\documentclass{article}
\usepackage[utf8]{inputenc}
\usepackage{jfrExamplee}
\usepackage{apalike}
\usepackage{graphicx}
\usepackage{amsmath,amssymb}
\usepackage{commath}
\usepackage[range-phrase=--,range-units=single,per-mode=symbol]{siunitx}
\usepackage{bm}
\usepackage{subfigure}
\usepackage{bigints}
\usepackage{todonotes}
\usepackage{booktabs}
\usepackage{hyperref}
\usepackage{xcolor}
\hypersetup{
  colorlinks,
  linkcolor={red!50!black},
  citecolor={blue!50!black},
  urlcolor={blue!80!black}
}

\newcommand{\T}{\mathsf{T}}

\title{Fast, Autonomous Flight in GPS-Denied and Cluttered Environments}
\author{
Kartik Mohta%\thanks{Use footnote for providing further information about author (webpage, alternative address). %Acknowledgments to funding agencies should go in the \textbf{Acknowledgments} section at the end of the paper.}%
\\
GRASP Lab\\
University of Pennsylvania\\
Philadelphia, PA, USA\\
\texttt{kmohta@seas.upenn.edu}\\
\And
Michael Watterson \\
GRASP Lab\\
University of Pennsylvania\\
Philadelphia, PA, USA \\
\texttt{wami@seas.upenn.edu}
\And
Yash Mulgaonkar \\
GRASP Lab\\
University of Pennsylvania\\
Philadelphia, PA, USA \\
\texttt{yashm@seas.upenn.edu}
\And
Sikang Liu \\
GRASP Lab\\
University of Pennsylvania\\
Philadelphia, PA, USA \\
\texttt{sikang@seas.upenn.edu}
\And
Chao Qu \\
GRASP Lab\\
University of Pennsylvania\\
Philadelphia, PA, USA \\
\texttt{quchao@seas.upenn.edu}
\And
Anurag Makineni \\
GRASP Lab\\
University of Pennsylvania\\
Philadelphia, PA, USA \\
\texttt{makineni@seas.upenn.edu}
\And
Kelsey Saulnier \\
GRASP Lab\\
University of Pennsylvania\\
Philadelphia, PA, USA \\
\texttt{saulnier@seas.upenn.edu}
\And
Ke Sun \\
GRASP Lab\\
University of Pennsylvania\\
Philadelphia, PA, USA \\
\texttt{sunke@seas.upenn.edu}
\And
Alex Zhu \\
GRASP Lab\\
University of Pennsylvania\\
Philadelphia, PA, USA \\
\texttt{alexzhu@seas.upenn.edu}
\And
Jeffrey Delmerico \\
Robotics and Perception Group \\
University of Zurich \\
Zurich, Switzerland \\
\texttt{jeffdelmerico@ifi.uzh.ch}
\And
Konstantinos Karydis \\
GRASP Lab\\
University of Pennsylvania\\
Philadelphia, PA, USA \\
\texttt{kkarydis@seas.upenn.edu}
\And
Nikolay Atanasov \\
GRASP Lab\\
University of Pennsylvania\\
Philadelphia, PA, USA \\
\texttt{atanasov@seas.upenn.edu}
\And
Giuseppe Loianno \\
GRASP Lab\\
University of Pennsylvania\\
Philadelphia, PA, USA \\
\texttt{loiannog@seas.upenn.edu}
\And
Davide Scaramuzza \\
Robotics and Perception Group \\
University of Zurich \\
Zurich, Switzerland \\
\texttt{sdavide@ifi.uzh.ch}
\And
Kostas Daniilidis \\
GRASP Lab\\
University of Pennsylvania\\
Philadelphia, PA, USA \\
\texttt{kostas@seas.upenn.edu}
\And
Camillo Jose Taylor \\
GRASP Lab\\
University of Pennsylvania\\
Philadelphia, PA, USA \\
\texttt{cjtaylor@seas.upenn.edu}
\And
Vijay Kumar \\
GRASP Lab\\
University of Pennsylvania\\
Philadelphia, PA, USA \\
\texttt{kumar@seas.upenn.edu}
}

\begin{document}

\maketitle

\begin{abstract}
One of the most challenging tasks for a flying robot is to autonomously navigate between target locations quickly and reliably while avoiding obstacles in its path, and with little to no a-priori knowledge of the operating environment. This challenge is addressed in the present paper. We describe the system design and software architecture of our proposed solution, and showcase how all the distinct components can be integrated to enable smooth robot operation. We provide critical insight on hardware and software component selection and development, and present results from extensive experimental testing in real-world warehouse environments. Experimental testing reveals that our proposed solution can deliver fast and robust aerial robot autonomous navigation in cluttered, GPS-denied environments.
\end{abstract}
%to get from point A to point B

\section{Introduction} \label{sec:introduction}
In recent times, there has been an explosion of research on micro-aerial vehicles (MAVs), ranging from low-level control \cite{Lee2010} to high-level, specification-based planning \cite{Wolff2014}. One class of MAVs, the quadrotor, has become popular in academia and industry alike due to its mechanical and control simplicity, high maneuverability and low cost of entry point compared to other aerial robots~\cite{Karydis_RSIF_17}. Indeed, there have been numerous applications of quadrotors to fields such as Intelligence, Surveillance and Reconnaissance (ISR), aerial photography, structural inspection \cite{Ozaslan2016}, robotic first responders \cite{Mohta2016}, and cooperative construction \cite{Augugliaro_CSM_14} and aerial manipulation \cite{Thomas_BB_2014}. Such recent advances have pushed forward the capabilities of quadrotors.

Most of these works rely on the availability of ground truth measurements to enable smooth robot performance. Ground truth can be provided by motion capture systems in lab settings, GPS in outdoor settings, or by appropriately placed special tags in known environments. However, real-world environments are dynamic, partially-known and often GPS-denied. Therefore, there is need to push further on developing fully autonomous navigation systems that rely on onboard sensing only in order to fully realize the potential of quadrotors in real-world applications. Our work aims at narrowing this gap.

Specifically, the focus of this paper is to provide a detailed description of a quadrotor system that is able to navigate at high speeds between a start and a goal pose (position and orientation) in cluttered 3D indoor and outdoor environments while using only onboard sensing and computation for state estimation, control, mapping and planning. The motivation for this problem comes from the recently announced DARPA Fast Lightweight Autonomy program\footnote{\url{http://www.darpa.mil/program/fast-lightweight-autonomy}}. The main challenge in creating such small, completely autonomous MAVs is due to the size and weight constraints imposed on the payload carried by these platforms. This restricts the kinds of sensors and computation that can be carried by the robot and requires careful consideration when choosing the components to be used for a particular application. Also, since the goal is fast flight, we want to keep the weight as low as possible in order to allow the robot to accelerate, decelerate and change directions quickly.

Any robot navigation system is composed of the standard building blocks of state estimation, control, mapping and planning. Each of these blocks builds on top of the previous ones in order to construct the full navigation system. For example, the controller requires a working state estimator while the planner requires a working state estimator, controller and mapping system. Initial works on state estimation for aerial robots with purely onboard sensing used laser rangefinders as the sensing modality due to the limited computational capacity available on the platforms \cite{Bachrach2009,Achtelik2009,Grzonka2009}. Due to the limitations of laser scan matching coupled with limited computational capability, these works were limited to slow speeds. Since the small and lightweight laser rangefinders that could be carried by the aerial robots only measured distances in a single plane, these methods also required certain simplifying assumptions about the environment, for example assuming 2.5D structure. As the computational power grew and more efficient algorithms were proposed, it became possible to use vision for state estimation for aerial robots \cite{Achtelik2009,Blosch2010}. Using cameras for state estimation allowed flights in 3D unstructured environments and also faster speeds \cite{Shen2013}. As the vision algorithms have improved \cite{Klein2007,Mourikis2007,Jones2011,Forster2014,Mur-Artal2015,Bloesch2015} and the computational power available on small computers has grown, cameras have now become the sensor of choice for state estimation for aerial robots. We use the visual odometry algorithm described in \cite{Forster2016svo} for our platform due to the fast run time, ability to use wide angle lenses without requiring undistortion of the full image, allowing the use of multiple cameras to improve robustness of the system and incorporation of edgelet features in addition to the usual point features.

The dynamics of the quadrotor are nonlinear due to the rotational degrees of freedom. In the control design for these robots, special care has to be taken in order to take this nonlinearity into account in order to utilize the full dynamics of the robot. Most early works in control design for quadrotors \cite{Bouabdallah2004,Bouabdallah2005,Escareno2006,Hoffmann2007,Bouabdallah2007} used the small angle approximation for the orientation controller to convert the problem into a linear one and proposed PID and backstepping controllers to stabilize the simplified system. Due to the small angle assumption, these controllers are not able to handle large orientation errors and have large tracking errors for aggressive trajectories. A nonlinear controller using quaternions (instead of Euler angles) was developed in \cite{Guenard2005} where the quadrotor was commanded to follow velocity commands. \cite{Lee2010} defined an orientation error metric directly in the SO(3) space and proposed a globally asymptotic controller that can stabilize the quadrotor from large position and orientation errors. Our controller is based upon this work and has good tracking performance even when following aggressive trajectories.

It has been shown that the trajectory generation for multi-rotor MAVs can be formulated as a Quadratic Program (QP) \cite{Mellinger2011}. Since the quadrotor is a differentially flat system, the trajectory can be optimized as an $n^{th}$ order polynomial parameterized in time \cite{Mellinger2011}. Generating a collision-free trajectory is more complicated, in which additional constraints for collision checking are required. Using Mixed Integer optimization methods to solve this problem has been discussed in \cite{Mellinger2012} and recently other approaches have been proposed to remove the integer variables and solve the QP in a more efficient way \cite{richter2016polynomial,Deits2015,Watterson2015}. Our pipeline uses a linear piece-wise path from a search-based planning algorithm to guide the convex decomposition of the map to find a safe corridor in free space as described in \cite{Sikang2016}. The safe corridor is formed as linear equality constraints in the QP for collision checking. We also consider dynamic constraints on velocity, acceleration and jerk in the QP to ensure that the generated trajectory does not violate the system's dynamics. In order to increase the safety, we propose a modified cost functional in the trajectory generation step such that the generated trajectory will be close to the center of the safe corridor.
%We show that these trajectories can decrease the probability of collision by a large amount.

We couple our trajectory generation method described above with a receding horizon method \cite{Bellingham2002} for replanning. As the robot moves, we only keep a local robot centric map and use a local planner to generate the trajectory. The main reasons behind this approach are: first, updating and planning in global map is expensive; second, the map far away from the robot is less accurate and less important. Since we are using a local planning algorithm, dead-ends are a well-known challenge. In order to efficiently solve this, we build a hybrid map consisting of a 3D local map and a 2D global map and our planner searches in this hybrid map to provide a globally consistent local action.

The purpose of a navigation system is to enable a robot to successfully traverse from a start pose to a goal pose in either a known or an unknown environment. The problem of navigating in an unknown environment is especially difficult, because in addition to having good state estimation and control, the robot needs to build an accurate map as it moves, and also generate collision-free trajectories quickly in the known map so that the replanning can be done at a high rate as the robot gets new sensor data. The initial works on navigation in unknown environments with quadrotors used offboard computation in order to run the planning due to the limited computation capability available on the platforms. \cite{He2008} presented a navigation system that uses a known map and takes the localization sensor model into account when planning so as to avoid regions that would lead to bad localization quality. \cite{Grzonka2009} demonstrated a quadrotor system that is able to localize and navigate in a known map using laser rangefinders as the main source of localization and mapping. Both of these transferred the sensor data to an offboard computer for processing. With a more powerful computer onboard the robot, \cite{Bachrach2011} were able to run a scan matching based localization system and the position and orientation controllers on the robot while the planner and a SLAM system to produce a globally consistent map ran on an offboard computer. \cite{Shen2011} was the first to demonstrated a full navigation system running onboard the robot without a known map of the environment. Since then, multiple groups have demonstrated similar capabilities \cite{Valenti2014,Schmid2014}.

In this paper, we describe our navigation system that allows a quadrotor to go from a starting position to a goal location while avoiding obstacles during the flight. We believe this is one of the first systems that is capable of fast aerial robot navigation through cluttered GPS-denied environments using only onboard sensing and computation. The navigation system has been tested thoroughly in the lab and in real world obstacle-rich environments that were set up as part of the DARPA FLA program.

This paper is organized as follows. In Section~\ref{sec:system_design} we describe our platform and the design decisions made in order to choose the current configuration. In Section~\ref{sec:estimation_and_control}, we describe our estimation and control algorithms. In this section, we also elaborate upon our sensor fusion methodology that is crucial to get good state estimates in order to control the robot. Section~\ref{sec:planning} describes our mapping, planning and trajectory generation modules. In Section~\ref{sec:experimental_results} we show results from various experiments performed in order to benchmark and test our full navigation system. Finally, we conclude in Section~\ref{sec:discussion_conclusion} with some discussion about the results and give some directions of future work that would help improve our system.

\section{System Design} \label{sec:system_design}
In this section we describe our overall system design. Specifically, we discuss platform design considerations, describe our computation, sensing and communication modules, and highlight critical software architecture components that enable the system to operate smoothly.

\begin{figure}[ht]
  \centering
  \includegraphics[width=0.5\columnwidth]{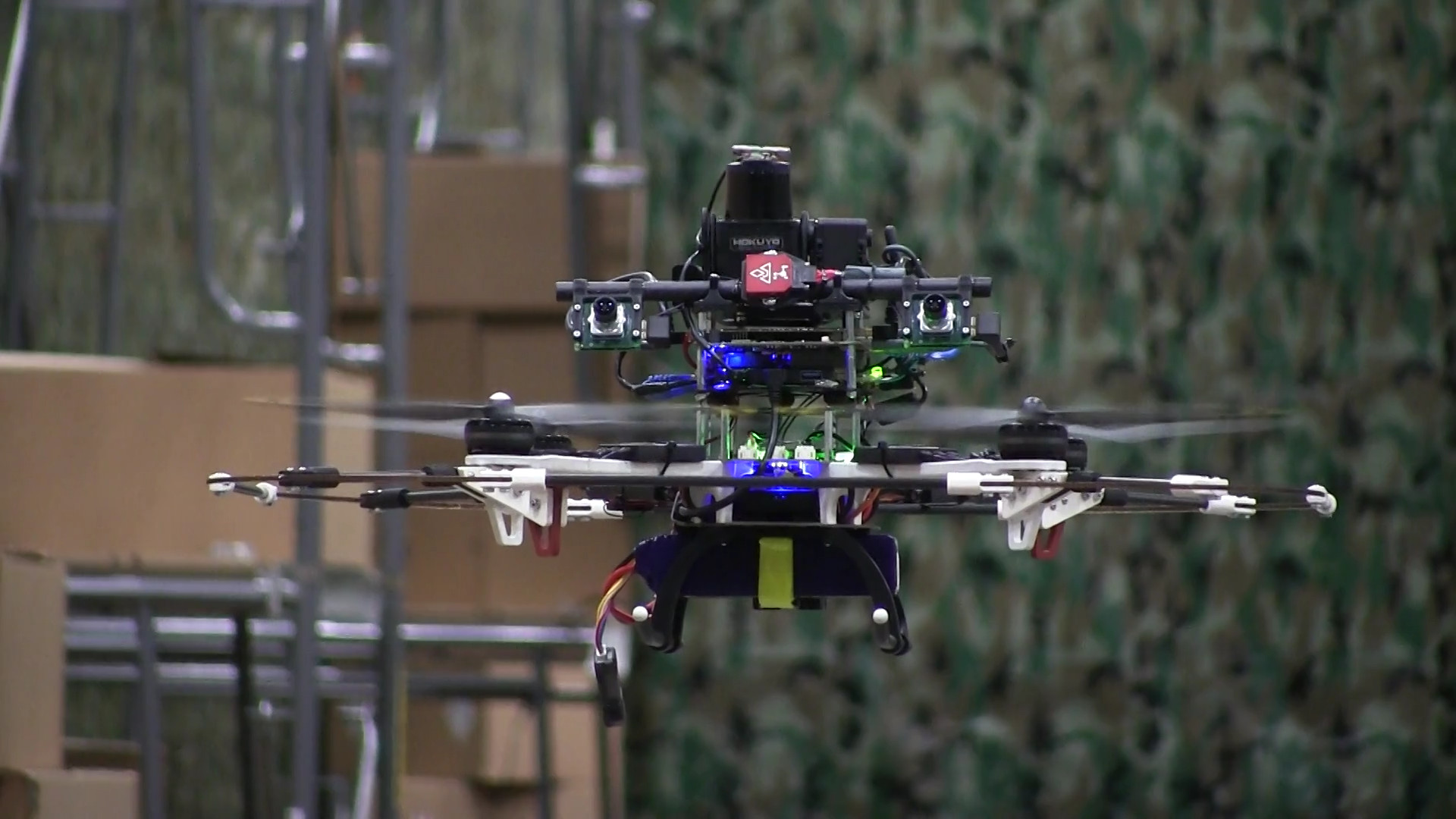}
  \caption{Our robot configuration showing the platform with stereo cameras and a nodding lidar.}
  \label{fig:robot}
\end{figure}

\subsection{Platform Design}
The guiding principle in the design of the platform was fast and agile flight. The desired capability of the platform was to be able to reach speeds of more than \SI{20}{\meter\per\second} while avoiding obstacles. This leads to a secondary and stronger requirement that the platform has to be able to stop from those speeds within typical sensor detection distances, which are around \SIrange{20}{25}{\meter}. This implies that the platform should be capable of accelerations of up to \SI{10}{\meter\per\square\second}. Reaching such high accelerations while maintaining the altitude requires a thrust-to-weight ratio of at around $1.5$. In order to have some margin for control during these high acceleration phases, we searched for an off-the-shelf platform that had sufficient thrust to provide a thrust-to-weight ratio of more than $2.0$ when fully loaded. This included an expected sensing and computation payload of up to \SI{1}{\kilo\gram} and a battery sufficient for desired flight time of around \SI{5}{\minute}. A list of various commercially available options is shown in Table~\ref{tab:platform_choices}.

\begin{table}[ht]
  \caption{Specifications of different commercially available off the shelf platforms. We expect a sensing and computation payload of approximately \SI{1}{\kilo\gram}, which has been added in the All-up mass. The mass of the battery is based upon the recommended battery for each platform.}
\label{tab:platform_choices}
\begin{center}
\begin{tabular}{lccccc}
\toprule
  \textbf{Platform} & \textbf{Frame} & \textbf{Battery} & \textbf{All-up} & \textbf{Max Thrust} & \textbf{Thrust/Weight} \\
  & \textbf{(kg)} & \textbf{(kg)} & \textbf{(kg)} & \textbf{(kgf)} & \textbf{Ratio} \\
  \midrule
3DR X8+ & 1.855 & 0.817 (4S) & 3.672 & 10.560 & 2.876 \\
DJI F550 + E310 & 1.278 & 0.600 (4S) & 2.878 & 5.316 & 1.847 \\
DJI F550 + E600 & 1.494 & 0.721 (6S) & 3.215 & 9.600 & 2.986 \\
DJI F450 + E310 & 0.826 & 0.400 (3S) & 2.226 & 3.200 & 1.438 \\
DJI F450 + E600 & 0.970 & 0.721 (6S) & 2.691 & 6.400 & 2.378 \\
\bottomrule
\end{tabular}
\end{center}
\end{table}

Based on the survey of the available platforms, we selected the platform configuration consisting of the DJI Flamewheel 450 base along with the DJI E600 motors, propellers and speed controllers since it closely matches our performance requirement. Each of the E600 motor and propeller combination has a rated maximum thrust of approximately \SI{1.6}{\kilo\gram f}. This leads to total thrust of around \SI{6.4}{\kilo\gram f} for our quadrotor configuration. For the low-level controller, we selected the Pixhawk \cite{pixhawk} which is a popular open-source autopilot. The main reason behind choosing the Pixhawk is that the firmware is open-source and customizable, giving us the capability of easily modifying or adding low-level capabilities as desired. In comparison, most of the commercially available autopilot boards are usually black boxes with an interface to send commands and receive sensor data. The base platform consisting of the F450 frame,  E600 propulsion system and the Pixhawk has a mass of approximately \SI{1.1}{\kilo\gram}. Adding the sensing and computation payload leads to a platform weight of \SI{2.1}{\kilo\gram} without the battery. In order to achieve the flight time requirement, we need to select the correct battery taking into account that the maximum total mass of the platform should be below \SI{3}{\kilo\gram}.

\begin{figure}[ht]
  \centering
  \includegraphics[width=0.5\columnwidth]{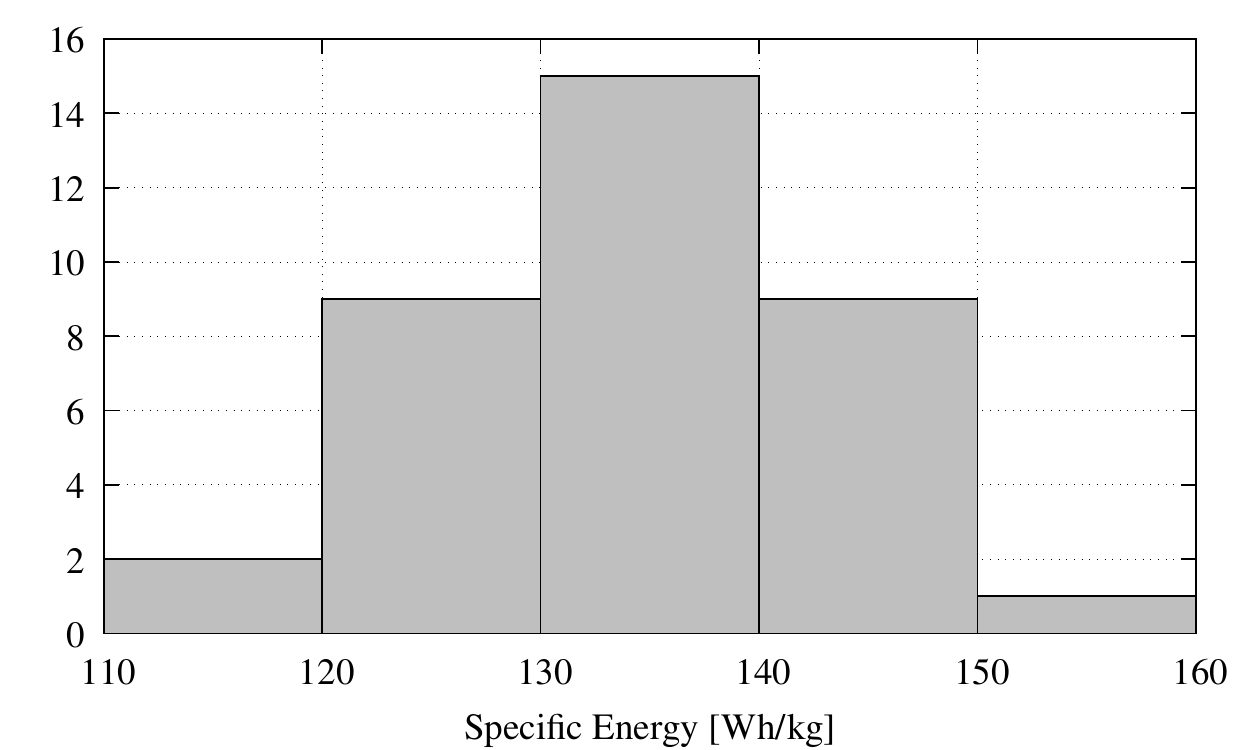}
  \caption{Histogram of specific energy values for a set of 36 6S rated hobby grade lithium polymer batteries.}
  \label{fig:specific_energy}
\end{figure}

The batteries used in MAVs are based on lithium polymer chemistry due to their high energy and power densities. The DJI E600 propulsion system requires a 6S battery, i.e.\ a battery with rated voltage of approximately \SI{22.2}{\volt}. Given that, the main design choice available is the battery capacity. Typical hobby grade lithium polymer batteries have specific energy values are around \SIrange{130}{140}{\watt\hour\per\kilo\gram} (Fig.~\ref{fig:specific_energy}). The power required to hover for quadrotors is approximately \SI{200}{\watt\per\kg} \cite{Mulgaonkar2014SPIEDSS}, so for a platform with a total mass between \SIrange{2.5}{3}{\kilo\gram}, the power consumption would be \SIrange{500}{600}{\watt}. Assuming an overall efficiency of around $60\%$, \cite{Theys2016} going from the supplied power from the battery to the mechanical power output at the propellers, the energy capacity of the battery for a \SI{5}{\minute} flight time needs to be approximately \SIrange{69.4}{83.3}{\watt\hour}. In practice, we never use the full capacity of the battery in order to preserve the life of the battery and also to have some reserve capacity for unforeseen circumstances. If we only use $80\%$ of the rated capacity of the battery, it leads to a required battery energy capacity of \SIrange{86.8}{104.1}{\watt\hour}. Using the average specific energy value of \SI{135}{\watt\hour\per\kilo\gram}, we expect the mass of the battery to be between \SIrange{0.64}{0.77}{\kilo\gram} which fits in well with our total mass budget. Based on the available battery capacities, we selected batteries with capacities of \SI{88.8}{\watt\hour} and \SI{99.9}{\watt\hour} in order to provide some flexibility in terms of having higher performance or higher endurance.

\begin{figure}[ht]
  \centering
  \includegraphics[width=0.5\columnwidth]{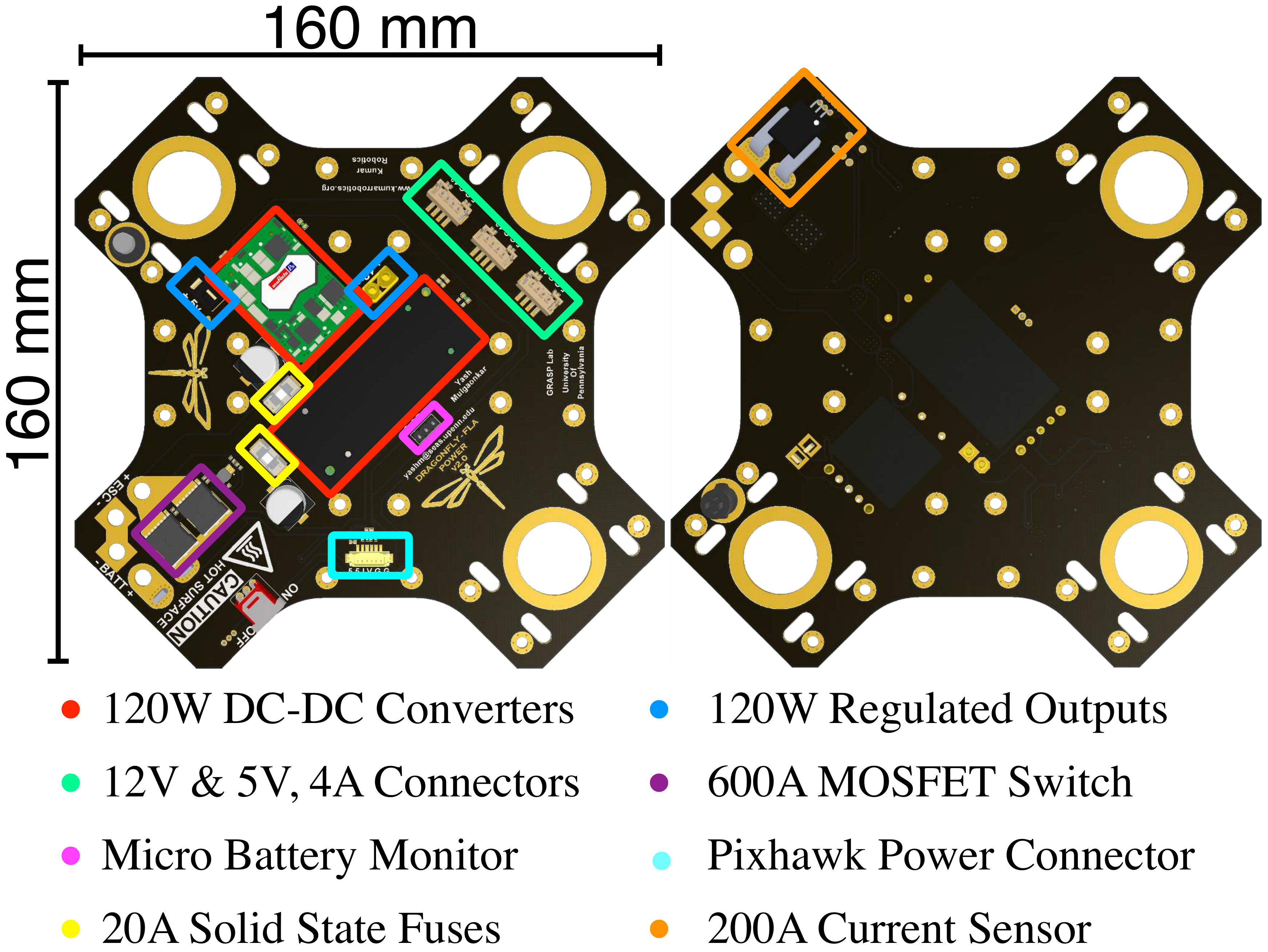}
  \caption{Power Distribution Board.}
  \label{fig:pdb}
\end{figure}

In order to power all the sensors and the computer onboard the robot, we required regulated power supplies for \SI{12}{\volt} and \SI{5}{\volt}. We designed a custom power distribution board, shown in Fig.~\ref{fig:pdb}, consisting of a power conditioning circuit, DC-DC converters, power connectors, and a battery monitor. The board is capable of providing filtered \SI{12}{\volt} and \SI{5}{\volt} supply at a maximum of \SI{120}{\watt} each. In addition to power management, for weight saving reasons, the board replaces the top plate of the standard commercially available configuration, forming an integral part of the robot frame.

\subsection{Sensing, Computation and Communication}
The robot needs to navigate through cluttered 3D environments with purely on-board sensing and computation. This requires the correct selection of sensors and the onboard computer in order to be able to perform the desired task while keeping the mass low. The two tasks that the robot has to perform which require proper sensor selection are state estimation and mapping. The two solutions for state estimation for MAVs are either vision based or lidar based. For unstructured 3D environments, the vision based systems have been more successful that lidar based ones, so we decided on using cameras as our primary state estimation sensors. More details about why the stereo configuration was selected are provided in Section~\ref{sec:estimation_and_control}. In addition to the cameras, we added a downward pointing lidar (Garmin Lidar-Lite) and a VectorNav VN-100 IMU for state estimation. The VN-100 IMU is also used to trigger the capture from the cameras in order to have time synchronization between the cameras and IMU.

\begin{figure}[ht]
  \centering
  \includegraphics[width=0.4\columnwidth]{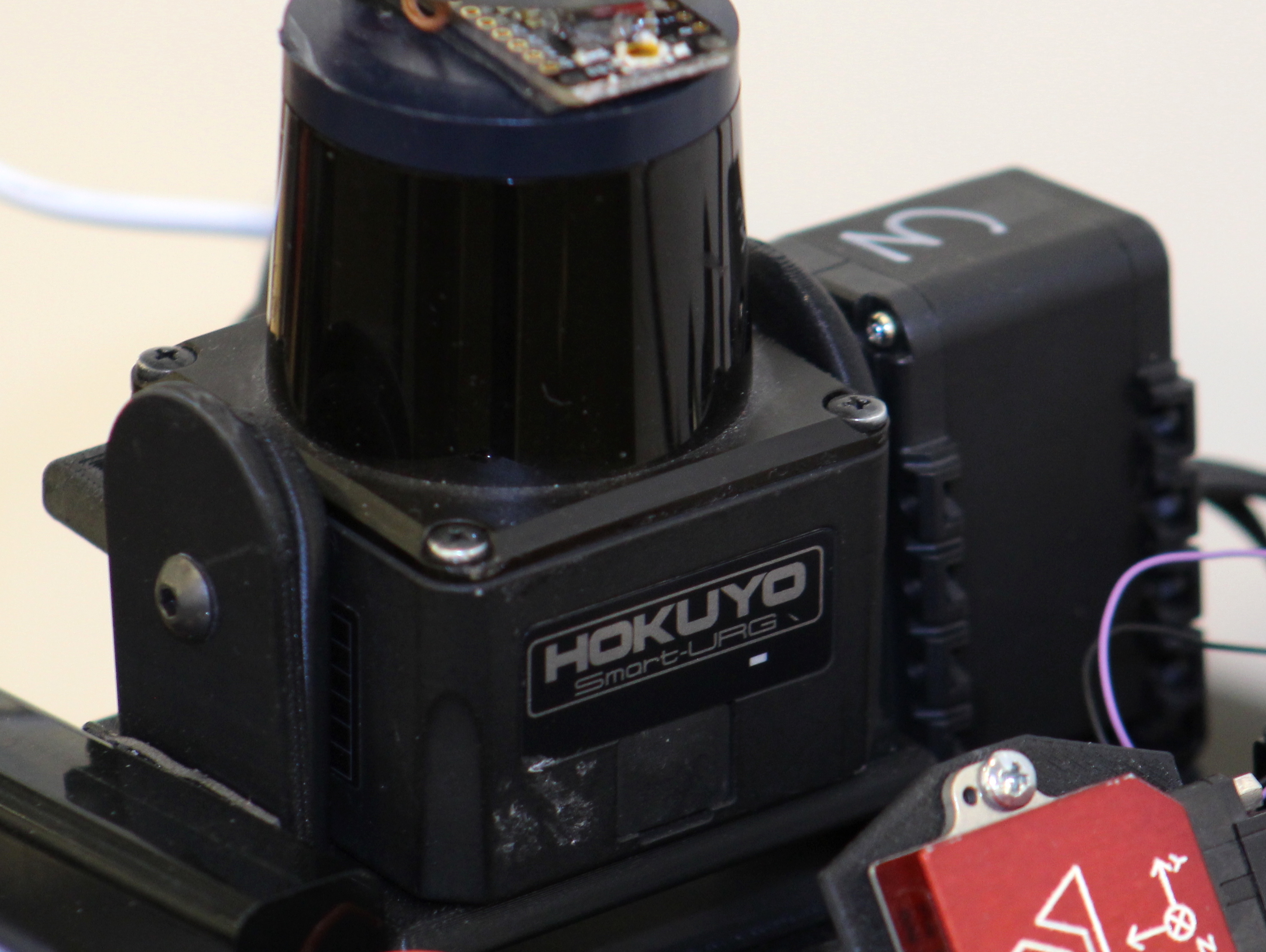}
  \caption{Our mapping solution consisting of a 2D lidar mounted on a nodding gimbal.}
  \label{fig:nodding_lidar}
\end{figure}

The situation for mapping is a bit different. Current vision based dense mapping algorithms are either not accurate enough or too computationally expensive to run in real time, so lidar based mapping is still the preferred choice for MAVs. In order to keep our weight low, we decided to use a Hokuyo 2D lidar instead of a heavy 3D lidar. We still required a 3D map for planning, so we decided to mount the 2D lidar on a one degree of freedom nodding gimbal as shown in Fig.~\ref{fig:nodding_lidar}.

In order to handle all the computations for estimation, control, mapping and planning onboard the robot, we selected the Intel NUC i7 computer. This single board computer is based on the Intel i7-5557U processor and supports up to 16GB of RAM and an M.2 SSD for storage. This provides sufficient computing power to run our full software stack on the robot without overloading the CPU and also gives us ample amount of storage for recording sensor data for long flights. While the robot is flying, we need to have a communication link in order to monitor the status of the various modules running on the robot. We wanted a link that has good bandwidth, so that during development we can stream the sensor data back to the base station, but also good range so that we do not loose the link when running long range (up to \SI{200}{\meter}) experiments. In addition, since we use ROS as our software framework, having a wireless link that behaves like a wireless local area network was preferred in order to be able to use the standard ROS message transport mechanism. Based on these requirements, we selected the Ubiquity Networks Picostation~M2 for the robot side and the Nanostation~M2 for the base station. These are high power wireless radios that incorporate Ubiquity Networks' proprietary airMAX protocol, which improves latency and reliability for long range wireless links compared to the 802.11 protocol, which was designed mainly for indoor use. The Picostation is the smaller and lighter of the two, weighing at around \SI{50}{\gram} (after taking off the outer plastic case) compared to \SI{400}{\gram} for the Nanostation. This lower weight comes with the compromise of lower transmit power and lower bandwidth, but the performance was sufficient for our purpose, providing a bandwidth of more than \SI{50}{\mega bps} up to distances of \SI{200}{\meter}.

\subsection{Software Architecture}

\begin{figure}[ht]
  \centering
  \includegraphics[width=0.6\columnwidth]{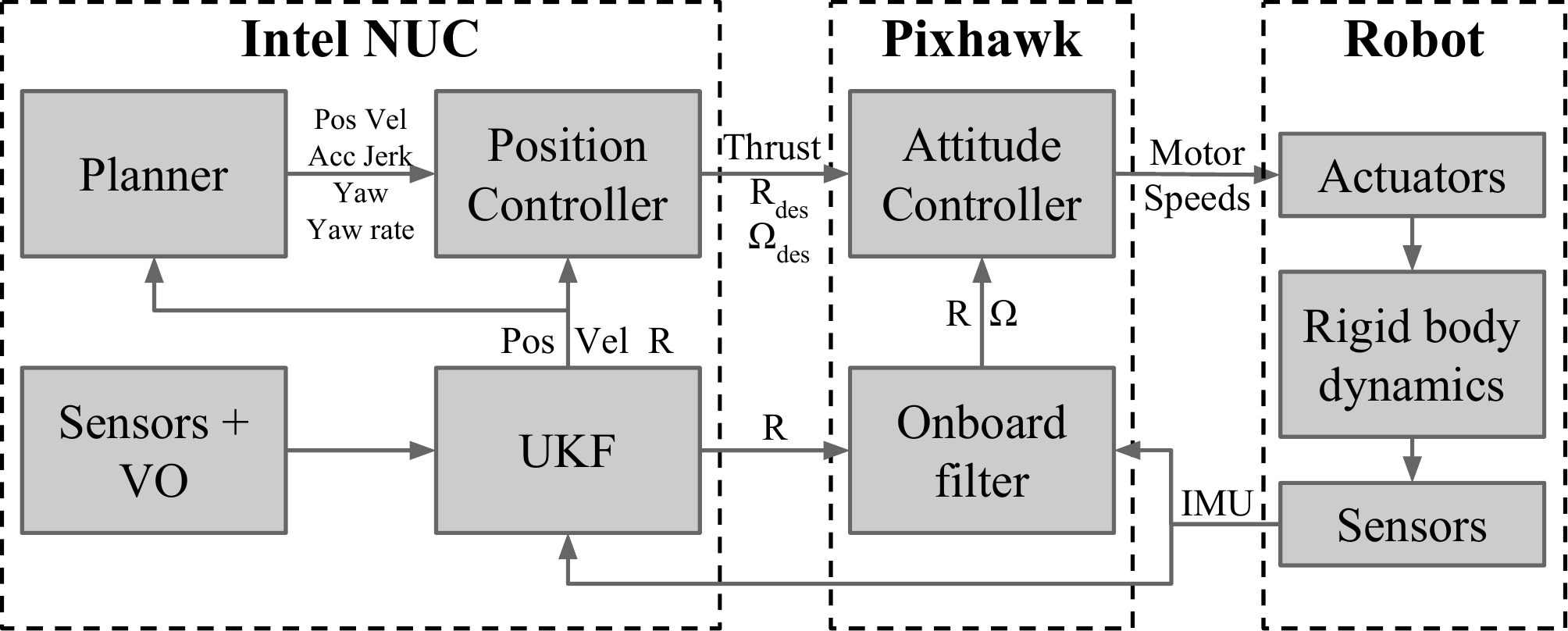}
  \caption{A high level block diagram of our system architecture.}
  \label{fig:software_architecture}
\end{figure}

Any big system requires all of the individual components to work together in order to allow the full system to function properly. Figure~\ref{fig:software_architecture} shows a high level block diagram of our system illustrating the different components and how they are connected to each other. The software components in our system can be grouped under four categories: Estimation, Control, Mapping and Planning. Each of these is in turn separated into smaller parts, and we use ROS as the framework for all the high level software running on the robot. ROS is chosen because it provides a natural way to separate each component into its own package allowing distributed development and ease of testing and debugging. Each executable unit in a ROS system is called a node and different nodes communicate with each other using message passing. In this way, a ROS system can be thought of as a computational graph consisting of a peer-to-peer network of nodes processing and passing data among them. One convenient feature of this system is that the nodes can be run on different computers, since the message passing uses the TCP transport, which allows us to run a subset of the nodes on the robot while the remaining can be run on a workstation computer making it easier to analyze and debug problems leading to a faster development phase. We also benefit from the whole ROS ecosystem of tools and utilities that have been developed in order to perform routine but useful tasks when developing a system such as tools for logging and playing back the messages passed between nodes or tools to visualize the data being sent between nodes.

\section{Estimation and Control} \label{sec:estimation_and_control}
There has been a lot of research in recent times on visual and visual-inertial odometry for MAVs with a variety of proposed algorithms \cite{Klein2007,Mourikis2007,Jones2011,Forster2014,Mur-Artal2015,Bloesch2015}. The algorithms can be classified based on the number and type of cameras required into three groups: monocular, stereo, or multi-camera. There are also algorithms using depth cameras but these cameras don't work well outdoors with sunlight so we do not consider them. An overview of the advantages and disadvantages of the algorithms is shown in Table~\ref{tab:vo_algorithms}. Looking at these, it is clear that the multi-camera setup would be the most preferred but the software complexity is still a hurdle in terms of real-world usage. Monocular algorithms have received a lot of research attention in the last few years and have improved to a level that they can be reliably used as the only source of odometry for a MAV system. One problem of the monocular algorithms is that they require an initialization process during which the estimates are either not available or are not reliable. In comparison, the stereo algorithms can be initialized using a single frame making them much more robust in case the algorithm needs to be reinitialized while flying if, for example, there is a sudden rotation. One more advantage of using stereo cameras is that in the extreme case that stereo matching is not possible due to features being too far away, we can use the input from only one of the cameras from the stereo pair and treat it as a monocular camera setup.

\begin{table}[ht]
\caption{Advantages and disadvantages of different visual odometry algorithms.}
\label{tab:vo_algorithms}
\begin{center}
\begin{tabular}{lccc}
\toprule
& \textbf{Monocular} & \textbf{Stereo} & \textbf{Multi-camera} \\
\midrule
{Mechanical complexity} & Low & Medium & Low \\
{Software complexity} & Medium & Low-Medium & High \\
{Robustness} & Low-Medium & Medium & High \\
{Feature distance} & High & Medium-High & High \\
\bottomrule
\end{tabular}
\end{center}
\end{table}

\subsection{Visual Odometry}

% SVO notation
\newcommand{\Camera}{\text{C}}
\newcommand{\camera}{\text{\tiny{C}}}
\newcommand{\cam}{\text{\tiny{C}}}
\newcommand{\Body}{\text{B}}
\newcommand{\body}{\text{\tiny{B}}}
\newcommand{\Image}{\mathtt{I}}
\newcommand{\pt}{\boldsymbol{\rho}}
\newcommand{\px}{\mathbf{u}}
\newcommand{\Real}{\mathbb{R}}
\newcommand{\depth}{\rho}
\newcommand{\proj}{\pi}
\newcommand{\World}{\text{W}}
\newcommand{\world}{\text{\tiny{W}}}
\newcommand{\Tran}{\mathtt{T}}
\newcommand{\Tcb}{\Tran_{\cam\body}}
\newcommand{\Tbc}{\Tran_{\body\cam}}
\newcommand{\SEthree}{\mathrm{SE}(3)}
\newcommand{\Trel}{{\Tran_{kk-1}}}
\newcommand{\Trelopt}{{\Tran_{kk-1}^\star}}
\newcommand{\residual}{\mathbf{r}}
\newcommand{\Ic}{\Image^\cam}
\newcommand{\Patch}{\mathcal{P}}
\newcommand{\transpose}{\mathsf{T}}
\newcommand{\inlierprob}{\gamma}
\newcommand{\R}{\mathtt{R}}
\newcommand{\tran}{\mathbf{p}}
\newcommand{\Identity}{\mathbf{I}}

\subsubsection{Overview of SVO}
% Summarize first 4 PPs of intro, cite TRO paper and refer there for more info
To estimate the six degree-of-freedom motion of the platform, we use the Semi-direct Visual Odometry (SVO) framework proposed in \cite{Forster2016svo}.
SVO combines the advantages of both \textit{feature-based} methods, which minimize the reprojection error of a sparse set of features, and\textit{direct} methods, which minimize photometric error between image pixels.
This approach estimates frame-to-frame motion of the camera by first aligning the images using a sparse set of salient features (e.g. corners) and their neighborhoods in the images, estimating the 3D positions of the feature points with recursive Bayesian depth estimation, and finally refining the structure and camera poses with optimization (e.g. bundle adjustment).
Our efficient implementation of this approach is capable of estimating the pose of a camera frame in as little as 2.5 milliseconds, while achieving comparable accuracy to more computationally intensive methods.

\begin{figure}[ht]
\centering
\includegraphics[trim={0cm 5cm 0cm 0cm}, clip, width=0.35\textwidth]{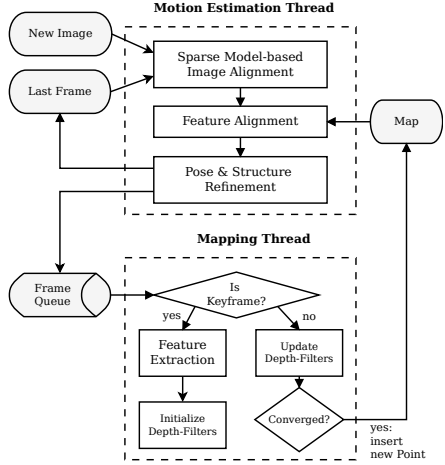}
\caption{A high level diagram of the SVO software architecture.}
\label{fig:svo_system_overview}
\end{figure}

% System overview + notation summary, reproduce figs 1
The system is decomposed into two parallel threads: one for estimating camera motion, and one for mapping the environment (see Fig. \ref{fig:svo_system_overview}).
The motion-estimation thread proceeds by first performing a sparse image alignment between the two most recent frames, then obtaining sub-pixel feature correspondence using direct methods on patches around each sparse feature, and finally pose refinement on the induced reprojection error.

A brief overview of the notation used in describing the method is presented here.
The intensity image from camera $\Camera$ at timestep $k$ is denoted by $\Image^\cam_k: \Omega^\cam \subset \mathbb{R}^2 \mapsto  \mathbb{R}$, where $\Omega^\cam$ is the image domain.
A point in 3D space $\pt\in\Real^3$ maps to image coordinates $\px\in\Real^2$ through the camera projection model, $\px = \pi(\pt)$, which we assume is known.
Given the inverse scene depth $\depth > 0$ for a pixel $\px\in \mathcal{R}^\cam_k$, the position of the corresponding 3D point is obtained using the back-projection model $\pt = \proj^{-1}_\depth(\px)$.
We denote $\mathcal{R}^\cam_k \subseteq \Omega$ the set of pixels for which the depth is known at time $k$ in camera $\Camera$.
The position and orientation of the world frame $\World$ with respect to the $k^{\text{th}}$ camera frame is described by the rigid body transformation $\Tran_{k\world} \in \SEthree$.
A 3D point ${}_\world\pt$ that is expressed in world coordinates can be transformed to the $k^{\text{th}}$ camera frame using: ${}_k\pt = \Tran_{k\world} \ {}_\world\pt$.

\subsubsection{Motion Estimation}
% Sparse image alignment: Fig. 2, eq 1, 2, 3 (skip eq describing backprojection to subset in FOV)
\begin{figure}[t]
  \centering
  \includegraphics[width=0.35\textwidth]{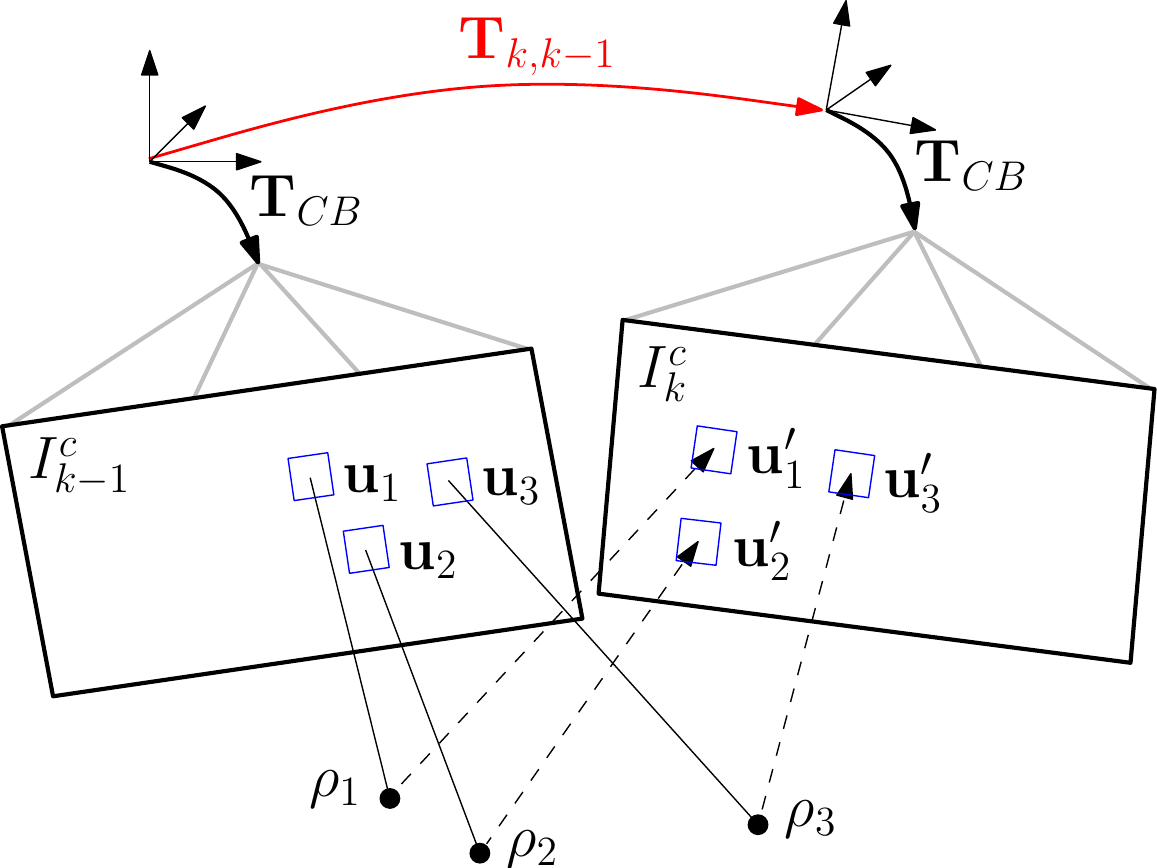}
  \caption{Changing the relative pose $\T_{k,k-1}$ between the current and the previous frame implicitly moves the position of the reprojected points in the new image $\px_i'$. Sparse image alignment seeks to find $\T_{k,k-1}$ that minimizes the photometric difference between image patches corresponding to the same 3D point (blue squares).
  }
  \label{fig:sparse_img_alignment}
\end{figure}

Consider a \emph{body frame} $\Body$ that is rigidly attached to the camera frame $\Camera$ with known extrinsic calibration $\Tcb \in \SEthree$ (see Fig. \ref{fig:sparse_img_alignment}).
Our goal is to estimate the incremental motion of the body frame $\Trel \doteq \Tran_{\body_k\body_{k-1}}$ such that the photometric error is minimized:
\begin{equation}\label{eq:img_align_cost}
    \Trelopt = \arg\min_\Trel
    \sum_{\px \in \bar{\mathcal{R}}_{k-1}^\cam}
    \frac{1}{2} \| \residual_{\Image_\px^\cam}(\Trel) \|^2_{\Sigma_\Image},
\end{equation}
where the photometric residual $\residual_{\Image_\px^\cam}$ is defined by the intensity difference of pixels in subsequent images $\Image_k^\cam$ and $\Image_{k-1}^\cam
$ that observe the same 3D point $\pt_\px$:
\begin{equation}
  \residual_{\Image_\px^\cam}(\Trel)
  \doteq
  \Ic_k\big( \proj ( \Tcb \Trel \ \pt_\px ) \big)  - \Ic_{k-1} \big( \proj ( \Tcb \ \pt_\px ) \big).
\end{equation}
The corresponding 3D point $\pt_\px$ for a pixel with known depth, expressed in the reference $\Body_{k-1}$ frame, is computed by means of back-projection:
\begin{equation}
  \pt_\px = \Tbc \ \pi^{-1}_\depth(\px), \qquad \forall \ \px \in \mathcal{R}_{k-1}^\cam.
\end{equation}

Sparse image alignment solves the non-linear least squares problem in Eq. \eqref{eq:img_align_cost} with $\mathcal{R}_{k-1}^\cam$ corresponding to small patches centered at features with known depth, using standard iterative non-linear least squares algorithms.

% Relaxation and refinement for corner and edge features
In the next step, we relax the geometric constraints given by the reprojection of 3D points and perform an individual 2D alignment of corresponding feature patches.
In order to minimize feature drift over the camera trajectory, the alignment of each patch in the new frame is performed with respect to a reference patch from the frame where the feature was first extracted.
However, this step generates a reprojection error, representing the difference between the projected 3D point and the aligned feature position.

We detect both edge and corner features, representing points of strong gradient in the image.
Feature alignment in the image minimizes the intensity difference of a small image patch $\Patch$ that is centered at the projected feature position $\px'$ in the newest frame $k$ with respect to a reference patch from the frame~$r$ where the feature was first observed.
For corner features, the optimization computes a correction $\delta\px^\star \in \Real^2$ to the predicted feature position $\px'$ that minimizes the photometric cost:
\begin{align}
  \label{eq:patch_alignment_2d}
  {\px'}^\star
    &= \px' + \delta\px^\star, \quad \text{with} \quad \px' = \proj\big(\Tcb \ \T_{kr} \ \Tbc \ \proj^{-1}_\depth(\px)\big) \\
  \delta\px^\star
    &=  \arg\min_{\delta\px} \sum_{\Delta\px \in \Patch} \!
        \frac{1}{2}
        \Big\|
          \Image_{k}^\cam \left( \px' \!+\! \delta\px \!+\! \Delta\px \right)
          - \Image_{r}^\cam(\px + \mathbf{A} \Delta\px)
        \Big\|^2 \!\! \nonumber,
\end{align}
where $\Delta\px$ is the iterator variable that is used to compute the sum over the patch $\Patch$.
For features on edges, we therefore optimize for a scalar correction $\delta u^\star \in \Real$ in the direction of the edge normal $\mathbf{n}$ to obtain the corresponding feature position ${\px'}^\star$ in the newest frame:
\begin{align}
  \label{eq:patch_alignment_1d}
  {\px'}^\star
    &= \px' + \delta u^\star \cdot \mathbf{n}, \quad \text{with} \\
  \delta u^\star \!
    &= \! \arg\min_{\delta u} \!\!\! \sum_{\Delta\px \in \Patch} \!
        \frac{1}{2}
        \Big\|
          \Image_{k}^\cam \left( \px' \!+\! \delta u \!\cdot\! \mathbf{n} \!+\! \Delta\px \right)
          \!-\! \Image_{r}^\cam(\px + \mathbf{A} \Delta\px)
        \Big\|^2 \!\! \nonumber.
\end{align}

% Optimization: Eq. 6, mention that it's possible to optimize the full trajectory, but we only optimize the latest pose and 3D points separately.
In the previous step, we established feature correspondence with subpixel accuracy, but this feature alignment violated the epipolar constraints and introduced a reprojection error $\delta\px$, which is typically well below 0.5 pixels.
Therefore, in the last step of motion estimation, we refine the camera poses and landmark positions $\mathcal{X} = \{\T_{k\world}, \pt_i\}$ by minimizing the squared sum of reprojection errors:
\begin{align}\label{eq:svo_bundle_adjustment}
  \mathcal{X}^\star
  = \arg\min_\mathcal{X}
  &\sum_{k\in\mathcal{K}} \sum_{i\in\mathcal{L}_k^C}
  \frac{1}{2} \|
    {\px'}^\star_i
    - \proj\big(\Tcb \ \T_{k\world} \ \pt_i \big)
   \|^2 \\
   +
   &\sum_{k\in\mathcal{K}} \sum_{i\in\mathcal{L}_k^E}
  \frac{1}{2} \|
    \mathbf{n}^\transpose_i \big( {\px'}^\star_i
    - \proj\big(\Tcb \ \T_{k\world} \ \pt_i \big) \big)
   \|^2 \nonumber
\end{align}
where $\mathcal{K}$ is the set of all keyframes in the map, $\mathcal{L}_k^C$ the set of all landmarks corresponding to corner features, and $\mathcal{L}_k^E$ the set of all edge features that were observed in the $k^\text{th}$ camera frame.

While optimization over the whole trajectory in Eq. \eqref{eq:svo_bundle_adjustment} with bundle adjustment results in higher accuracy, for MAV motion estimation, it is sufficient to only optimize the latest camera pose and the 3D points separately, which permits more efficient computation.

\subsubsection{Mapping}
% Depth filter, cite [55]: eq 7-10, Fig. 5
In the derivation of motion estimation, we assumed that the depth at sparse feature locations in the image was known.
Here, we describe how the mapping thread estimates this depth for newly detected features, assuming that the camera poses are known from the motion estimation thread.

\begin{figure}[ht]
  \centering
  \includegraphics[width=0.33\textwidth]{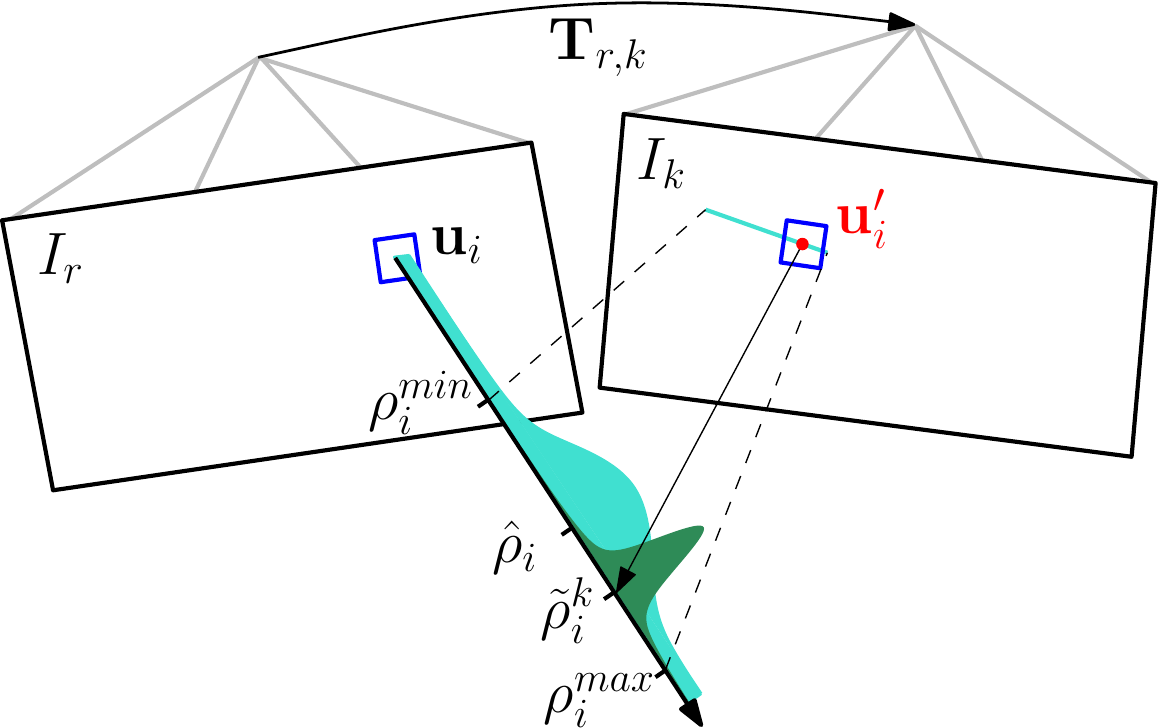}
  \caption{Probabilistic depth estimate $\hat{\depth}_i$ for feature $i$ in the reference frame $r$. The point at the true depth projects to similar image regions in both images (blue squares). Thus, the depth estimate is updated with the triangulated depth $\tilde{\depth}^k_{i}$ computed from the point $\px'_i$ of highest correlation with the reference patch. The point of highest correlation lies always on the epipolar line in the new image.}
  \label{fig:vogiatzis}
\end{figure}

The depth at a single pixel is estimated from multiple observations by means of a recursive Bayesian \emph{depth filter}.
When the number of tracked features falls below some threshold, a new keyframe is selected and new depth filters are initialized at corner and edge features in that frame.
Every depth filter is associated to this reference keyframe~$r$, and the initial depth uncertainty is initialized with a large value.
For a set of previous keyframes as well as every subsequent frame with known relative pose $\{\Image_k, \T_{kr}\}$, we search for a patch along the epipolar line that has the highest correlation via zero mean sum of squared differences (see Fig. \ref{fig:vogiatzis}).
From the pixel with maximum correlation, we triangulate the depth measurement $\tilde{\depth}_i^k$, which is used to update the depth filter.
If enough measurements have been obtained such that uncertainty in the depth is below a certain threshold, we initialize a new 3D point at the estimated depth in our map, which subsequently can be used for motion estimation (see system overview in Fig. \ref{fig:svo_system_overview}).

We model the depth filter according to \cite{Vogiatzis11jivc} with a two dimensional distribution: the first dimension is the inverse depth~$\depth$ \cite{Civera08tro}, while the second dimension~$\inlierprob$ is the inlier probability.
Hence, a measurement $\tilde{\depth}_i^k$ is modeled with a \emph{Gaussian + Uniform} mixture model distribution: an inlier measurement is normally distributed around the true inverse depth $\depth_i$ while an outlier measurement arises from a uniform distribution in the interval $[\depth_i^\text{min}, \depth_i^\text{max}]$:
\begin{equation}
  p(\tilde{\depth}_i^k|\depth_i,\inlierprob_i)
  =  \inlierprob_i\mathcal{N} \big( \tilde{\depth}_i^k \big| \depth_i,\tau_i^2 \big)
  + (1\!-\!\inlierprob_i) \mathcal{U}\big( \tilde{\depth}_i^k \big| \depth_i^\text{min}\!, \depth_i^\text{max} \big),
\end{equation}
where $\tau_i^2$ the variance of a good measurement that can be computed geometrically by assuming a disparity variance of one pixel in the image plane \cite{Pizzoli14icra}.
We refer to the original work \cite{Vogiatzis11jivc} and the \cite{Forster2016svo} for more details.

\subsubsection{Implementation Details}
% Multi-camera system: Fig. 9, eq. 11,
\begin{figure}[ht]
  \centering
  \includegraphics[trim=0pt 0pt 0pt 0pt, clip=true, width=0.4\linewidth]{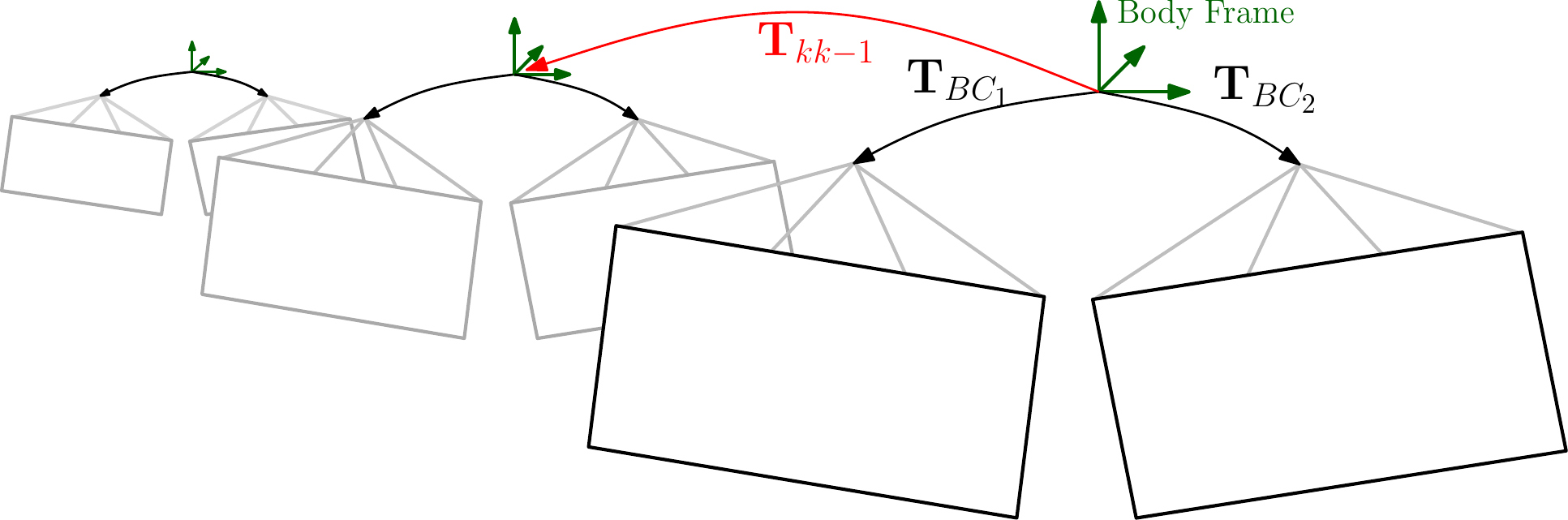}
  \caption{Visual odometry with multiple rigidly attached and synchronized cameras.
  The relative pose of each camera to the body frame $\T_{\body\cam_j}$ is known from extrinsic calibration and the goal is to estimate the relative motion of the body frame $\T_{kk-1}$.}
  \label{fig:camera_rig}
\end{figure}
Our system utilizes multiple cameras, so consider a camera rig with $M$ cameras (see Fig. \ref{fig:camera_rig}).
We assume that the relative pose of the individual cameras $c \in \Camera$ with respect to the body frame $\Tcb$ is known from extrinsic calibration.
To generalize sparse image alignment to multiple cameras, we simply need to add an extra summation in the cost function of Eq. \eqref{eq:img_align_cost}:
\begin{equation}
\label{eq:svo_img_align_multicam}
    \Trelopt = \arg\min_\Trel
    \sum_{\camera \in \Camera} \sum_{\px \in \bar{\mathcal{R}}_{k-1}^\cam}
    \frac{1}{2} \| \residual_{\Image_\px^\cam}(\Trel) \|^2_{\Sigma_\Image}.
\end{equation}
The same summation is necessary in the bundle adjustment step to sum the reprojection errors from all cameras.
The remaining steps of feature alignment and mapping are independent of how many cameras are used, except that more images are available to update the depth filters.
An initial map is computed during initialization using stereo matching.

% Motion Priors: eq. 12 - constant velocity, gyro for rot
We additionally apply motion priors within the SVO framework by assuming a constant velocity relative translation prior $\tilde\tran_{kk-1}$ and a relative rotation prior $\tilde\R_{kk-1}$ from a gyroscope.
We employ the motion prior by adding additional terms to the cost of the sparse image alignment step:
\begin{align}\label{eq:prior}
  \Trelopt
  =  \arg\min_{\T_{kk-1}}
    & \sum_{\camera \in \Camera} \sum_{\px \in \bar{\mathcal{R}}_{k-1}^\cam}
        \frac{1}{2}\| \residual_{\Image_\px^\cam}(\Trel) \|^2_{\Sigma_\Image}\\
    & + \frac{1}{2}\| \tran_{kk-1} - \tilde\tran_{kk-1}\|^2_{\Sigma_\tran} \nonumber \\
    & + \frac{1}{2}\| \log(\tilde\R_{kk-1}^\transpose \R_{kk-1})^\vee \|^2_{\Sigma_\R}, \nonumber
\end{align}
where the covariances $\Sigma_\tran, \Sigma_\R$ are set according to the uncertainty of the motion prior, the variables $(\tran_{kk-1}, \R_{kk-1}) \doteq \T_{kk-1}$ are the current estimate of the relative position and orientation (expressed in body coordinates $\Body$), and the \emph{logarithm map} maps a rotation matrix to its rotation vector.

% Summarize Sec. X
We apply the sparse image alignment algorithm in a coarse-to-fine scheme, half-sampling the image to create an image pyramid of five levels, and use a patch size of $4 \times 4$ pixels.
The photometric cost is then optimized at the coarsest level until convergence, starting from the initial condition $\T_{kk-1} = \Identity_{4\times 4}$, before continuing at the finer levels to improve the precision.

In the mapping thread, we divide the image in cells of fixed size (e.g., $32 \times 32$ pixels).
For every keyframe a new depth-filter is initialized at the FAST corner \cite{Rosten10pami} with highest score in the cell, unless there is already a 2D-to-3D correspondence present.
In cells where no corner is found, we detect the pixel with highest gradient magnitude and initialize an edge feature.
This results in evenly distributed features in the image.
To speed up the depth-estimation we only sample a short range along the epipolar line; in our case, the range corresponds to twice the standard deviation of the current depth estimate.
We use a $8 \times 8$ pixel patch size for the epipolar search.

We refer the reader to the original paper \cite{Forster2016svo} for further details about both the approach and its performance.

\subsection{Sensor Fusion}
\begin{figure}[ht]
\centering
  \includegraphics[width=0.7\columnwidth]{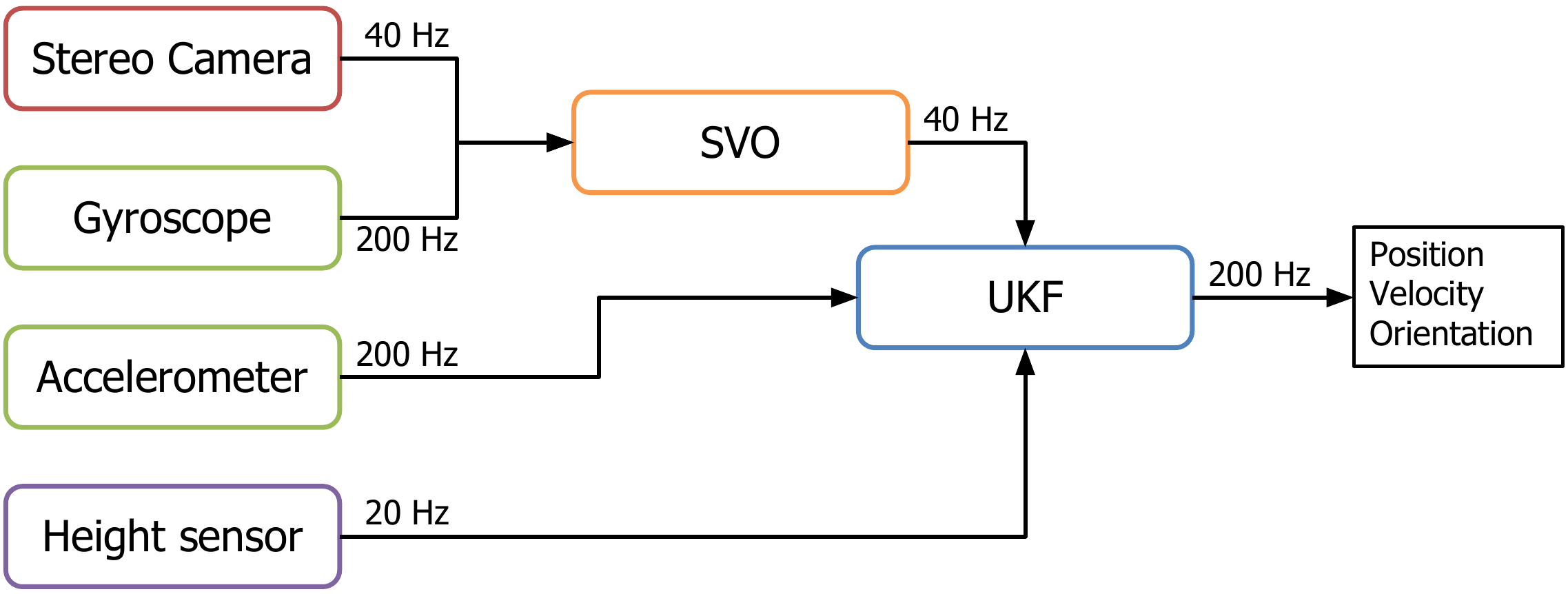}
\caption{Data flow diagram of the UKF used on the robot.}
\label{fig:ukf}
\end{figure}

We have multiple sensors on the platform, each providing partial information about the state of the robot. Moreover, the sensors provide output at different rates, for example, we run the stereo cameras at \SI{40}{\hertz} while the downward pointing distance sensor runs at \SI{20}{\hertz}. We need to merge these pieces of partial information into a single consistent estimate of the full state of the robot. The typical method used for such sensor fusion tasks is some variant of the Kalman filter. The quadrotor is a nonlinear system due to its rotational degrees of freedom. This requires the use of either an Extended Kalman filter (EKF) or an Unscented Kalman filter (UKF). The UKF has the advantage of better handling the system nonlinearities with only a small increase in computation, so we chose the UKF for our system. Fig~\ref{fig:ukf} shows the inputs and outputs of the UKF module running on the robot. The state vector used in the UKF is,
\begin{equation*}
  \bm{x} = \begin{bmatrix} \bm{p}^{\T} & \dot{\bm{p}}^{\T} & \phi & \theta & \psi & \bm{b}_a^{\T} & \bm{b}_{\omega}^{\T}\end{bmatrix}^{\T}
\end{equation*}
where $\bm{p}$ is the world-frame position of the robot, $\dot{\bm{p}}$ is the world-frame velocity, $\phi$, $\theta$ and $\psi$ are the roll, pitch and yaw respectively, $\bm{b}_a$ is the accelerometer bias while $\bm{b}_{\omega}$ is the gyroscope bias. We use the ZYX convention for representing the rotations in terms of the Euler angles $\phi$, $\theta$ and $\psi$. The Euler angle representation was chosen for representing the orientation primarily because of its simplicity. The well-known problem of gimbal lock when using Euler angles is not an issue in this case since the desired and expected roll and pitch of the robot is always less than \SI{90}{\degree}.

The UKF consists of a prediction step which uses the IMU data as the input and multiple update steps, one for each of the other sensors. The update step is performed whenever the corresponding sensor measurement arrives. The prediction step is nonlinear since the accelerometer and gyroscope measurements are in the body frame while the position and velocity in the state are in the world frame, which requires the transformation of the measured quantities from body to world frame using the estimated orientation.

Given that the state at iteration $k$, $\bm{x}_{k}$ (dimension $n$), has mean $\bar{\bm{x}}_{k}$ and covariance $\bm{P}_{k}$, we augment it with the process noise (dimension $p$) having mean $\bar{\bm{v}}_{k}$ and covariance $\bm{Q}_{k}$, creating the augmented state $\bm{x}^a_{k}$ and covariance matrix $\bm{P}^a_{k}$,
\begin{equation*}
  \bar{\bm{x}}^a_{k} = \begin{bmatrix}\bar{\bm{x}}_{k} \\ \bar{\bm{v}}_{k}\end{bmatrix}, \qquad
    \bm{P}^a_{k} = \begin{bmatrix} \bm{P}_{k} & \bm{0} \\ \bm{0} & \bm{Q}_{k}\end{bmatrix}
\end{equation*}
Then, we generate a set of sigma points by applying the Unscented transform \cite{Julier1995} to the augmented state,
\begin{equation}
\begin{aligned}
  \bm{\mathcal{X}}_{0}^{a}(k) &= \bar{\bm{x}}^a_{k} \\
  \bm{\mathcal{X}}_{i}^{a}(k) &= \bar{\bm{x}}^a_{k} + \sqrt{\left(L + \lambda\right) \bm{P}^a_{k}} \qquad i = 1, \ldots, L \\
  \bm{\mathcal{X}}_{i}^{a}(k) &= \bar{\bm{x}}^a_{k} - \sqrt{\left(L + \lambda\right) \bm{P}^a_{k}} \qquad i = L+1, \ldots, 2L
\end{aligned} \label{eq:sigma_point_generation}
\end{equation}
where $L = n + p$ is the dimension of the augmented state and $\lambda$ is a scaling parameter \cite{Wan2000}.

These sigma points are then propagated through the process model with the accelerometer and gyroscope measurements as input.
\begin{equation*}
  \bm{\mathcal{X}}_{i}^{x}(k+1 \,|\, k) = f\left(\bm{\mathcal{X}}_{i}^{x}(k), \bm{u}(k), \bm{\mathcal{X}}_{i}^{v}(k)\right)
\end{equation*}
where $\bm{\mathcal{X}}_{i}^{x}$ is the state part of the augmented state while $\bm{\mathcal{X}}_{i}^{v}$ is the process noise part.
The process model, $f\left(\bm{x}_k, \bm{u}_k, \bm{v}_k\right)$, for our system is given by
\begin{gather*}
\bm{u}_k = \left[\bm{a}_{meas}^{\T} \quad \bm{\omega}_{meas}^{\T}\right]^{\T} \qquad
\bm{v}_k = \left[\bm{v}_a^{\T} \quad \bm{v}_\omega^{\T} \quad \bm{v}_{b_a}^{\T} \quad \bm{v}_{b_\omega}^{\T}\right]^{\T} \nonumber \\
\bm{a} = \bm{a}_{meas} - \bm{b}_{a} + \bm{v}_{a} \nonumber \\
\bm{\omega} = \bm{\omega}_{meas} - \bm{b}_{\omega} + \bm{v}_{\omega} \nonumber \\
\begin{aligned}
\bm{p}_{k+1} &= \bm{p}_{k} + \dot{\bm{p}}_{k} \dif t \\
\dot{\bm{p}}_{k+1} &= \dot{\bm{p}}_{k} + \left(\bm{R}_k \bm{a} - \bm{g}\right)  \dif t\\
\bm{R}_{k+1} &= \bm{R}_{k} \left(\bm{I}_3 + \left[\bm{\omega}\right]_{\times} \dif t \right) \\
\bm{b}_{a_{k+1}} &= \bm{b}_{a_{k}} + \bm{v}_{b_a} \dif t\\
\bm{b}_{\omega_{k+1}} &= \bm{b}_{\omega_{k}} + \bm{v}_{b_\omega} \dif t
\end{aligned}
\end{gather*}
where $\bm{R}_{k} = \bm{R}\left(\phi_k, \theta_k, \psi_k \right)$ is the rotation matrix formed by using the ZYX convention for the Euler angles while $\bm{v}_a$, $\bm{v}_\omega$, $\bm{v}_{b_a}$ and $\bm{v}_{b_\omega}$ are the individual process noise terms.

From the transformed set of sigma points, $\bm{\mathcal{X}}_{i}^{x}(k+1 \,|\, k)$, we can calculate the predicted mean and covariance,
\begin{gather*}
  \bar{\bm{x}}_{k+1 \,|\, k} = \sum_{i=0}^{2L} w_{i}^{m} \bm{\mathcal{X}}_{i}^{x}(k+1 \,|\, k) \\
  \bm{P}_{k+1 \,|\, k} = \sum_{i=0}^{2L} w_i^{c} \left[\bm{\mathcal{X}}_{i}^{x}(k+1 \,|\, k) - \bar{\bm{x}}_{k+1 \,|\, k}\right] \left[\bm{\mathcal{X}}_{i}^{x}(k+1 \,|\, k) - \bar{\bm{x}}_{k+1 \,|\, k}\right]^{\T}
\end{gather*}
where $w_{i}^{m}$ and $w_i^{c}$ are scalar weights \cite{Wan2000}.

Whenever a new sensor measurement, $\bm{y}_{k+1}$, arrives, we run the update step of the filter. First we generate a new set of sigma points in the same way as done during the prediction step, \eqref{eq:sigma_point_generation}, with the augmented state and covariance given by,
\begin{equation*}
  \bar{\bm{x}}^a_{k+1 | k} = \begin{bmatrix}\bar{\bm{x}}_{k+1 \,|\, k} \\ \bar{\bm{n}}_{k}\end{bmatrix}, \qquad
    \bm{P}^a_{k+1 | k} = \begin{bmatrix} \bm{P}_{k+1 | k} & \bm{0} \\ \bm{0} & \bm{R}_{k}\end{bmatrix}
\end{equation*}
where $\bar{\bm{n}}_k$ is the mean of the measurement noise and $\bm{R}_k$ is the covariance. The generated sigma points are then used to generate the predicted measurement using the measurement function $h\left(\bm{x}, \bm{n}\right)$,
\begin{gather*}
\bm{\mathcal{Y}}_{i}(k+1 \,|\, k) = h\left(\bm{\mathcal{X}}_{i}^{x}(k+1 \,|\, k), \bm{\mathcal{X}}_{i}^{n}(k+1 \,|\, k)\right) \\
\bar{\bm{y}}_{k+1 | k} = \sum_{i=0}^{2L} w_{i}^{m} \bm{\mathcal{Y}}_{i}(k+1 \,|\, k) \\
\bm{P}_{yy} = \sum_{i=0}^{2L} w_i^{c} \left[\bm{\mathcal{Y}}_{i}(k+1 \,|\, k) - \bar{\bm{y}}_{k+1 | k}\right] \left[\bm{\mathcal{Y}}_{i}(k+1 \,|\, k) - \bar{\bm{y}}_{k+1 | k}\right]^{\T}
\end{gather*}
And finally the state is updated as follows,
\begin{gather*}
\bm{P}_{xy} = \sum_{i=0}^{2L} w_i^{c} \left[\bm{\mathcal{X}}_{i}^{x}(k+1 \,|\, k) - \bar{\bm{x}}_{k+1 \,|\, k}\right] \left[\bm{\mathcal{Y}}_{i}(k+1 \,|\, k) - \bar{\bm{y}}_{k+1|k}\right]^{\T} \\
\bm{K} = \bm{P}_{xy} \bm{P}_{yy}^{-1} \\
\bar{\bm{x}}_{k+1} = \bar{\bm{x}}_{k} + \bm{K} \left(\bm{y}_{k+1} - \bar{\bm{y}}_{k+1 | k}\right) \\
\bm{P}_{k+1} = \bm{P}_{k+1 \,|\, k} - \bm{K} \bm{P}_{yy} \bm{K}^{\T}
\end{gather*}

Note that for each sensor input to the UKF except the IMU, which is used for the prediction step, there is a separate measurement function, $h\left(\bm{x}, \bm{n}\right)$, and the full update step is performed, with the corresponding measurement function, when an input is received from any of those sensors.

The attitude filter running on the Pixhawk is a simple complementary filter which can take an external reference orientation as an input. This allows us to provide the estimate from the UKF to the Pixhawk in order to improve the orientation estimate on the Pixhawk. This is important for good control performance since the orientation controller running on the Pixhawk uses the Pixhawk's estimate of the orientation while our control commands are calculated using the UKF estimates. Without an external reference being sent to the Pixhawk, the orientation estimates on the Pixhawk can be different from the UKF which would lead to an incorrect interpretation of the control commands.

\subsection{Control}
The controller used for the robot has the cascade structure, as shown in Figure~\ref{fig:software_architecture}, which is has become standard for MAVs. In this structure, we have an inner loop controlling the orientation and angular velocities of the robot while an outer loop controls the position and linear velocities. In our case, the inner loop runs at a high rate (\SI{400}{\hertz}) on the Pixhawk autopilot while the outer loop runs at a slightly slower rate (\SI{200}{\hertz}) on the Intel NUC computer.

%\begin{figure}[ht]
%\centering
%\missingfigure{Control}
%\caption{Block diagram of the controller consisting of an inner loop controlling the orientation running on the Pixhawk and an outer loop controlling the position running on the Intel NUC computer.}
%\label{fig:controller}
%\end{figure}

At every time instance, the outer loop position controller receives a desired state, which consists of a desired position, velocity, acceleration and jerk, from the planner and using the estimated state from the UKF, computes a desired force, orientation and angular velocities which are sent to the orientation controller. The inner loop orientation controller receives these and computes the thrust and moments required to achieve the desired force and orientation. These are then converted into individual motor speeds that are sent to the respective motor controllers.

The controller formulation we use is based on the controller developed in \cite{Lee2010} with some simplifications. The thrust command of the position controller is calculated as,
\begin{gather}
  \bm{e}_{pos} =  \hat{\bm{p}} - \bm{p}_{des} \, , \quad \bm{e}_{vel} = \hat{\dot{\bm{p}}} - \dot{\bm{p}}_{des} \nonumber \\
  \bm{f} = m \left(-k_{pos} \bm{e}_{pos} - k_{vel} \bm{e}_{vel} + g\bm{e}_{3} + \ddot{\bm{p}}_{des}\right) \nonumber \\
  \mathrm{Thrust} = \bm{f} \cdot \hat{\bm{R}} \bm{e}_{3} \label{eq:thrust_des}
\end{gather}
where $\bm{e}_{3} = \left[\begin{smallmatrix}0 & 0 & 1\end{smallmatrix}\right]^{\T}$ and $\hat{\bm{R}}$ is the rotation matrix which converts vectors from body frame to world frame calculated using the estimated roll, pitch and yaw. The desired attitude is calculated as,
\begin{gather}
  \bm{b}_{2,des} = \begin{bmatrix}-\sin\psi_{des}, & \cos\psi_{des}, & 0\end{bmatrix}^{\T} \nonumber \\
  \bm{b}_{3} = \frac{\bm{f}}{\norm{\bm{f}}}, \quad
  \bm{b}_{1} = \frac{\bm{b}_{2,des} \times \bm{b}_{3}}{\norm{\bm{b}_{2,des} \times \bm{b}_{3}}}, \quad
  \bm{b}_{2} = \bm{b}_{3} \times \bm{b}_{1} \nonumber \\
  \bm{R}_{des} = \begin{bmatrix}\bm{b}_{1}, & \bm{b}_{2}, & \bm{b}_{3} \end{bmatrix} \label{eq:R_des} \\
  \dot{\bm{b}}_{2,des} = \begin{bmatrix}-\cos\psi_{des} \dot{\psi}_{des}, & -\sin\psi_{des} \dot{\psi}_{des}, & 0\end{bmatrix}^{\T} \nonumber \\
  \dot{\bm{b}}_3 = \bm{b}_{3} \times \frac{\dot{\bm{f}}}{\norm{\bm{f}}} \times \bm{b}_{3}, \quad
  \dot{\bm{b}}_1 = \bm{b}_{1} \times \frac{\dot{\bm{b}}_{2,des} \times \bm{b}_{3} + \bm{b}_{2,des} \times \dot{\bm{b}}_{3}}{\norm{\bm{b}_{2,des} \times \bm{b}_{3}}} \times \bm{b}_{1}, \quad
  \dot{\bm{b}}_2 = \dot{\bm{b}}_3 \times \bm{b}_1 + \bm{b}_3 \times \dot{\bm{b}}_1 \nonumber \\
  \left[\bm{\Omega}_{des}\right]_{\times} = \bm{R}_{des}^{\T} \dot{\bm{R}}_{des} \label{eq:omega_des}
\end{gather}
Note that here we have to define $\bm{b}_{2,des}$ based on the yaw instead of defining $\bm{b}_{1,des}$ as done in \cite{Mellinger2011} due to the different Euler angle convention, we use the ZYX convention while they used ZXY.

The thrust and attitude commands, from \eqref{eq:thrust_des}, \eqref{eq:R_des} and \eqref{eq:omega_des}, are then sent to the Pixhawk autopilot through mavros\footnote{\url{https://github.com/mavlink/mavros}}. The attitude controller running on the Pixhawk takes these commands and converts them to commanded motor speeds. First, using the $\bm{R}_{des}$ and the estimate of the current orientation, $ \hat{\bm{R}}$, we calculate the desired moments as follows,
\begin{gather*}
  \left[\bm{e}_{R}\right]_{\times} = \frac{1}{2}\left(\bm{R}_{des}^{\T} \,  \hat{\bm{R}} -  \hat{\bm{R}}^{\T} \bm{R}_{des}\right) \, , \quad
  \bm{e}_{\Omega} = \bm{\Omega} - \hat{\bm{R}}^{\T} \bm{R}_{des}\bm{\Omega}_{des} \\
  \bm{M} = -k_{R} \bm{e}_{R} - k_{\Omega} \bm{e}_{\Omega}
\end{gather*}
Then, from the desired thrust and moments, we can calculate the thrust required from each propeller which allows us to compute the desired motor speed as shown in \cite{Michael2010}.

\section{Mapping and Planning} \label{sec:planning}
\begin{figure}[ht]
\centering
\includegraphics[width=0.6\columnwidth]{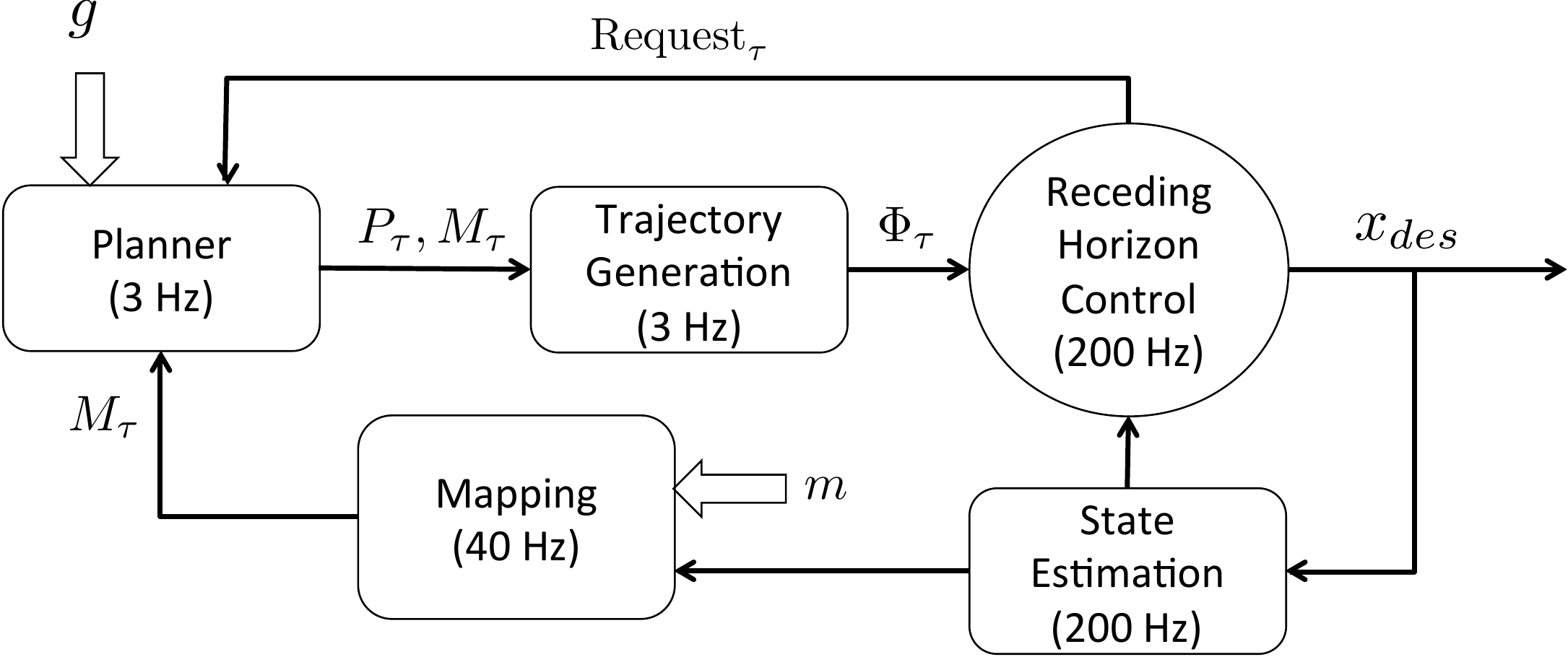}
\caption{Our navigation framework. A desired goal $g$ is sent to the planner at the beginning of the task. The planner generates a path, $P_\tau$, using the map, $M_\tau$, and sends it to the trajectory generator. The trajectory generator converts the path into a trajectory, $\Phi_\tau$, and sends it to the receding horizon controller. The controller then derives the desired state $x_{\text{des}}$ at \SI{200}{\hertz} from this trajectory which is sent to the robot controller. The input $m$ to the mapping block denotes the sensor measurements.}
\label{fig: planning_diagram}
\end{figure}

Our navigation system consists of five parts as shown in Fig.~\ref{fig: planning_diagram}. In this section, we discuss the mapping, planner and trajectory generation threads.

\subsection{Mapping}
\begin{figure}[ht]
  \centering
  \subfigure[Local point cloud $M^c$ ($\SI{40}{m}\times\SI{20}{m}\times\SI{4}{m}$).]{ \includegraphics[width=0.3\columnwidth]{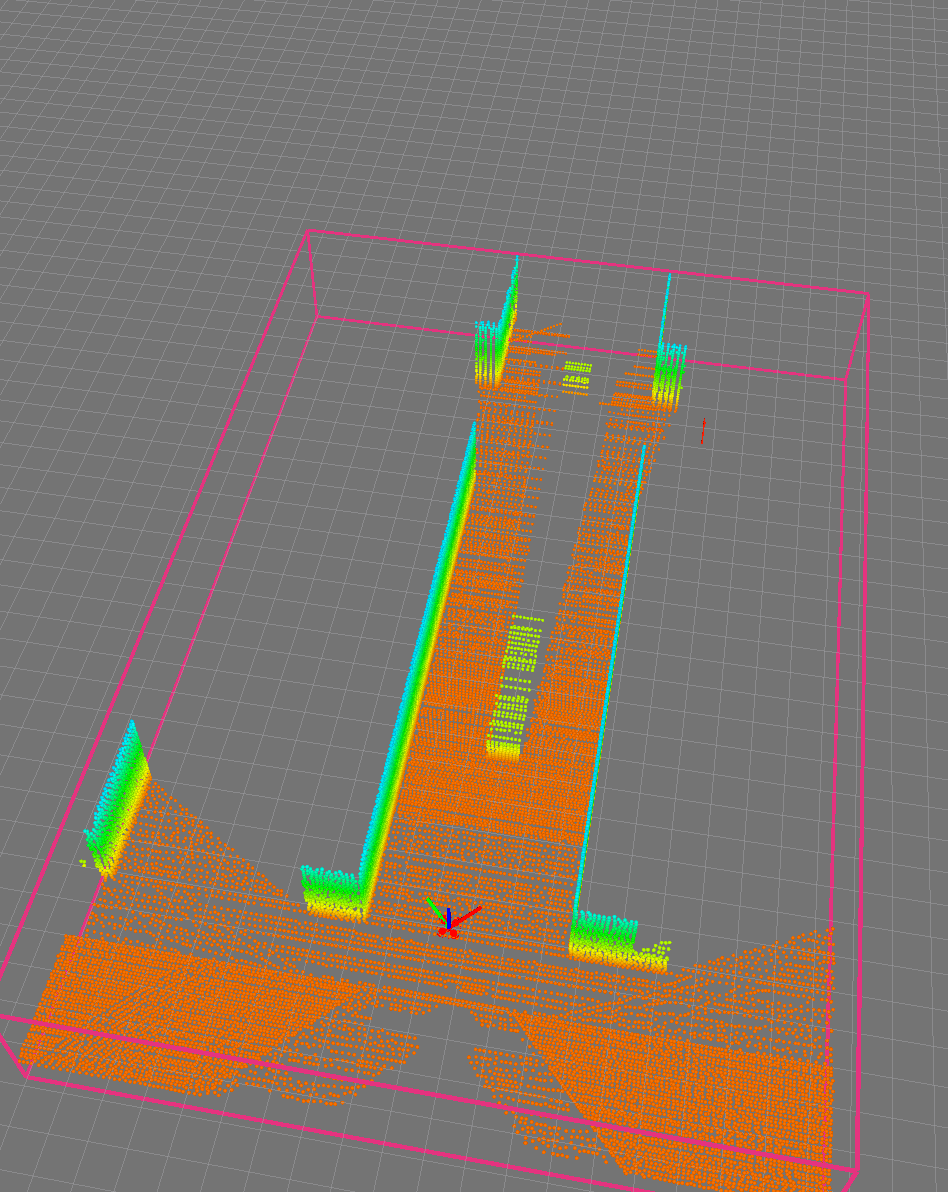}}
  \subfigure[Local map $M^l$ ($\SI{15}{m}\times\SI{10}{m}\times\SI{3}{m}$) and global map $M^w$($\SI{80}{m} \times \SI{40}{m}$).]{\includegraphics[width=0.3\columnwidth]{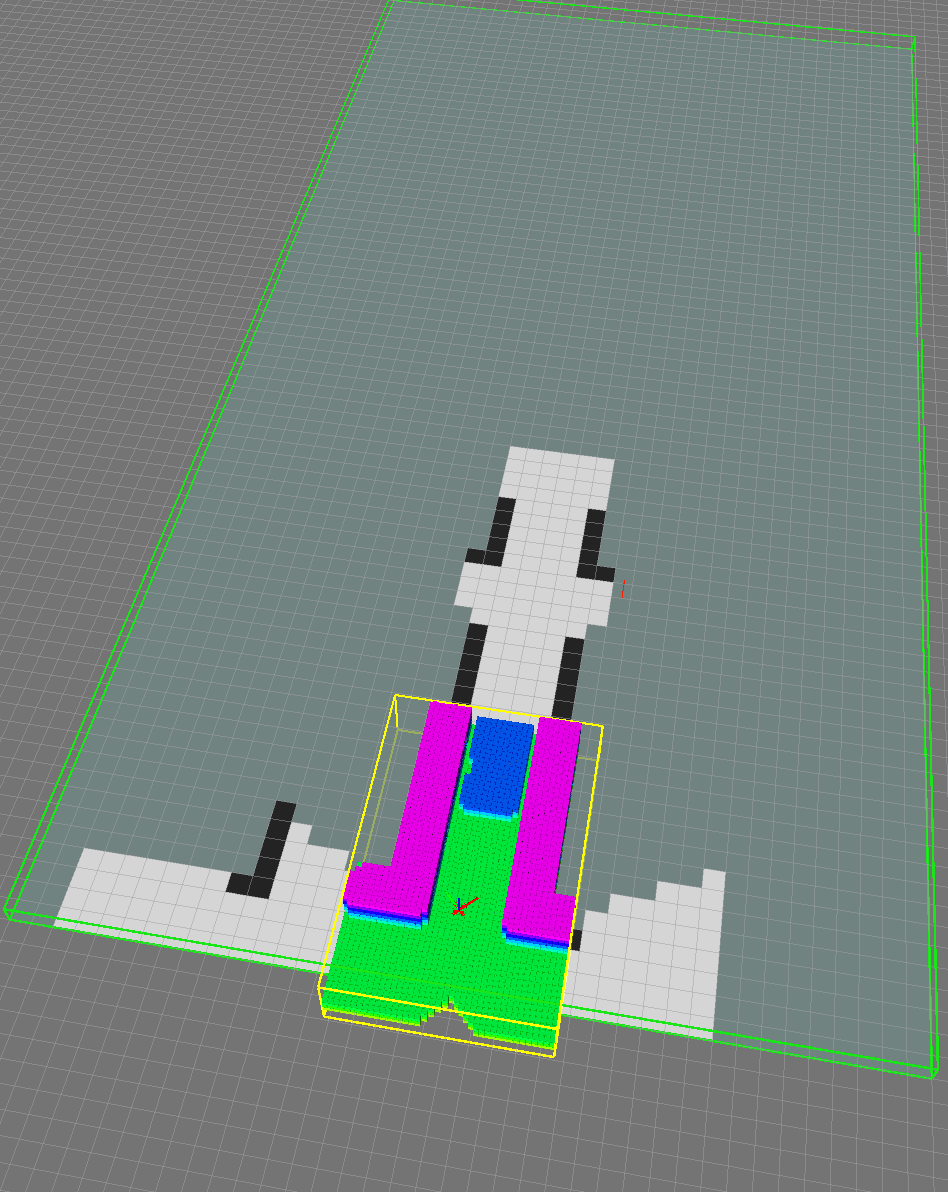}}
  \subfigure[Path planned using both $M^l, M^w$.]{ \includegraphics[width=0.3\columnwidth]{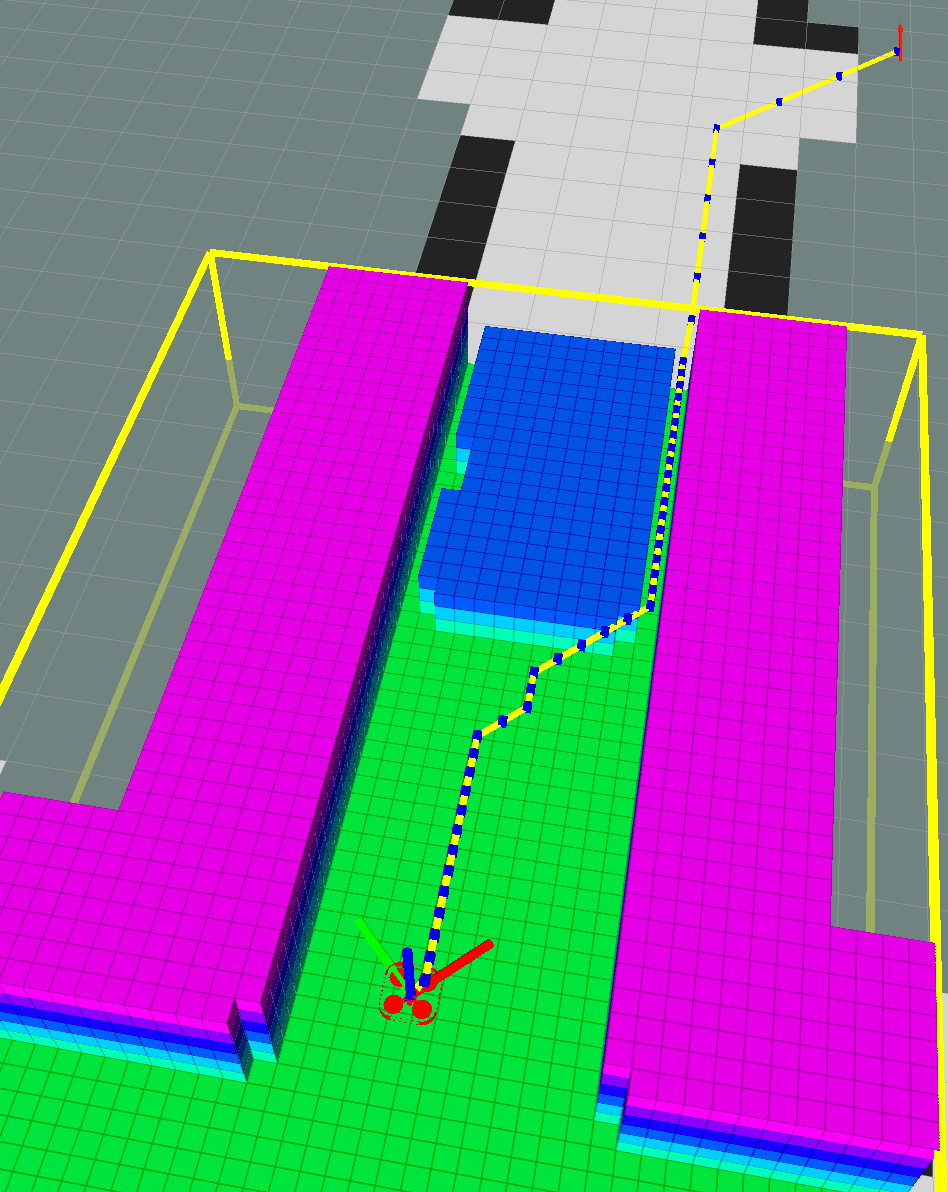}}
  \caption{We keep the range of the local point cloud equal to the sensor range (e.g \SI{30}{m} for a laser rangefinder). The size of local map $M^l$ is smaller than the point cloud $M^c$ because of the computational limitation. For planning, we dilate the occupied voxel in $M^l$ by the robot radius. The global information map is much larger but with much coarser resolution (\SI{1}{m}). For each map, we draw a bounding box to visualize the size. \label{fig: maps}}
\end{figure}

We have mounted a LIDAR on a servo such that we can generate a 3D voxel map by rotating the laser. Updating the map and planning using the 3D global map are both computationally expensive and in addition, with noise and estimation drift, the global map can be erroneous. Hence we utilize a local mapping technique that generates a point cloud around current robot location (Fig.~\ref{fig: maps}). This local point cloud, $M^c$, has fixed size and fine resolution and is used to build a 3D occupancy voxel map, $M^l$, centered at current robot location. Since the local map only records the recent sensor measurements with respect to current robot location, the accumulated error in mapping is small. We also generate a coarse 2D map, $M^w$, in global frame in order to solve the dead-end problem caused by local planning. We call this map the ``global information map'' since it contains two pieces of information: one is the known and unknown spaces so that we know which part has been explored, the other is the location of walls detected from $M^c$ that the robot cannot fly over.

\subsection{Planner}
\begin{figure}[ht]
  \centering
  \subfigure[Local map $M^l_{\tau_1}$]{ \includegraphics[width=0.23\columnwidth, trim=0 0 0 0, clip=true]{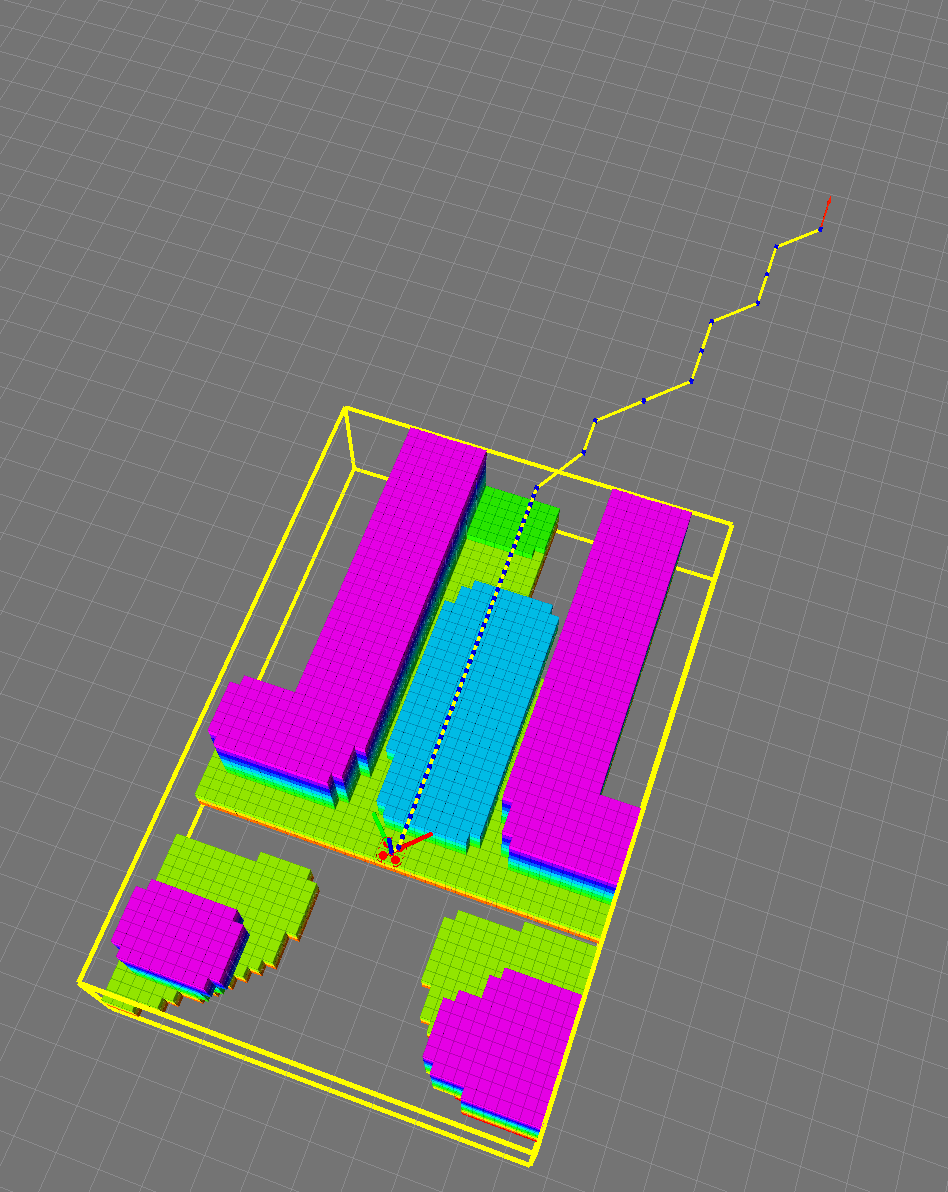}}
  \subfigure[Global map $M^w_{\tau_1}$]{ \includegraphics[width=0.23\columnwidth, trim=0 0 0 0, clip=true]{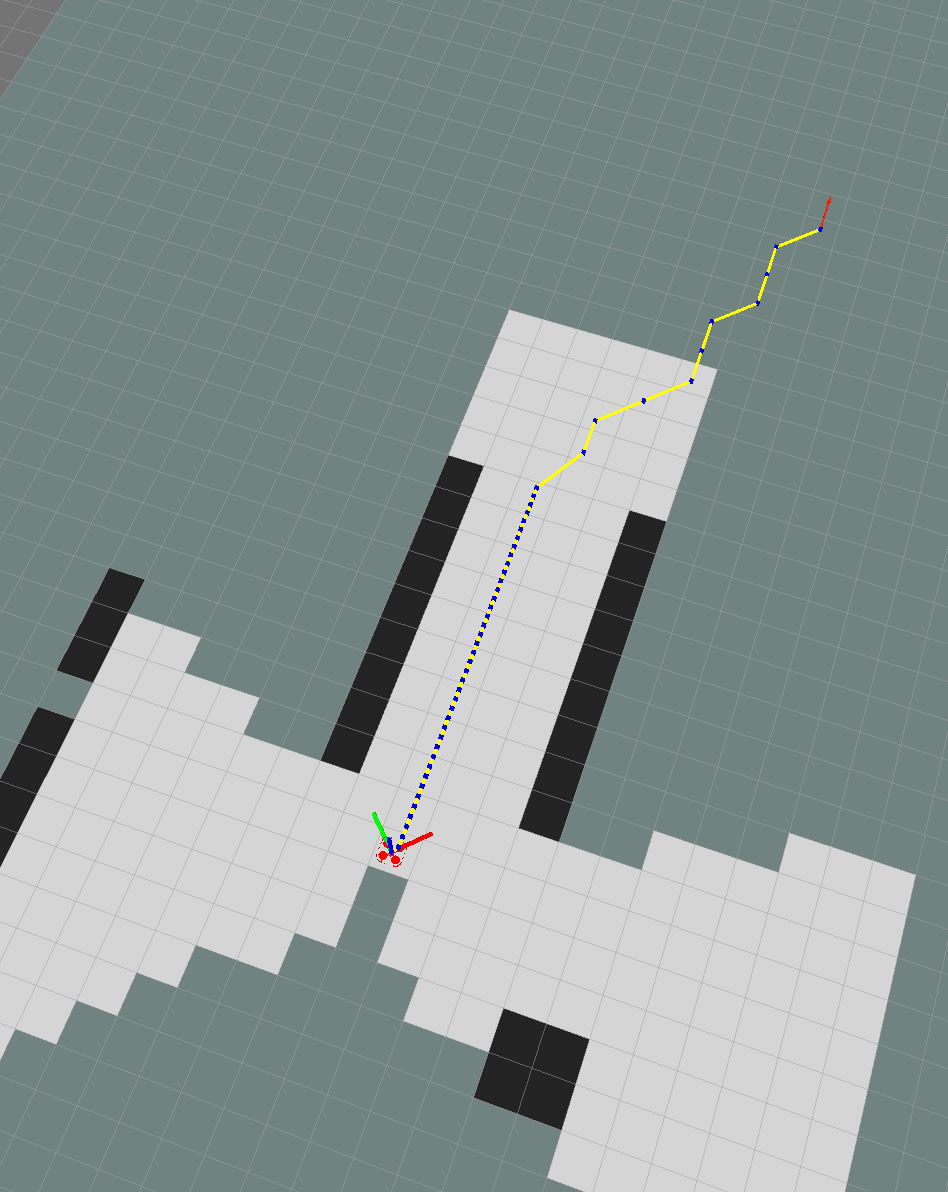}}
  \subfigure[Local map $M^l_{\tau_2}$]{ \includegraphics[width=0.23\columnwidth, trim=0 0 0 0, clip=true]{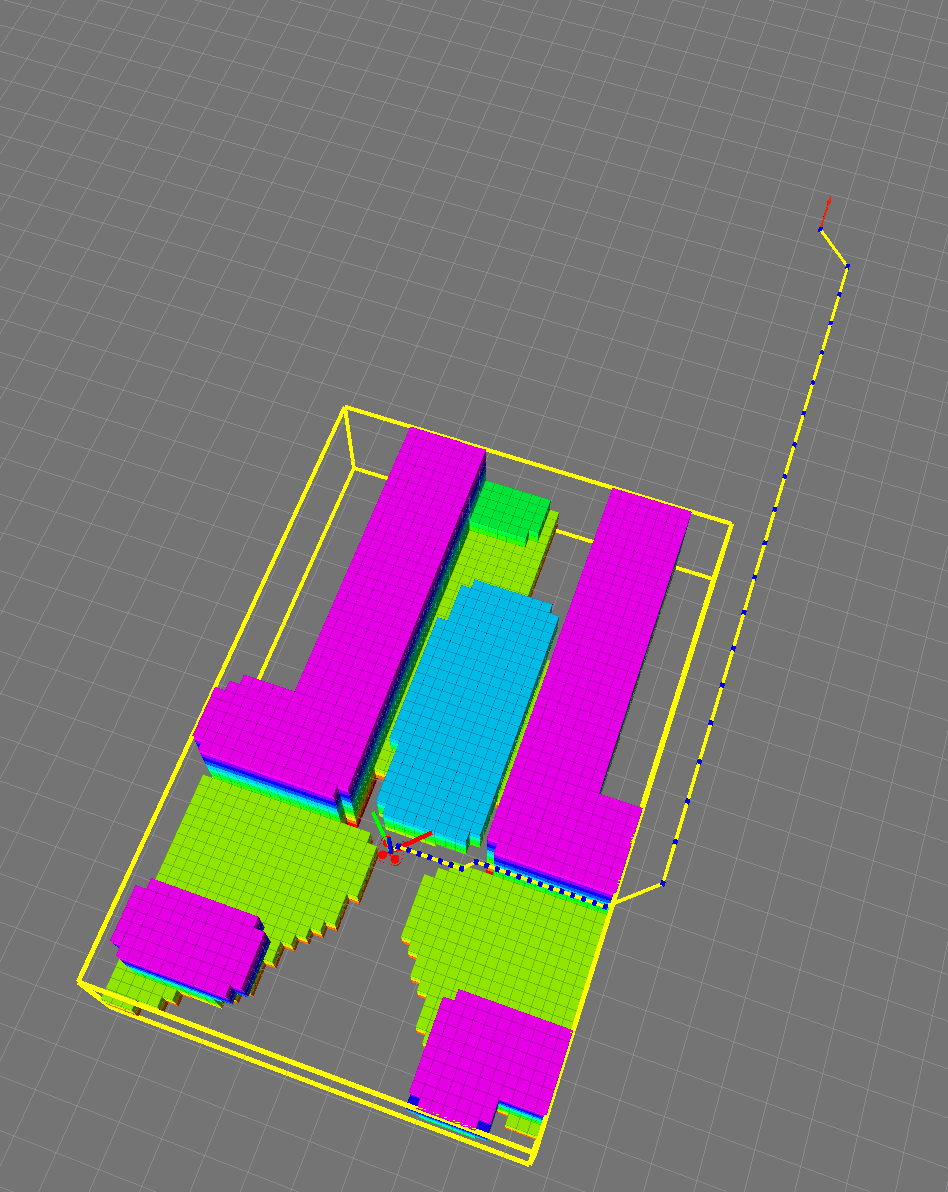}}
  \subfigure[Global map $M^w_{\tau_2}$]{ \includegraphics[width=0.23\columnwidth, trim=0 0 0 0, clip=true]{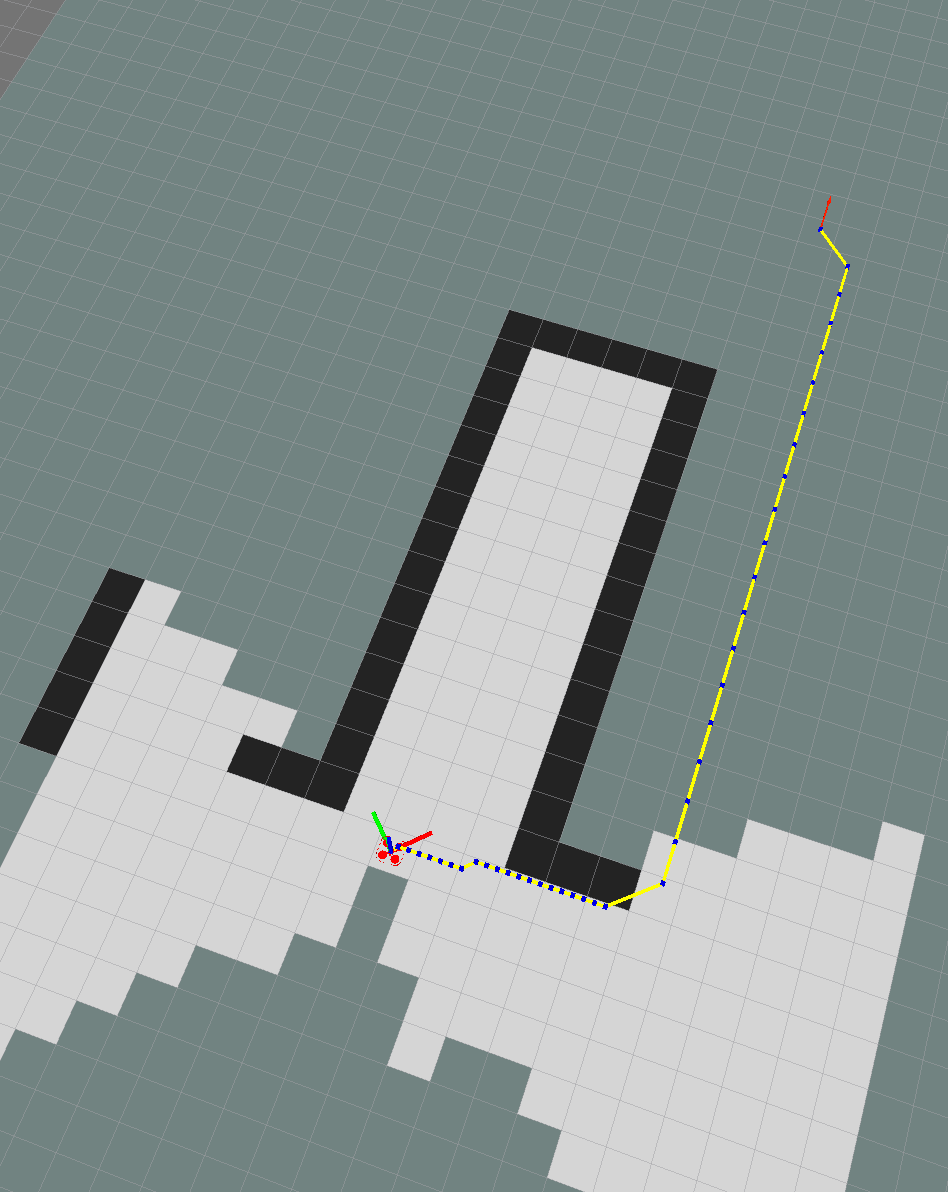}}
  \caption{At planning epoch $\tau_1$, the end of the corridor cannot be viewed by the sensor with limited range and the path leads the quadrotor to go forward (a)-(b). At planning epoch $\tau_2$, with similar local map $M^l_{\tau_2}$ but different global map $M^w_{\tau_2}$ which contains the dead-end geometry, planning to the same goal, results in a path which is totally different. \label{fig: paths}}
\end{figure}

We use $A^\star$ to plan a path in a hybrid graph $G(V, E)$ that links the voxels in both local 3D map $M^l$ and global information map $M^w$ (result is shown in Fig.~\ref{fig: maps}(c)). We can efficiently derive the path $P$ in local map that is globally consistent. Fig.~\ref{fig: paths} shows an example of using this method to solve the dead-end corridor problem.

\subsection{Trajectory Generation}
\begin{figure}[ht]
	\centering
	\includegraphics[width=0.7\columnwidth]{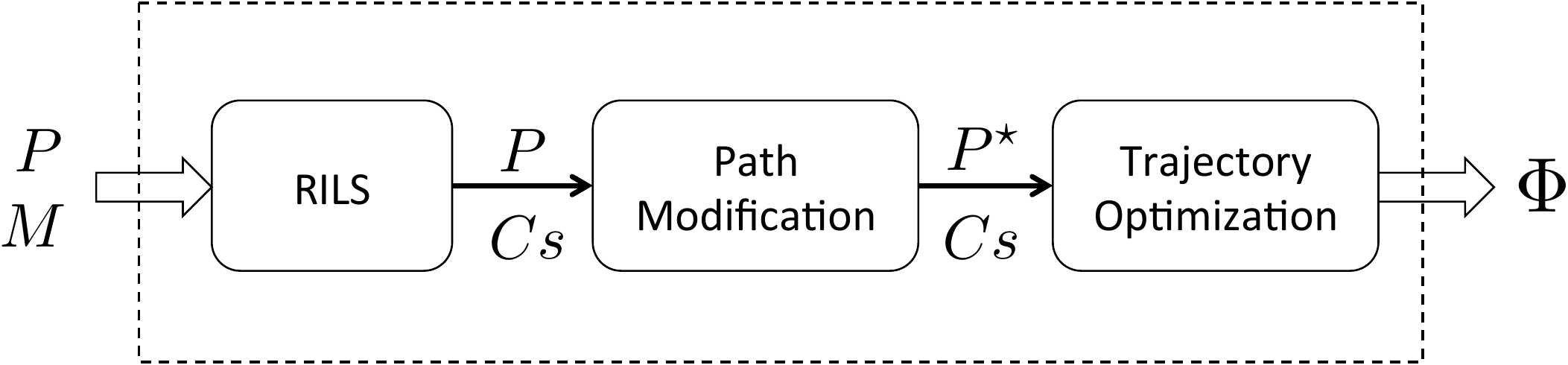}
	\caption{Trajectory generation process, which can be treated as a black box (dashed rectangle) as in Fig.~\ref{fig: planning_diagram}. The inputs are a path $P$ and a discrete map $M$, output is the dynamically feasible trajectory $\Phi$. \label{fig: traj_pip}}
\end{figure}

In this subsection, we are going to introduce the trajectory generation method given the map $M$ and a prior path $P$. The trajectory generation process is shown in Fig.~\ref{fig: traj_pip}. Through regional inflation, a safe corridor is found in $M$ that excludes all the obstacle points. As the intermediate waypoints in $P$ can be close to the obstacles, we shift the intermediate waypoints towards the center of safe corridor. The new path $P^\star$ and the safe corridor are used to generate the trajectory.

\subsubsection{Regional Inflation by Line Segments} \label{sec:rils}
\begin{figure}[ht]
  \centering
  \subfigure[Grow ellipsoid for each line segment.]{ \includegraphics[width=0.3\columnwidth, trim=0 0 0 0, clip=true]{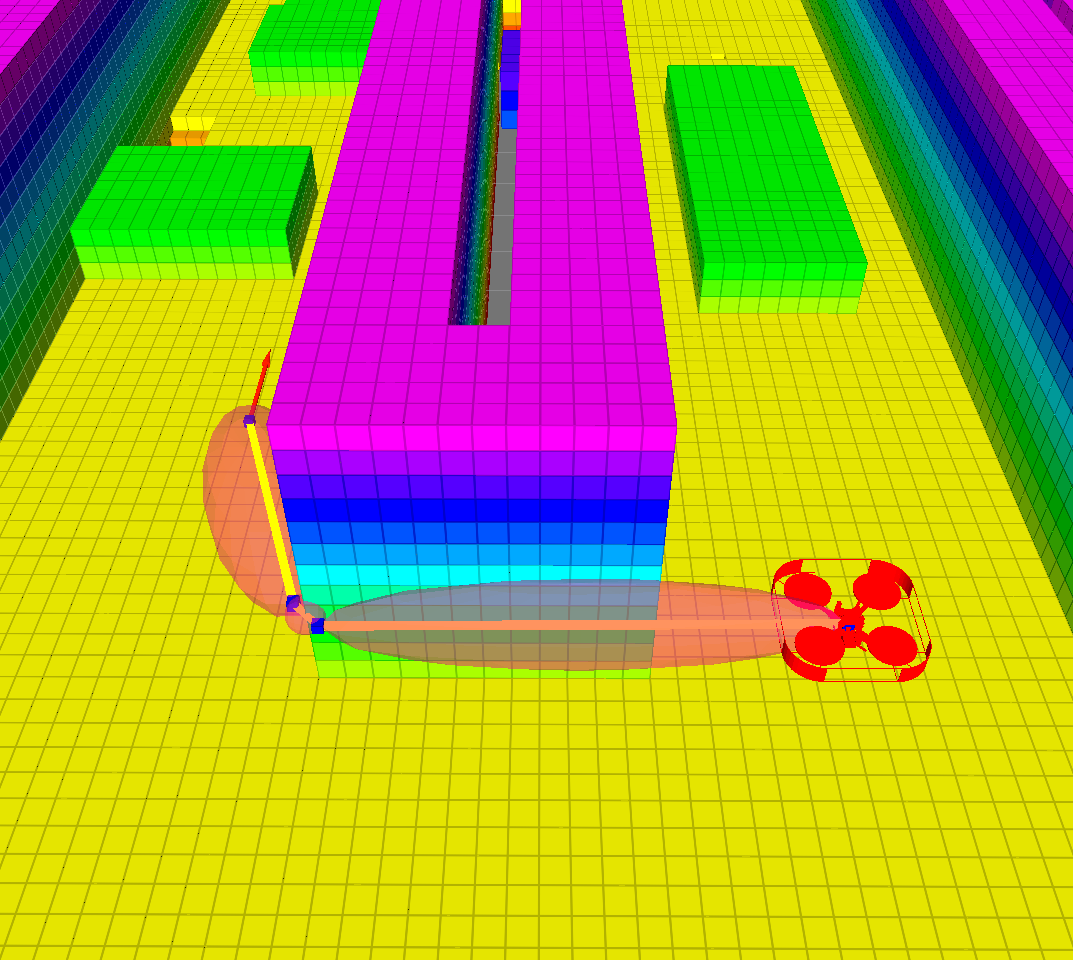}}
  \subfigure[Inflate the ellipsoid to generate the convex polyhedra.]{ \includegraphics[width=0.3\columnwidth, trim=0 0 0 0, clip=true]{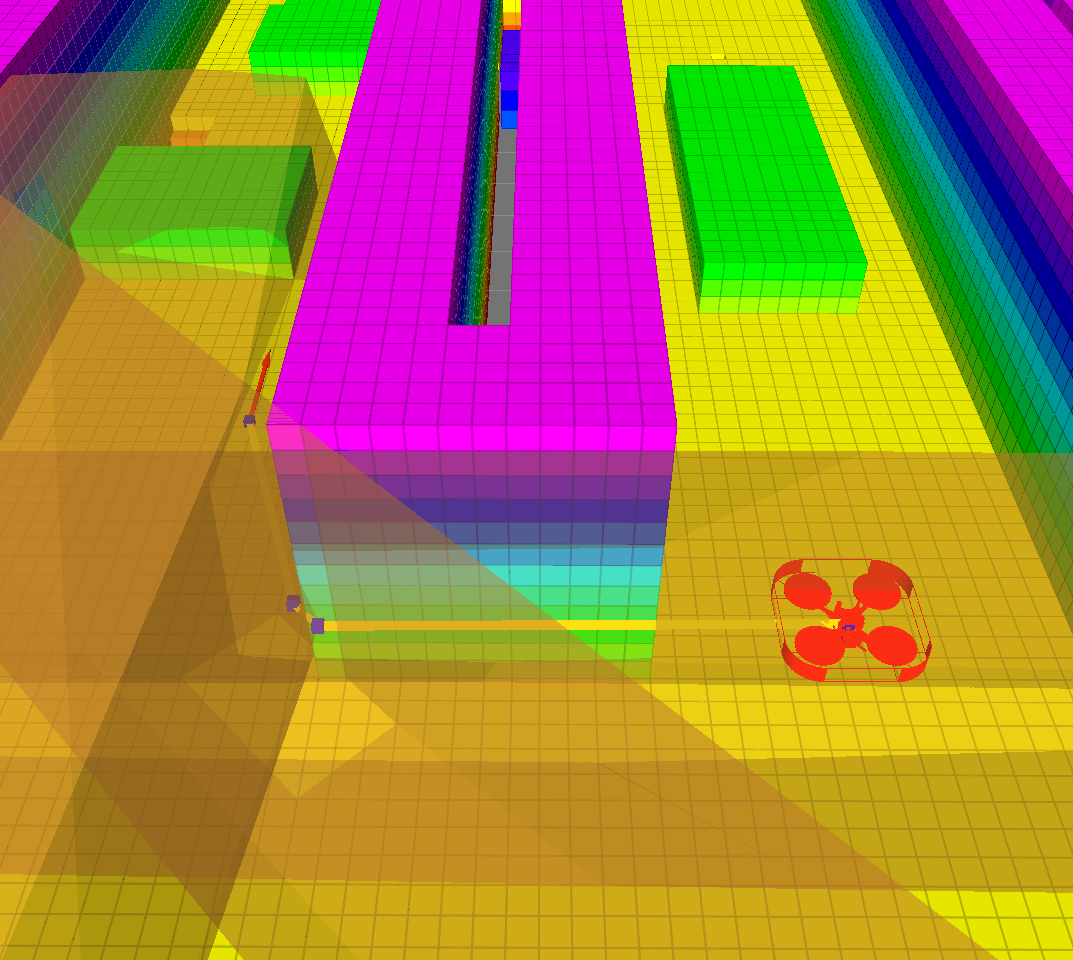}}
  \subfigure[Modified path $P^\star$.]{ \includegraphics[width=0.3\columnwidth, trim=0 0 0 0, clip=true]{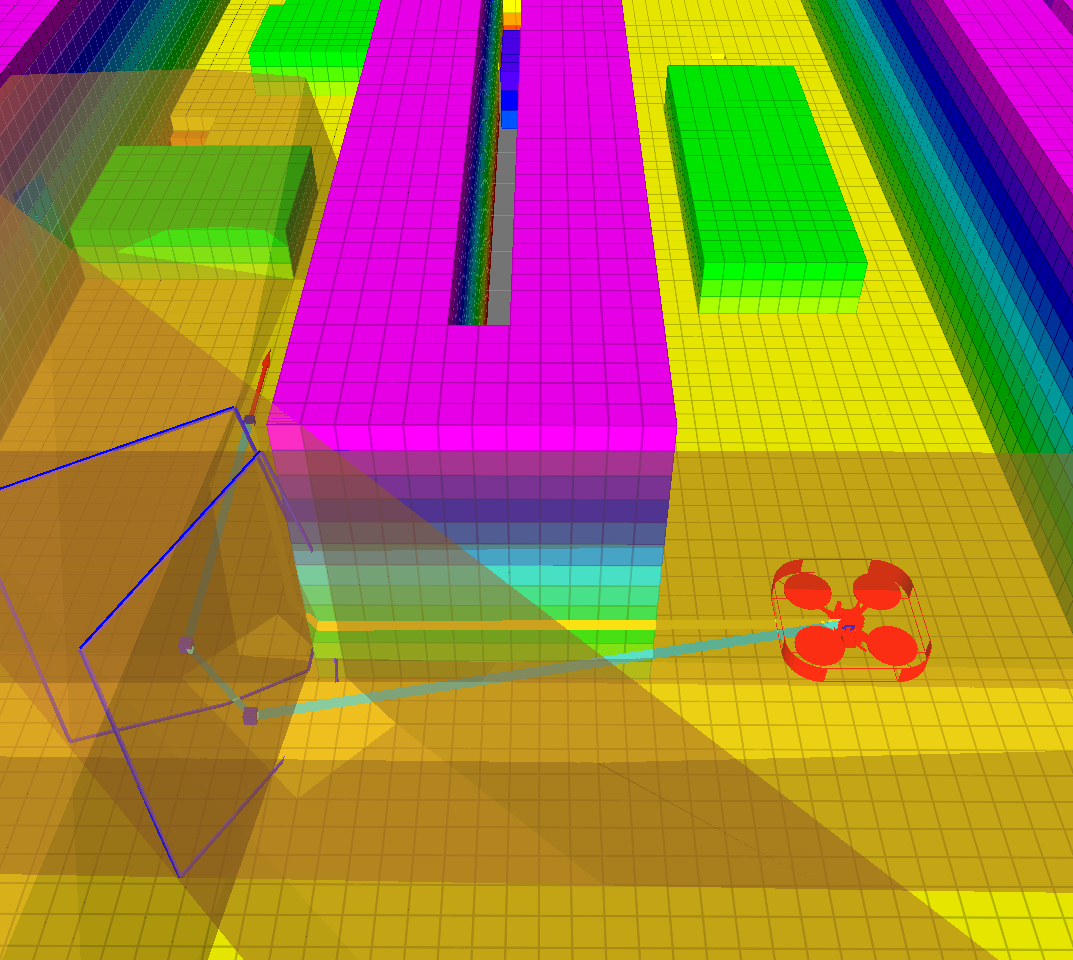}}
     \caption{We generate the safe corridor by inflating the free region around path using RILS. The ellipsoid is colored as purple in (a), the transparent orange region on (b) shows the polyhedra for safe corridor. The cyan path $P^\star$ is modified from the original path $P$ by shifting the corner to the centroid of blue polygon in (c). \label{fig: rils}}
\end{figure}

Inspired by IRIS in~\cite{deits2015computing}, we developed the algorithm to dilate a path in free space using ellipsoids. For each line segment in the path $P$, we generate a convex polyhedron that includes the whole segment but excludes any occupied voxel in $M$ through two steps:
\begin{itemize}
  \item Grow ellipsoid for each line segment (Fig.~\ref{fig: rils}(a))
  \item Inflate the ellipsoid to generate the polyhedron (Fig.~\ref{fig: rils}(b))
\end{itemize}

The ellipsoid is described as $\xi_r(E, d) = \{E\tilde{x} + d \ |\  \|\tilde{x}\| \leq r\}$ which is the projection of a unit sphere with radius $r$ into $\mathbb{R}^3$. A polyhedron $C$ is the intersection of $m$ half-planes: $C = \{x \ | \ A_j^Tx \leq b_j,\ j = 1\hdots m\}$. The half-plane is found at the intersection of ellipsoid with the closest obstacle point $x_r$ as shown in equation~\eqref{eq: inter}.

\begin{equation}\label{eq: inter}
     A_j = \left.\frac{\dif\xi_{r}}{\dif x}\right\vert_{x = x_r}  = 2E^{-1}E^{-T}(x_r - d), \quad b_j =  A_j^T x_r
\end{equation}

\subsubsection{Path Modification}
The original path from the planner can be close to the obstacles. Although the trajectory generation does not require the final trajectory to go through the intermediate points, the path affects the route of the trajectory. The path modification step aims to modify the original path away from the obstacle by keeping the intermediate waypoints in the middle of a safe corridor. We use a bisector plane that passes through the waypoint $p$, this plane intersects with both polyhedra that connected through $p$. The point $p$ is moved to the centroid of the intersection polygon (Fig.~\ref{fig: rils}(c)).

\subsubsection{Trajectory Optimization}
\begin{figure}[ht]
  \centering
  \subfigure[$\epsilon = 0$]{ \includegraphics[width=0.3\columnwidth, trim=0 0 0 0, clip=true]{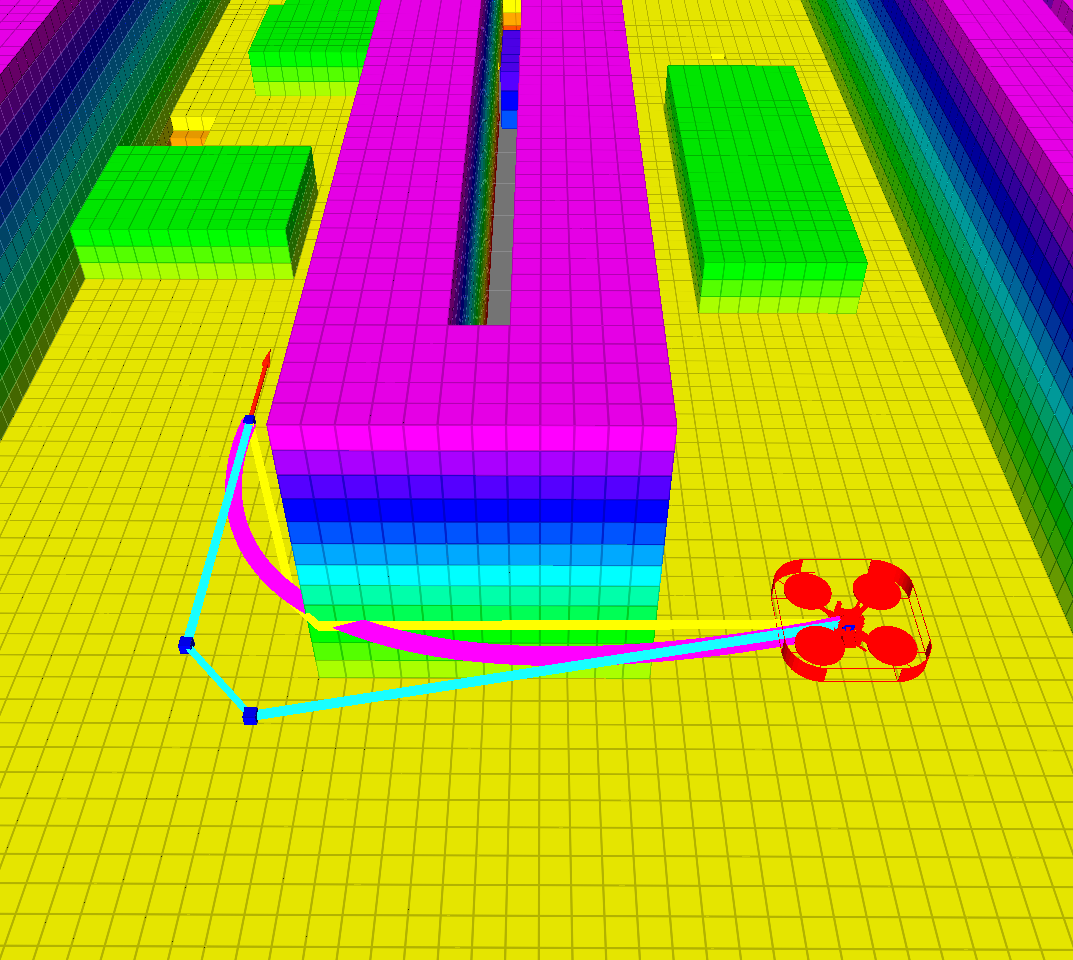}}
  \subfigure[$\epsilon = 20$]{ \includegraphics[width=0.3\columnwidth, trim=0 0 0 0, clip=true]{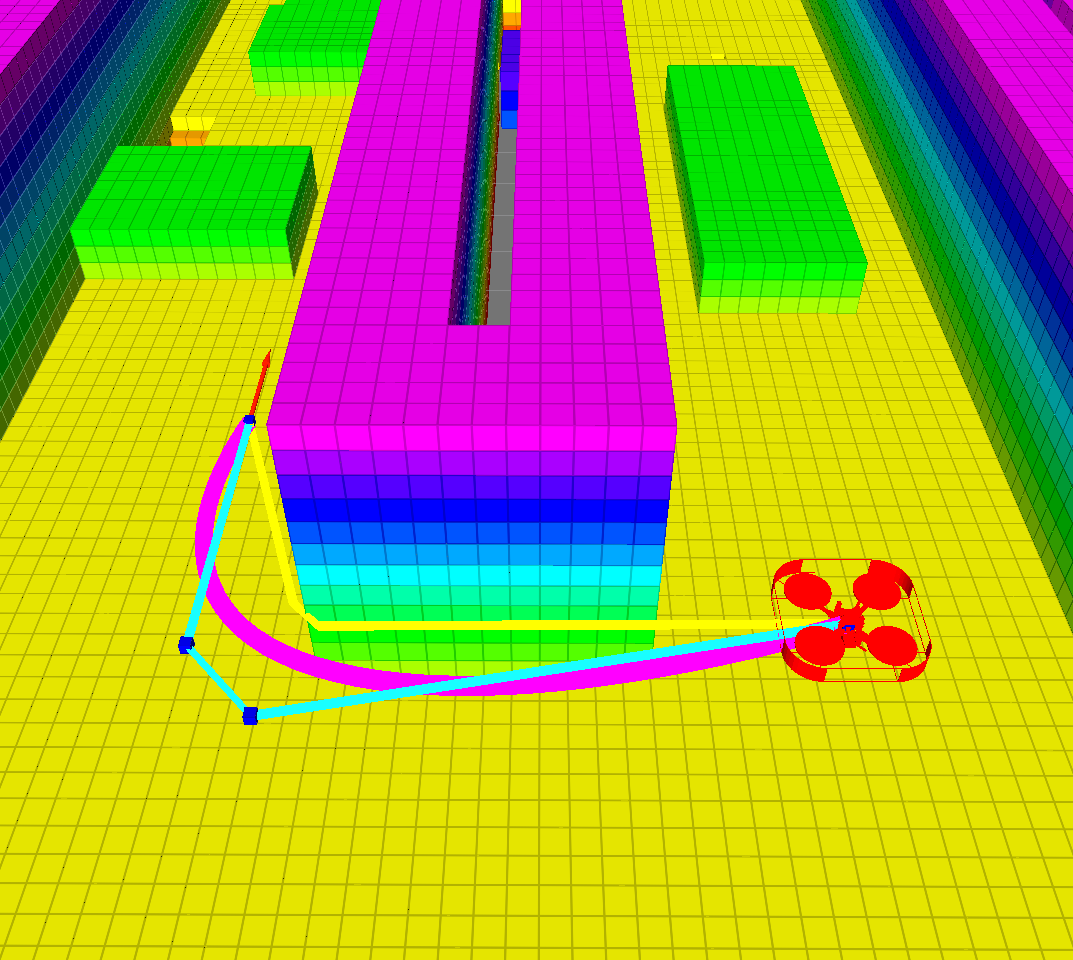}}
  \subfigure[$\epsilon = 100$]{ \includegraphics[width=0.3\columnwidth, trim=0 0 0 0, clip=true]{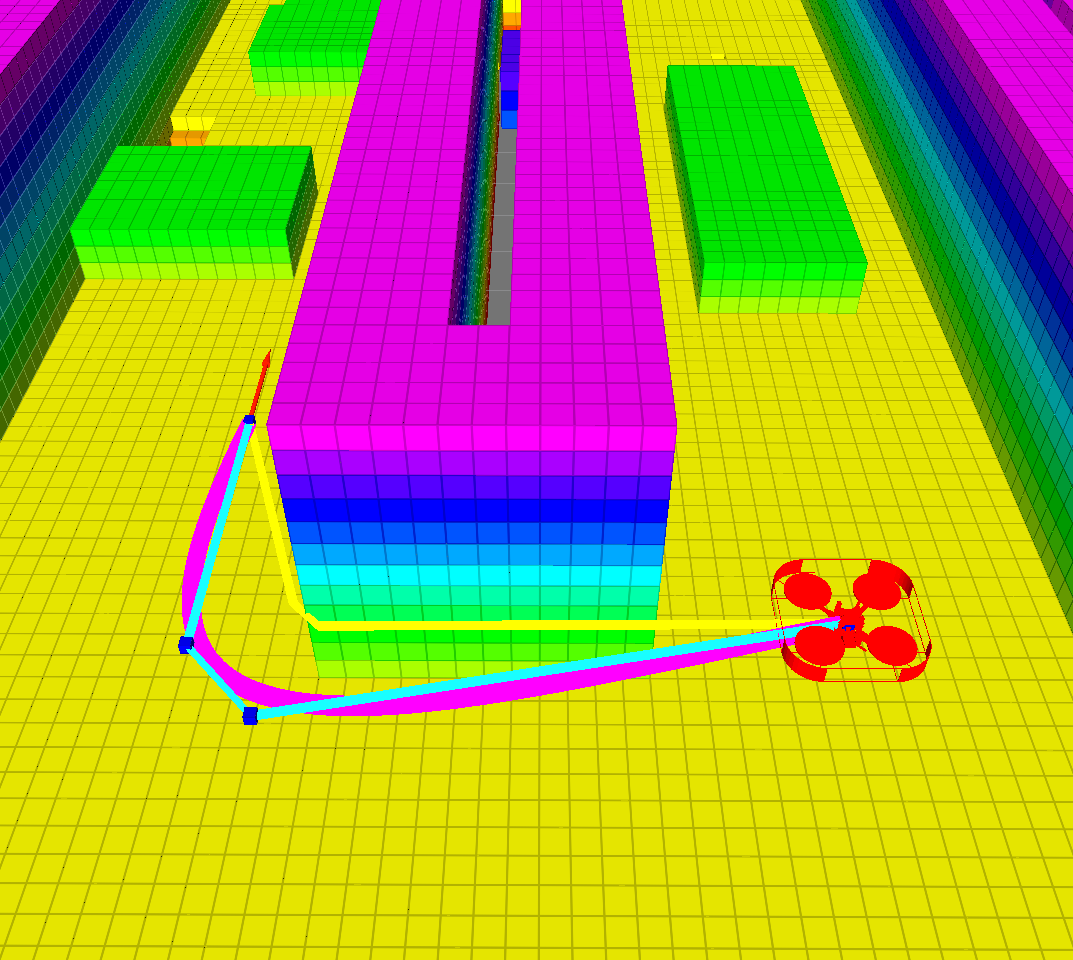}}
     \caption{The generated trajectories (purple) for different values of the weight $\epsilon$. As we increase the weight, the trajectory gets closer to the given path (cyan). \label{fig: traj_gen}}
\end{figure}

We formulate the trajectory generation as an optimal control problem as shown in equation~\eqref{eq:traj_min}. Compared to the standard formulation of this optimal control problem, we add a second term in the cost function which is the square of the distance between the trajectory $\Phi_i$ and the line segment $l_i: \{a_ix + b_iy+c_iz + d_i = 0\}$. This distance cost is weighted by a factor $\epsilon$. Fig.~\ref{fig: traj_gen} shows the affect on the trajectory by changing this weighting factor. Thus, we can control the shape of a trajectory to keep it close to the modified path $P^\star$ and away from obstacles. This process increases the safety of the trajectory such that the robot will not get too close to obstacles.

\begin{equation} \label{eq:traj_min}
\begin{gathered}
		\arg\min J =  \bigintsss_0^{T} \norm{\frac{\dif^{\,4} \Phi(t)}{\dif t^4}}^2 \dif t + \epsilon\left(\frac{l\cdot\Phi(t)}{\norm{l}}\right)^2  \\
       \begin{aligned}
		s.t. \quad \Phi(0) &= p_0, \quad \Phi(T) = p_f, \quad \Phi(t) \in C_3 \\
		\dot{\Phi}(t) &\leq v_{max}, \quad \ddot{\Phi}(t) \leq a_{max}, \quad \dddot{\Phi}(t) \leq j_{max}
	\end{aligned}
\end{gathered}
\end{equation}

\subsubsection{Continuous Optimization}

To to be able to compute Equation \ref{eq:traj_min} in real time, we have chosen to represent $\Phi$ as an $n^{th}$ order polynomial spline. Each spline segment $\Phi_j$ takes time $\Delta_j$ such that $\sum_j \Delta_j=T$. For our experiments, we have found that $n=7$ provides good performance for trajectory optimization and that increasing $n$ produces similar trajectories, but with longer computational time.

For this section, we will assume that our trajectory has $d$ dimensions, $m$ segments and is represented by an $n$th order polynomial.  Also we define the indices: $i = 0,\dotsc,n$, $j=1,\dotsc,m$,and $k=1,\dotsc,d$. Therefore we can write the $k$th dimension of $\Phi_i$ with respect to coefficients $\alpha_{ijk}$ and basis functions $p_{i}$.
\begin{equation}
\Phi_{jk}(t) = \sum\limits_{i} \alpha_{ijk} p_{i}(t)
\end{equation}
We have chosen $p_{i}$ to be shifted Legendre polynomials which are represented with a unit time scaling:
\begin{equation}
	s = \frac{t - \sum\limits_{j=0}^q \Delta_j}{\Delta_{q+1}}
\end{equation}
For whichever $q \in 1,\dotsc,m$ makes $s\in[0,1]$

\begin{equation}
  \frac{\dif^{\,4}}{\dif s^k}p_{i+4}(t) = (-1)^i \sum\limits_{l=0}^i \binom{l}{i} \binom{l+i}{i}(-s)^l
\end{equation}
This leaves $4$ basis polynomials undefined, which we just use:
\begin{equation}
p_i(t) = s^i \hspace{2mm} \text{for } i=0,\dotsc,3
\end{equation}
This definition results in a cost which is quadratic in $\alpha$
\begin{equation} \label{eqn:quad_cost}
  \bigintsss_0^T \norm{\frac{\dif^{\,4} \Phi(t)}{\dif t^4}}^2 \dif t = \sum_{ijk} \alpha_{ijk}^2 \frac{\Delta_j^{-7}}{2k+1}
\end{equation}
For the other constraints are concerned with constraining $\Phi$ at different times. We note that the evaluation of any derivative of $\Phi_i$ is linear with respect to $\alpha$
\begin{equation} \label{eqn:lin_con}
  \frac{\dif^{\,q}}{\dif t^q}\Phi_{jk}(t) = (\Delta_j)^q \sum\limits_i \alpha_{ijk} \cdot \frac{\dif^{\,q}}{\dif s^q} p_i(s)
\end{equation}

To ensure continuity of our spline up to $3$ derivatives, we need to add the following constraints for $j = 2..(m-1)$, $q = 0..3$, and $k = 1..d$:
\begin{equation} \label{eqn:quad_cont}
  (\Delta_j)^q \sum\limits_i \alpha_{ijk} \cdot \frac{\dif^{\,q}}{\dif s^q} p_i(1) = (\Delta_{j+1})^q \sum\limits_i \alpha_{i(j+1)k} \cdot \frac{\dif^{\,q}}{\dif s^q} p_i(0)
\end{equation}
The start and end constraints:
\begin{equation} \label{eqn:quad_boundary}
	\begin{array} {c}
    (\Delta_1)^q \sum\limits_i \alpha_{i1k} \cdot \frac{\dif^{\,q}}{\dif s^q} p_i(0) = \frac{\dif^{\,q}}{\dif t^q} p_0 \\
        (\Delta_m)^q \sum\limits_i \alpha_{imk} \cdot \frac{\dif^{\,q}}{\dif s^q} p_i(1) = \frac{\dif^{\,q}}{\dif t^q} p_f
	\end{array}
\end{equation}

For the inequality constraints and the centering part of the cost functional, we use the sub-sampling method proposed by \cite{Mellinger2011}.  Along the interval, we can select $g$ points at which to sample the trajectory.  In practice we found that $g = 10$ points worked well and they could be sample uniformly within $s_g \in [0,1]$ or by using Chebyshev sampling. The $a_{vk}$ and $b_v$ come from the polyhedra found in $\ref{sec:rils}$ for $q=0$ and $L_1$ bounds for $q>0$ \cite{boyd_convex_2004}.

\begin{equation} \label{eqn:quad_ineq}
  \frac{\dif^{\,q}}{\dif t^q} \sum_k \Phi_{jk}(s_g) a_{vk} \leq b_v
\end{equation}

To compute the centering part of the cost, we could use a Gaussian quadrature, but found that rectangular integration worked fine \cite{press_numerical_1992}.
\begin{equation}  \label{eqn:quad_epsilon}
\sum_{jg} \epsilon\left(\frac{l\cdot \Phi(s_g)}{\norm{l}}\right)^2
\end{equation}

This results in the following QP in $\alpha$

\begin{equation} \label{eqn:qp}
  \begin{array}{rll}
  	\min\limits_{\alpha} & \alpha^T Q \alpha & \text{Eqn. } \ref{eqn:quad_cost} + \text{Eqn. } \ref{eqn:quad_epsilon}\\
    s.t & A \alpha = b & \text{Eqn. } \ref{eqn:quad_cont}\text{ and Eqn. } \ref{eqn:quad_boundary} \\
    & C\alpha \leq d & \text{Eqn. } \ref{eqn:quad_ineq}
  \end{array}
\end{equation}

Choosing $\Delta_j$ is critical to the feasibility of Equation~\eqref{eqn:qp} and the quality of the resultant trajectory. To choose the $\Delta_j$ we use the times we get by fitting a trapezoidal velocity profile through the segments, so the time per segment is based on the length of each line segment $l_j$ in the path through the environment.

\section{Experimental Results} \label{sec:experimental_results}
\subsection{Estimation benchmarking}

\begin{figure}[ht]
  \centering
  \subfigure[Position]{\includegraphics[width=0.45\columnwidth]{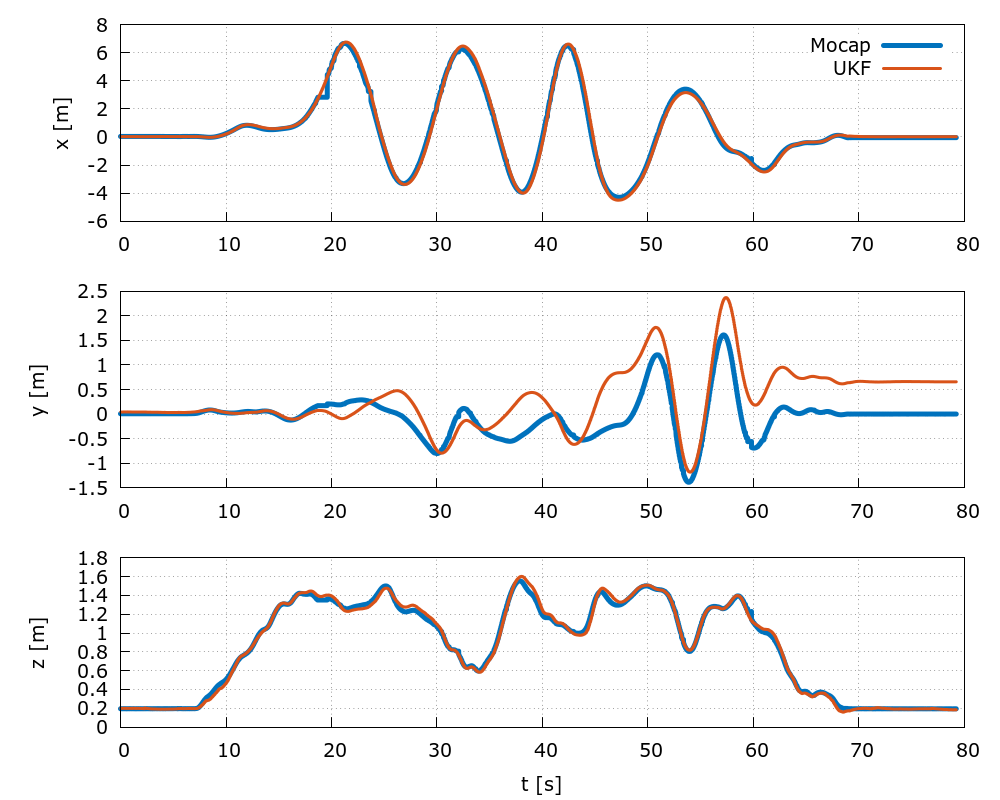}}\hspace{0.05\columnwidth}%
  \subfigure[Velocity]{\includegraphics[width=0.45\columnwidth]{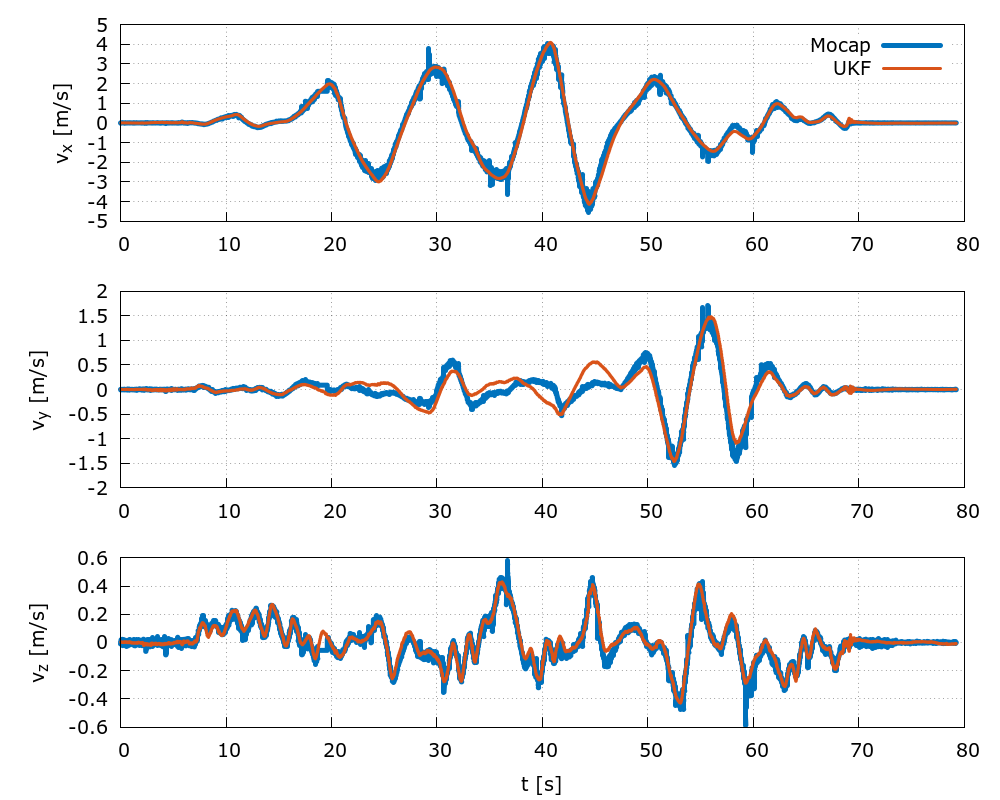}}
  \caption{Plots of position and velocity from our estimation system compared to ground truth from motion capture.} \label{fig:estimation_benchmark}
\end{figure}

The main task for the robot is to fly long distances to a goal point, so the estimation accuracy is very important. The drift in the estimator should be low so that the robot reaches the desired goal. In order to test the accuracy and drift in the estimator, we flew the robot in the motion capture space in the lab. The robot was flown manually along an aggressive trajectory reaching speeds of up to \SI{4}{\meter\per\second} and accelerations of \SI{4}{\meter\per\square\second}. The plots of the estimated and ground-truth position and velocity as shown in Figure~\ref{fig:estimation_benchmark}. As can be seen from the figure, the final drift after more than \SI{60}{\meter} of flight is less than \SI{0.6}{\meter}, giving us a drift of around $1 \%$. Note that there is almost no drift in the Z-axis due to the use of the downward pointing distance sensor, which gives us an accurate height estimate.

\begin{figure}[ht]
  \centering
  \subfigure[Onboard image with features during hover]{\includegraphics[width=0.31\columnwidth, trim=0 0 0 0, clip=true]{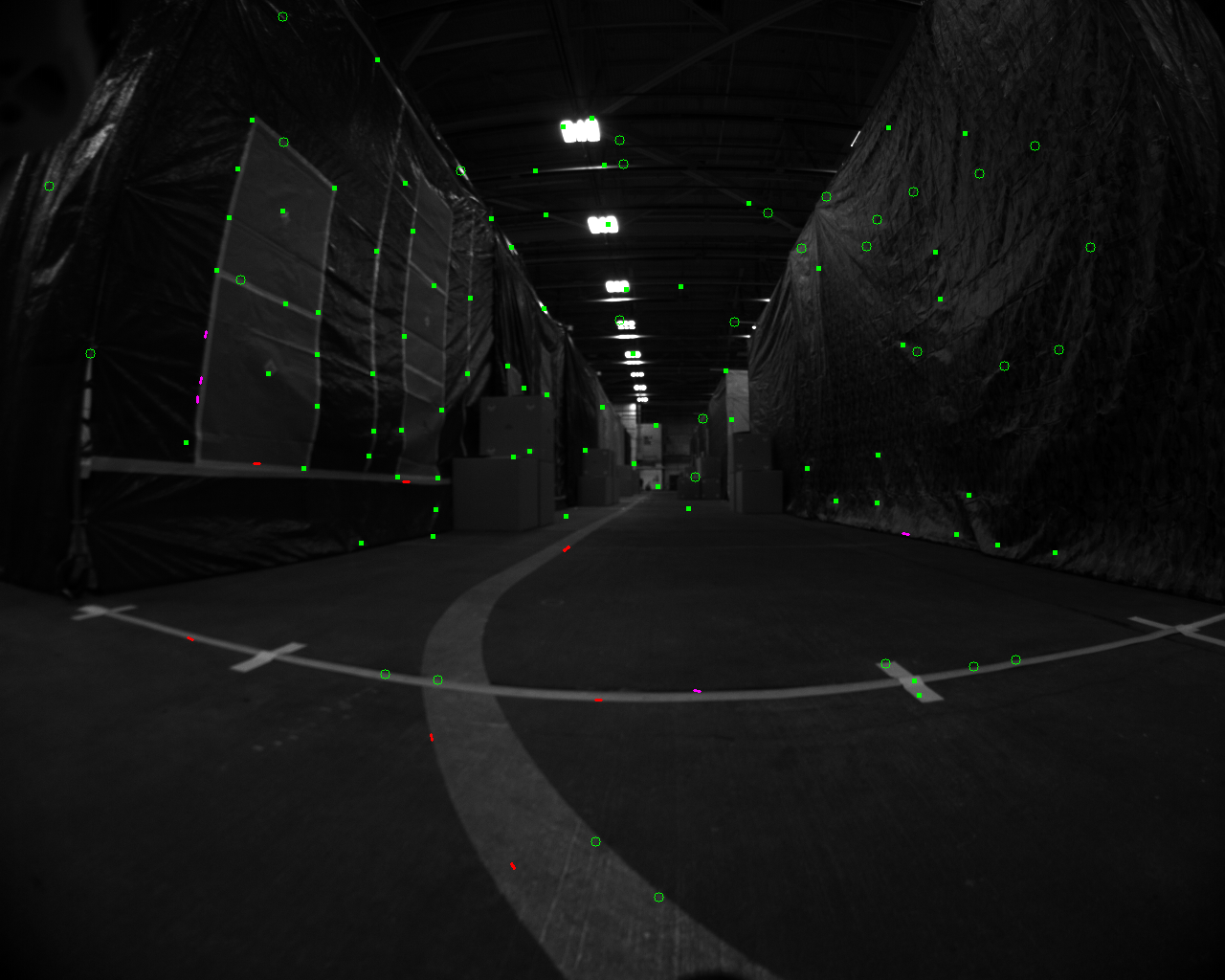}}
  \subfigure[Onboard image with features during aggressive flight]{\includegraphics[width=0.31\columnwidth, trim=0 0 0 0, clip=true]{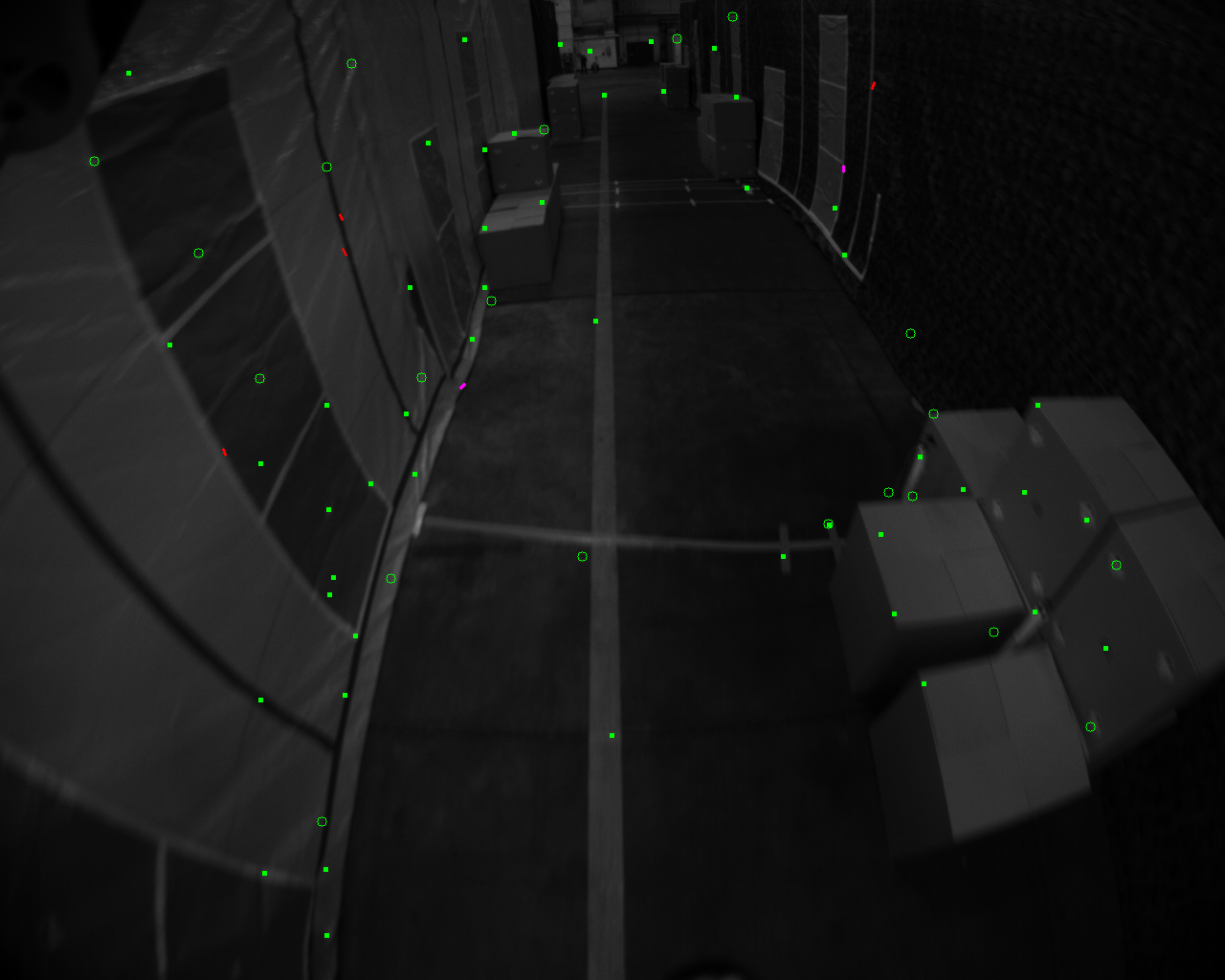}}
  \subfigure[Map of sparse 3D points]{ \includegraphics[width=0.31\columnwidth, trim=0 0 0 0, clip=true]{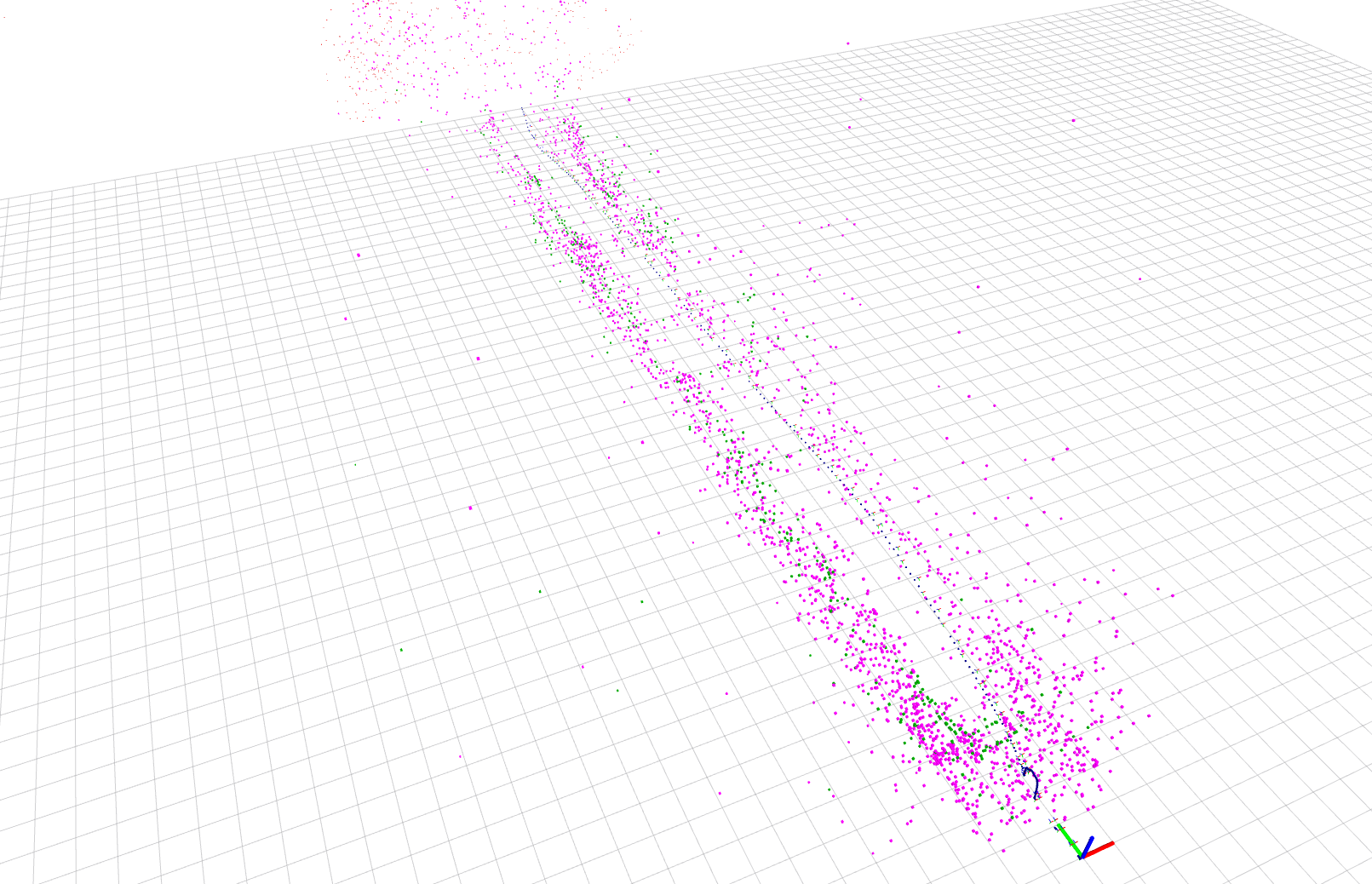}}
  \caption{Camera images from onboard the MAV show good feature tracking performance from SVO, even at high speed. The resulting sparse map of 3D points that have been triangulated is consistent and metrically accurate with the actual structure of the environment. \label{fig:svo_exp_imgs}}
\end{figure}

\begin{figure}[ht]
  \centering
  \subfigure[Estimated position of the MAV]{\includegraphics[width=0.4\columnwidth, trim=0 0 0 0, clip=true]{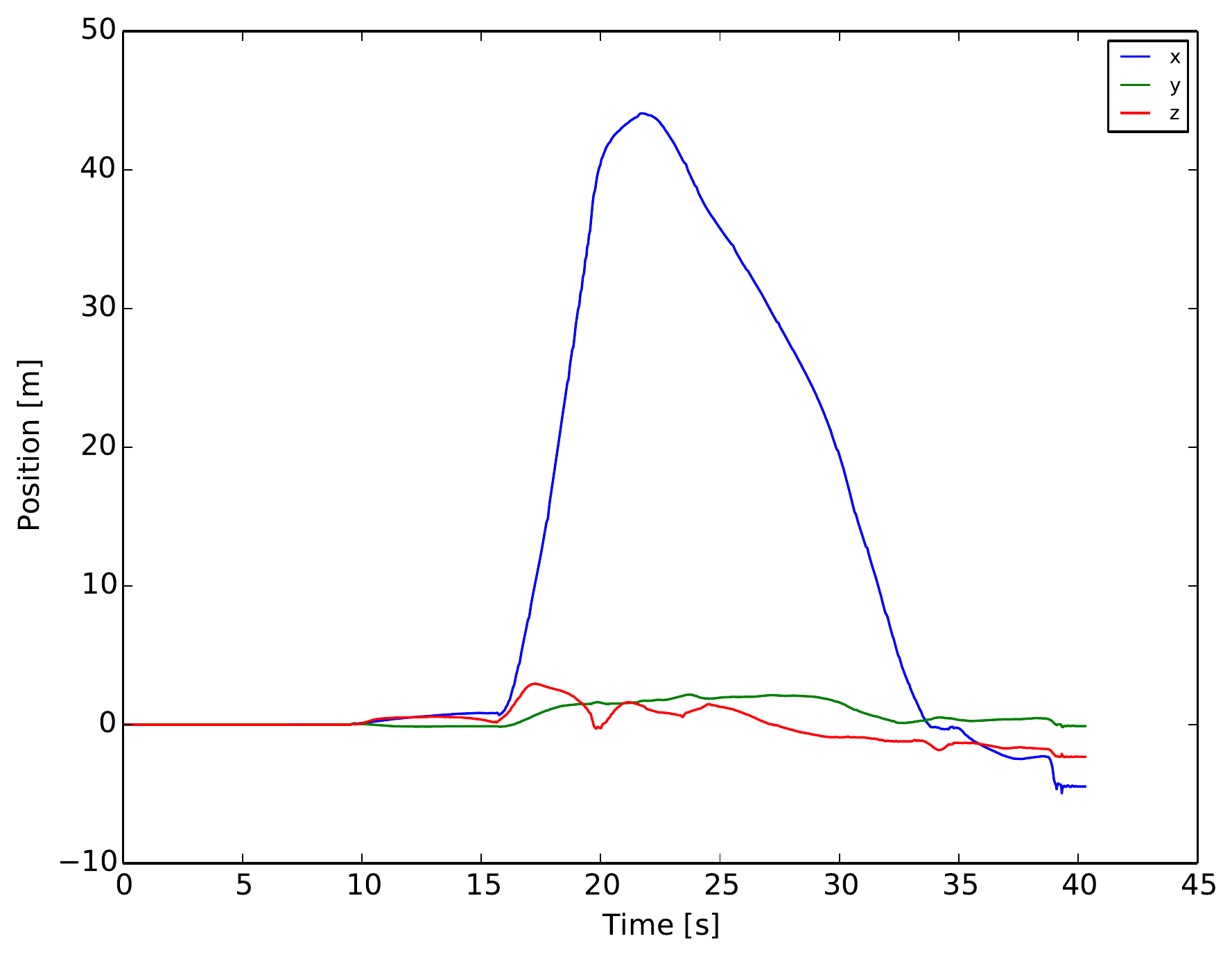}}
  \subfigure[Estimated velocity of the MAV]{\includegraphics[width=0.4\columnwidth, trim=0 0 0 0, clip=true]{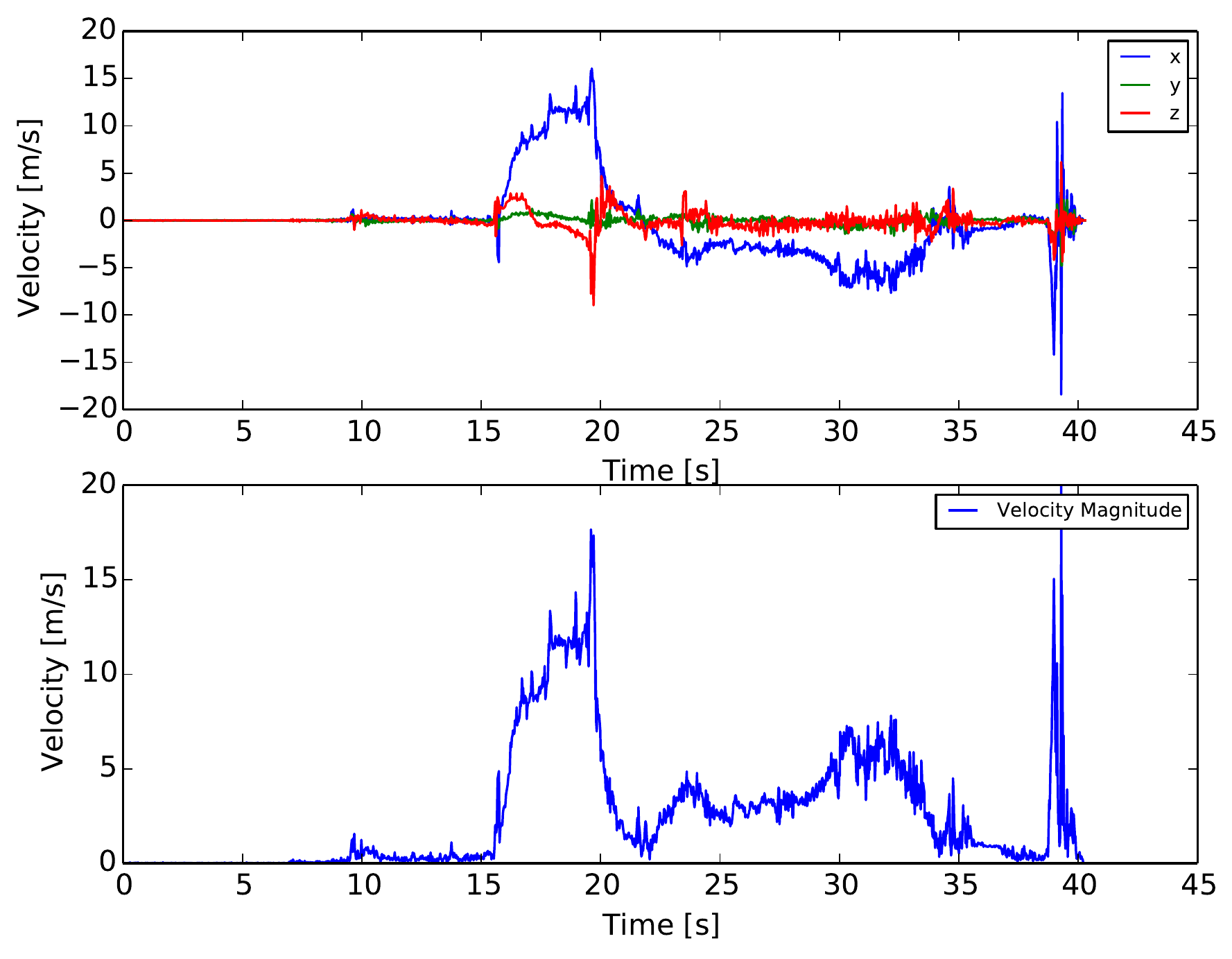}}
  \caption{Motion estimation of the MAV during a high-speed, straight line trajectory.  SVO provides a smooth pose estimate of this aggressive flight, which reached a speed of over \SI{15}{\meter\per\second} over \SI{50}{\meter}.}
   \label{fig:svo_exp_plots}
\end{figure}

The SVO framework was deployed on our MAV system for visual odometry using a forward-facing stereo camera configuration and onboard computation.
As a demonstration of the accuracy of the motion estimation, some high-speed maneuvers were flown manually in a warehouse environment.
The MAV accelerated aggressively along a \SI{50}{\meter} straight aisle in the warehouse, braked aggressively to a stop, and then returned to the starting location at a moderate speed.
Figure~\ref{fig:svo_exp_imgs} shows several onboard camera images marked up with the features that SVO is tracking, as well as the sparse map of 3D points in the environment that were mapped during this trajectory.
During this trial, the MAV reached a maximum speed of over \SI{15}{\meter\per\second}, as shown in Fig. \ref{fig:svo_exp_plots} even with such an aggressive flight, SVO only incurs around $\SI{2}{\meter}$ of position drift over the more than $\SI{100}{\meter}$ trajectory.
% A few still images of plots and from trials

\subsection{Real World Tests}

The quadrotor navigation system described in this paper has been tested extensively in the lab environment as well as in multiple real-world environments. The system has been used on our entry for the first test of the DARPA Fast Lightweight Autonomy (FLA) program and was able to successfully navigate multiple obstacle courses that were set up. The rules of the FLA program do not allow any human interaction after the robot is airborne, so the runs described in this section were fully autonomous.

\begin{figure}[ht]
  \centering
  \subfigure[An example obstacle course. The goal was to get to the other end of the aisle. The different types of obstacles along the length of the aisle can be seen.]{\includegraphics[width=0.4\columnwidth]{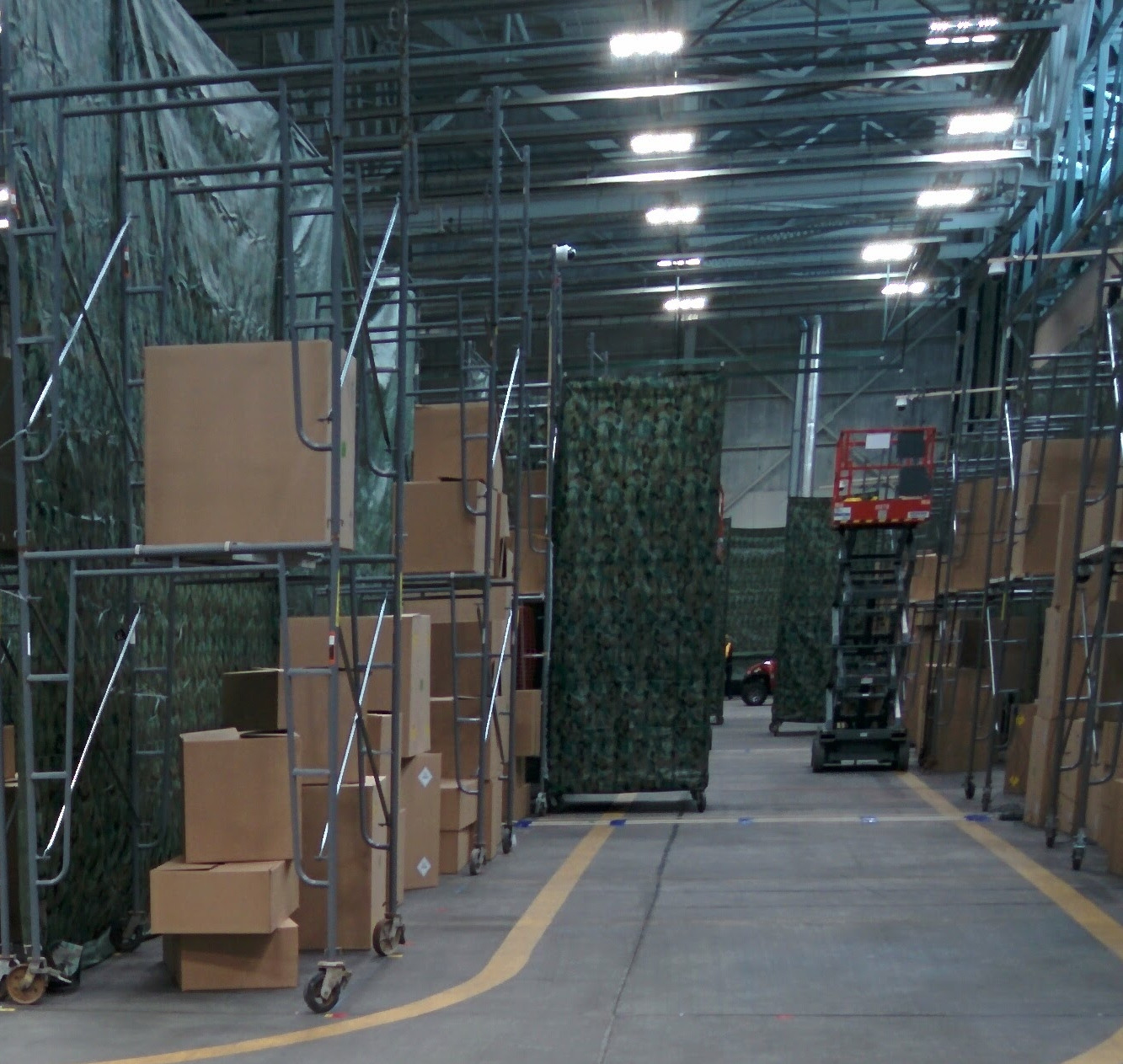}}\hspace{0.05\columnwidth}%
  \subfigure[Snapshot of the local 3D map. The color represents height, going from red on the floor to blue at \SI{4}{\meter} height and the axes in the middle of the figure represents the location of the robot.]{\includegraphics[width=0.4\columnwidth]{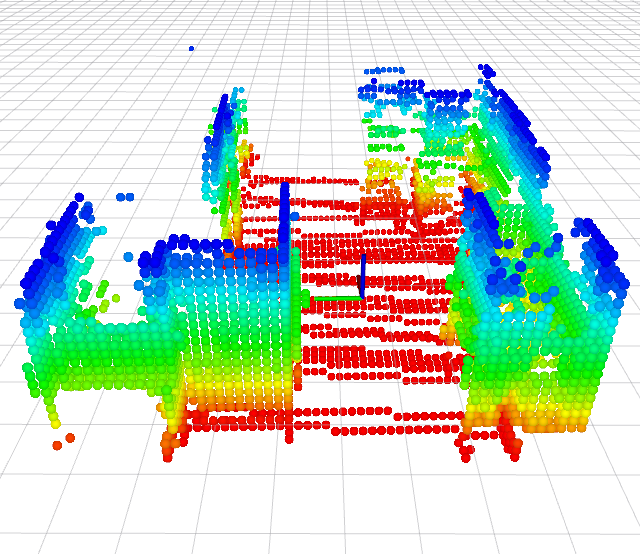}}
  \caption{An example obstacle course that the robot had to get through and a snapshot of the local 3D map constructed using the nodding laser as the robot was traversing the course. The robot was right next to the tower obstacle when the snapshot of the local map was taken. The tower obstacle (on the left) and the scissor lift (further away on the right) can be clearly seen in the 3D map.}
  \label{fig:fla_warehouse}
\end{figure}

The test environment was constructed so as to simulate the inside of a warehouse. There were two aisles separated by scaffolding of around \SI{5}{\meter} height with tarps on the back of the scaffolding and boxes placed on the shelves. The total length of the test course was around \SI{65}{\meter} while the width of each of the aisles, in between the scaffoldings, was \SI{3}{\meter}. Different types of obstacles such as a scaffolding tower or scissor lifts were placed along the aisles in order to test the obstacle avoidance performance of the robot. The minimum clearance between the obstacle in the aisle and the scaffolding on the side was set to be \SI{2.1}{\meter}. As a reference, the rotor tip-to-tip diameter of the platform is \SI{0.76}{\meter}. An example of the obstacles along the aisle is shown in Figure~\ref{fig:fla_warehouse}.

Different types of obstacle courses were set up using the aisles and the obstacles in order to challenge the robot. The simplest task was to just go straight down an empty aisle, while the most complicated ones involved changing aisles due to the first aisle being blocked in the middle.
%Some examples of the tasks are shown in Figure~\ref{fig:fla_tasks}.
The only prior information available for each task was the type of obstacles to expect along the course, but the actual layout of the test course was unknown. The goal position was specified as a bearing and a range from the starting position at the start of each run.
A particular task was deemed complete only when we were able to complete three successful runs of the task with different obstacle course layouts, thus ensuring that our system can work robustly.
In the following, we describe some of the specific tasks and also show results of our runs through them.

\subsubsection{Slalom}
\begin{figure}[ht]
  \centering
  \includegraphics[width=0.7\columnwidth]{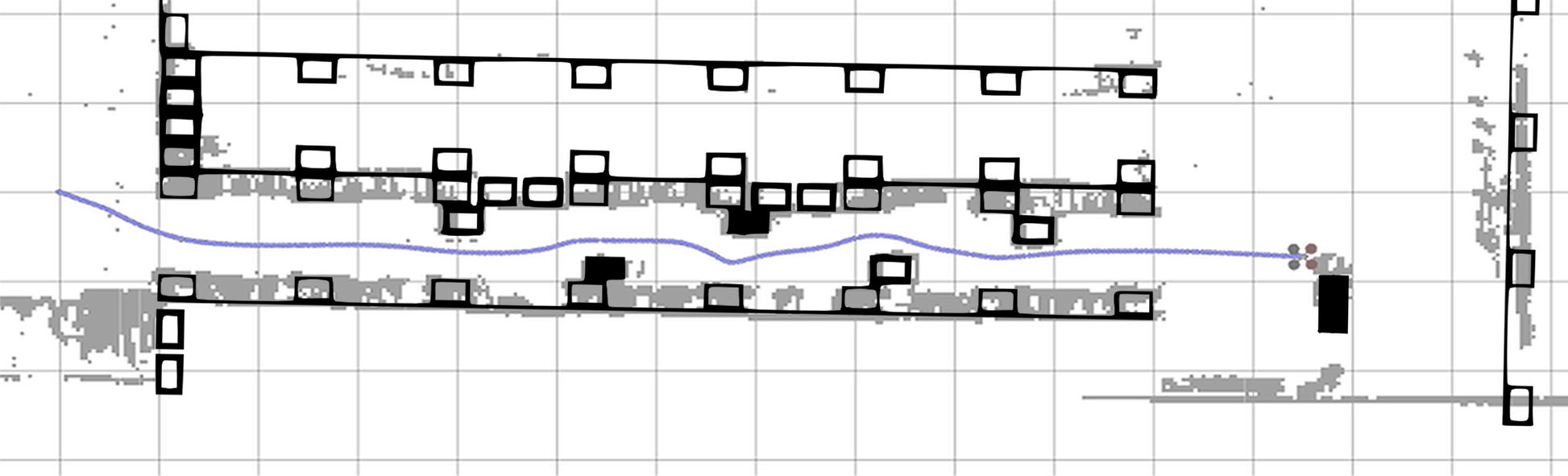}
  \caption{One of our runs for the slalom task. In black we show the actual obstacle course layout. The hollow obstacles in the aisle are similar to the tower shown in Figure~\ref{fig:fla_warehouse} while the filled black ones are scissor lifts. The gray regions are the projection of map created by the robot onto the 2D plane. The robot starts near the opening on the left and has to reach the target represented by the black rectangle on the right. The path of the robot shows it moving in a zigzag fashion in order to avoid the obstacles. Each grid cell is \SI{5}{\meter}$\,\times\,$\SI{5}{\meter}.}
  \label{fig:slalom}
\end{figure}

In the slalom task, the obstacles in an aisle were arranged in a manner such that robot is forced to move in a zigzag manner along the aisle, going to the right of the first obstacle, then left of the second, one and so on. Figure~\ref{fig:slalom} shows the result of one of our runs. Since there was no ground truth position data available, the only way to judge the performance of the system is to compare the map created by the robot with a map of the real obstacle course. From the figure, we can see that the projected map (in grey) matches the actual obstacle course layout (in black) showing the accuracy of our estimation and control algorithms.

\subsubsection{Aisle change with \SI{45}{\degree} transition}
\begin{figure}[ht]
  \centering
  \includegraphics[width=0.7\columnwidth]{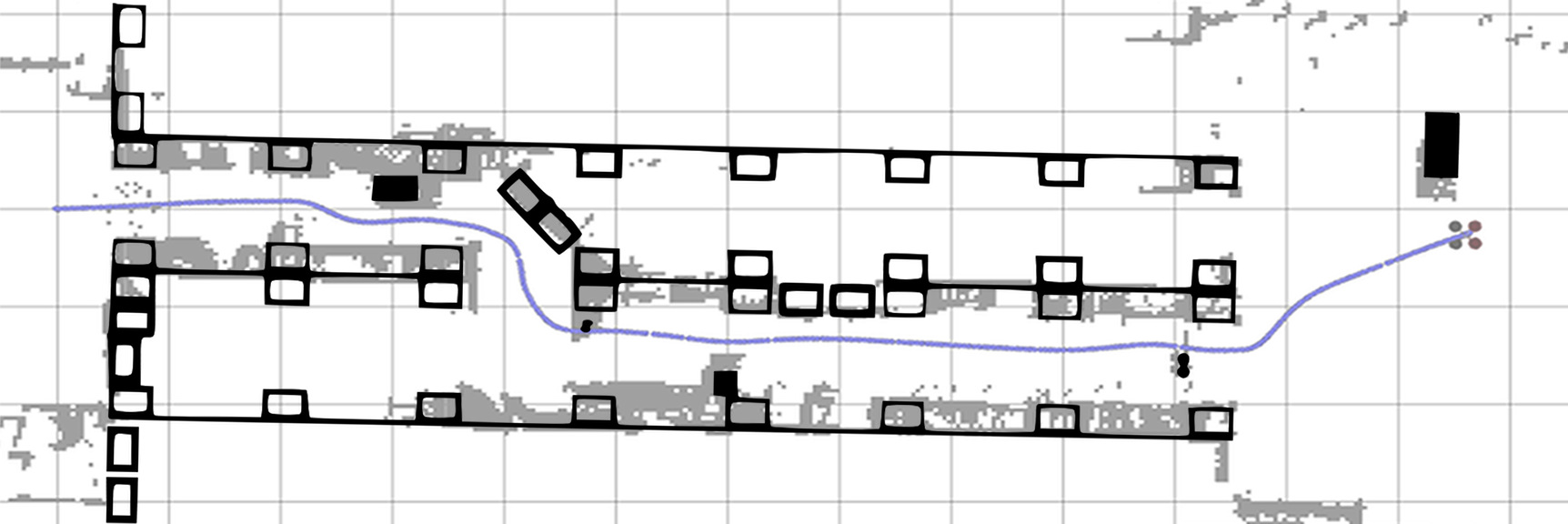}
  \caption{One of our runs for the aisle change with \SI{45}{\degree} transition task. In black we show the actual obstacle course layout. The small filled black objects along the aisles are short obstacles that the robot could fly over. The gray regions are the projection of map created by the robot onto the 2D plane. The robot starts near the opening on the left and has to reach the target represented by the black rectangle on the right. Each grid cell is \SI{5}{\meter}$\,\times\,$\SI{5}{\meter}.}
  \label{fig:aisle_change_45}
\end{figure}

In this task, the robot was required to change from the first aisle to the second one since the first aisle was blocked in the middle. The opening between the first and second aisles was constructed such that the robot could move diagonally along a \SI{45}{\degree} line from the first aisle into the second aisle. Figure~\ref{fig:aisle_change_45} shows one of our runs for this task. The robot successfully completes the transition and starts moving along the second aisle. Note that the goal was still in line with the first aisle, so the robot is always looking to move towards the left in order to get closer to the goal. This causes it to remain in the left part of the second aisle as is observed in the figure. After crossing the second aisle, the robot moves back to the left to reach the goal. Again we can see that the projected map (in grey) matches the actual course layout (in black).

\subsubsection{Aisle change with \SI{90}{\degree} transition}
\begin{figure}[ht]
  \centering
  \includegraphics[width=0.7\columnwidth]{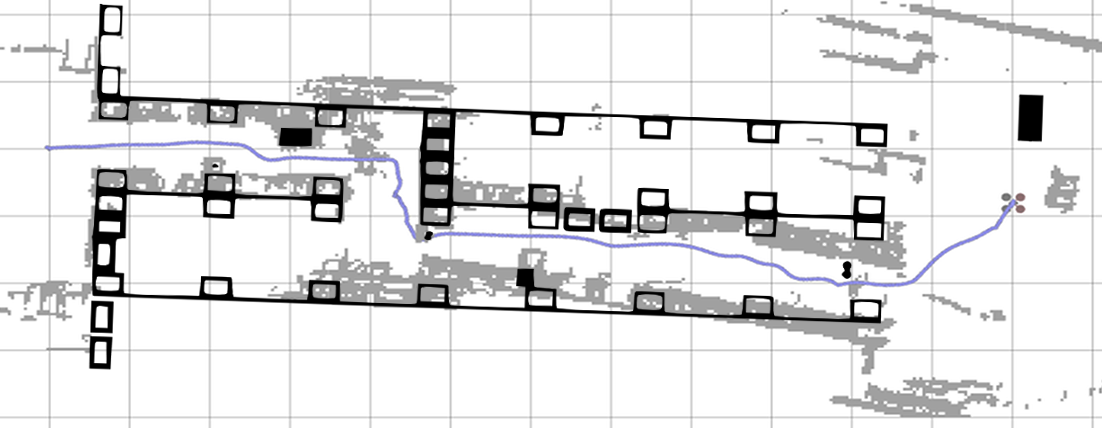}
  \caption{One of our runs for the aisle change with \SI{90}{\degree} transition task. In black we show the actual obstacle course layout. The small filled black objects along the aisles are short obstacles that the robot could fly over. The gray regions are the projection of map created by the robot onto the 2D plane. The robot starts near the opening on the left and has to reach the target represented by the black rectangle on the right. Each grid cell is \SI{5}{\meter}$\,\times\,$\SI{5}{\meter}.}
  \label{fig:aisle_change_90}
\end{figure}

This task was just a more challenging variation of the previous one. Here the aisle change required the robot to move sideways (see Figure~\ref{fig:aisle_change_90}). We were able to reach the goal, but our system did not perform very well for this task. As can be seen from the figure, the state estimate had small jumps and drifted during the transition between the aisles. There is some position drift but the main issue is the drift in yaw. Since the distance to the goal is large, even small drifts in yaw correspond to large position errors when the robot reaches the goal. The main reason for this drift was that when moving sideways in front of the obstacle during the transition, the vision system lost all the tracked features in the image and as it entered the second aisle, got new features which were far from the camera since it was looking along the aisle. Since the new features were far from the camera, they could not be triangulated accurately and hence caused bad estimates from the vision system. During this phase, there were a number of jumps in the output of vision system and hence led to drifts in the state estimate. As the robot started moving forward after the transition, the vision system was able to triangulate more features along the corridor and get good estimates again. This issue does not occur for the \SI{45}{\degree} transition case since the robot is able to see some part of the second aisle when moving diagonally and hence already has well triangulated features when it is completes the transition into the second aisle. One way to help with this issue would be to make the robot orient itself such that it is always facing the direction along which it is moving.

\subsection{High speed flight}
\begin{figure}[ht]
  \centering
  \subfigure[Position]{\includegraphics[width=0.45\columnwidth]{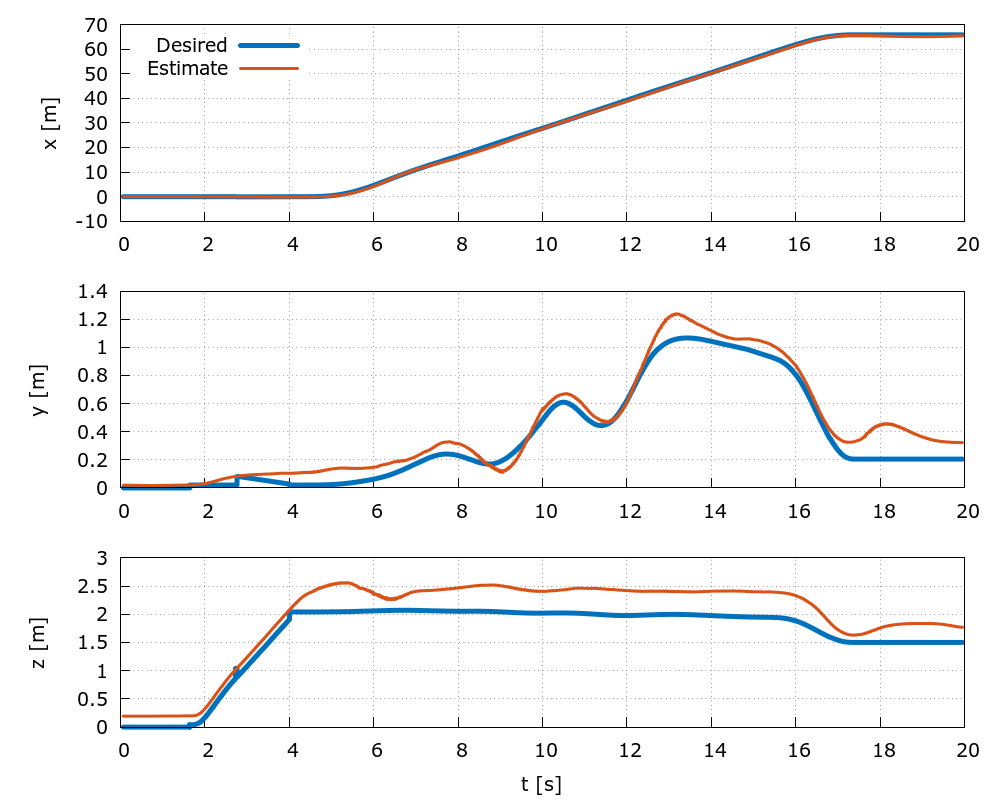}}\hspace{0.05\columnwidth}%
  \subfigure[Velocity]{\includegraphics[width=0.45\columnwidth]{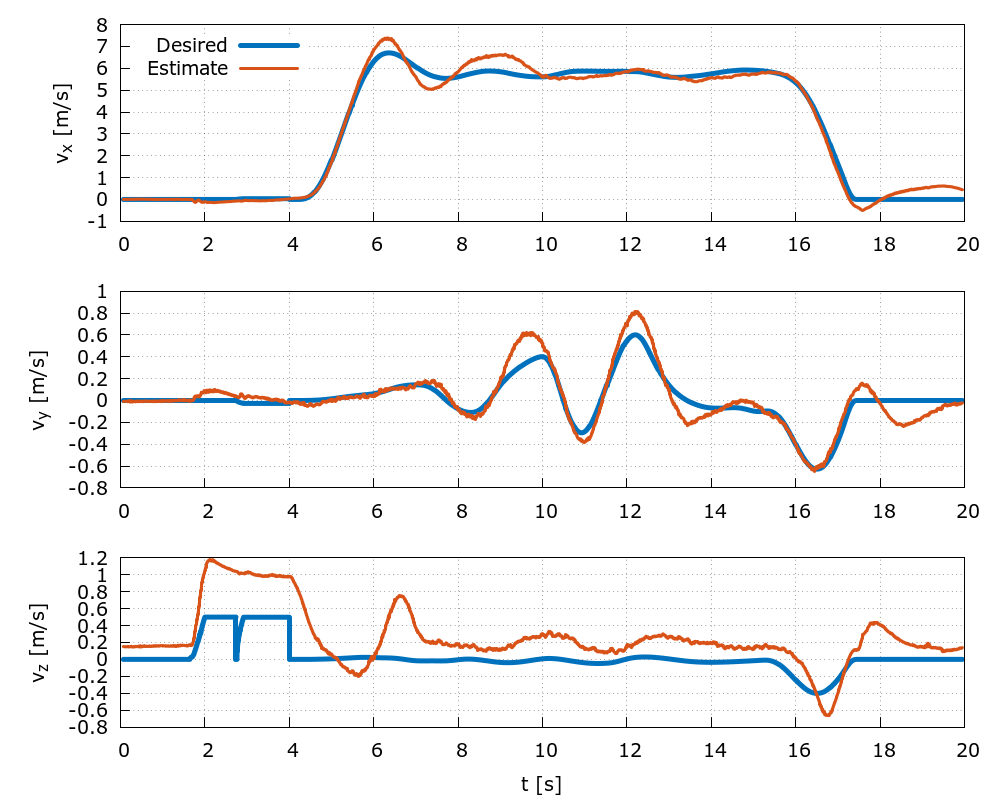}}
  \caption{Plots showing the control performance when running the full navigation system in an empty aisle. During the flight, the robot reaches speeds of up to \SI{7}{\meter\per\second}.}
  \label{fig:high_speed}
\end{figure}

%In order to test the estimation and control performance during high speed flight, we removed the lidar used for mapping and ran a straight line trajectory of around \SI{70}{\meter} with no replanning.
In order to test the high speed capability of the system, we performed a test run in an aisle with no obstacles. The goal provided to the robot was to go straight for a distance of \SI{65}{\meter}.
We were able to fly at speeds of up to \SI{7}{\meter\per\second} and reach the desired goal position.
A plot of the desired and estimated position and velocity is shown in Figure~\ref{fig:high_speed}, which shows that the performance of our controller is good enough to track such aggressive trajectories. The initial section of the plots, from \SIrange{0}{4}{\second}, is the autonomous takeoff and the forward trajectory begins at $t =$ \SI{4}{\second}. There was no source of ground truth during the test but based on the expected location of the goal, the net drift in the position estimates was less than \SI{2}{\meter}.

\section{Discussion and Conclusion} \label{sec:discussion_conclusion}
In this work, we developed a system that allows a quadrotor to navigate autonomously in GPS-denied and cluttered environments. Our navigation system consists of a set of modules that work together in order to allow the robot to go from a starting position to a specified goal location while avoiding obstacles on the way. After developing our system, we found the following points especially important in order to successfully build such a system:
\begin{itemize}
  \item \textbf{Modular architecture:} During our development process, each of the modules were separately developed. This was made possible by defining proper interfaces between the modules and using message passing to communicate among them. We used ROS as the framework for all the software running on the robot since it was designed to solve this exact problem. This separation of the modules allowed most of the planner development to happen in a simulator while the estimation and control modules were being developed. This accelerated the development since different modules could be implemented and tested in parallel.
  \item \textbf{Sensor selection:} The choice of sensors used for estimation and mapping plays an important role in determining the robustness of the system. As shown in Table~\ref{tab:vo_algorithms}, there are various advantages and disadvantages of different camera configurations for visual odometry. We selected a stereo configuration for our system since it provides increased robustness over a monocular camera setup, which is gaining popularity among the research community due to its minimalistic nature and also allows us to have simpler algorithms compared to multi-camera systems. The use of a dedicated height sensor makes it possible to maintain altitude even when there is drift in our visual odometry, allowing the robot to safely fly without hitting the ground or going too high. However, the downward pointing height sensor has jumps in the measurement when the robot goes over obstacles and this has to be properly taken care of in the sensor fusion module. For mapping, instead of using a fixed lidar, we mounted it on a servo in order to sweep it up and down allowing us to create 3D maps and navigate in 3D environments with obstacles above and below the robot. With a fixed lidar we would not have been able to safely avoid all the obstacles that we encountered during the tests.

\item \textbf{Local map for planning:} Using a local map for planning instead of a global map was an crucial decision in the design of our planner. The problem with creating a global map is that we need to explicitly maintain global consistency by making use of loop closures to eliminate drifts. By comparison, the local map approach helps the planner tolerate drifts in the state estimation since the drift is small in the short period of time that the local map is constructed in. This can be seen clearly in Figure~\ref{fig:aisle_change_90} where there is large drift in the yaw but the robot is still able to reach the goal. This also helps in reducing the computational complexity and allows us to run the planner at a higher rate. Faster replanning reduces the latency between an obstacle being seen by the mapping system and the robot reacting to it, thus improving the robustness of the system.
\end{itemize}

In addition to these positive points, we learned some lessons during the tests in the warehouse environment.
\begin{itemize}
  \item \textbf{Drift in the visual odometry:} The tests in the warehouse environment involved flying long trajectories while constantly moving forward. Since the floor of the building was smooth and did not have texture, very few image features could be detected on it. This led to most of our image features coming from the obstacles to the side and front of the robot. We even picked up edgelet features from structures on the ceiling of the building. Thus a large part of the image features were at a large distance from the robot. In order to get good depth estimates of these far-away features, either the stereo baseline needs to be large or there needs to be sufficient parallax between the feature observations due to the motion between frames. We were limited to a \SI{0.2}{\meter} stereo baseline due to the size of the robot. When moving along the long aisles, the image features were mainly in front of the robot, which sometimes led to insufficient parallax to get good depth estimates for the features. Due to the poor depth estimates for some of the features, the visual odometry was not able to detect the correct scale of the motion between frames, which led to drift in the estimates. This caused a failure to reach the goal in some cases. One solution to this, that we are already looking into, is to have a more tightly coupled visual odometry system where the accelerometer measurement is also used in order to provide another source of scale for the visual odometry system.
  \item \textbf{Local map size:} One factor that prevents us from reaching high speeds is the size of the map used for planning. Since we want to generate dynamically feasible trajectories for the robot, we have to take into account the maximum acceleration that the robot can safely achieve. Also, in order to guarantee safety, we have to plan trajectories such that the robot comes to a halt at the end of the known map since there can be undiscovered obstacles just outside the map. Thus, the combination of a map size and maximum acceleration puts a limit on the maximum speed that the robot can reach. The main factor limiting our local map size is the time required to plan in that map. The majority of the time in each planning step is taken by the $A^*$ algorithm, which is used to find a path through the hybrid graph (as described in Section~\ref{sec:planning}). In order to reduce this time, we are looking into using better heuristics for $A^*$ and other techniques such as Jump Point Search \cite{Harabor2011} which can significantly speed up the graph search.
\end{itemize}

In conclusion, we have presented a solution that consists of all the modules that are required for a robot to autonomously navigate in an unknown environment. The system has been designed such that all the sensing and computation occur onboard the robot. Once the robot has been launched, there is no human interaction necessary for the robot to navigate to the goal.

The system has been thoroughly tested in the lab as well as in the warehouse environment that was set up as part of the DARPA FLA program. Our robot was able to successfully navigate the various obstacle courses that were specifically designed to challenge the navigation system. The only input from the operator for each run was the goal position relative to the starting position. In fact, during some of the runs, we even lost the communication link between the base station (which was only used for monitoring purposes) and the robot, due to the long distance and the large scaffolding structures in between, but the robot kept on going and successfully completed the task.

The final goal is to be able to fly at speeds of around \SI{20}{\meter\per\second} through cluttered environments, and we believe that it would require more work in all the individual modules that make up the system. In estimation, we need to reduce the drift that the visual odometry system experiences when flying fast while following long trajectories. In control, we need to incorporate aerodynamic effects such as drag, which become increasingly important when flying fast. In the mapping part, the nodding lidar solution needs to be replaced by one which provides a denser representation of the environment in order to detect small obstacles reliably. And finally, the planning subsystem needs to be sped up in order to allow us to use a larger map for planning and also to allow faster replanning in order to make the system more robust. As these developments are made, we would be able to incorporate them into our system due to the modular architecture, thus providing a strong foundation for future research.

\subsubsection*{Acknowledgments}
We gratefully acknowledge support from DARPA grants HR001151626/HR0011516850.
%The authors would like to thank the following people for their support in this project: Anurag Makineni and Giuseppe Loianno for help with the Pixhawk autopilot, Kelsey Saulnier for implementing the mission file parser and Ke Sun for the Camera-IMU synchronization driver.

\bibliographystyle{apalike}
\bibliography{references}

\begin{thebibliography}{}

\bibitem[Achtelik et~al., 2009]{Achtelik2009}
Achtelik, M., Bachrach, A., He, R., Prentice, S., and Roy, N. (2009).
\newblock Stereo vision and laser odometry for autonomous helicopters in
  gps-denied indoor environments.
\newblock volume 7332, pages 733219--733219--10.

\bibitem[Augugliaro et~al., 2014]{Augugliaro_CSM_14}
Augugliaro, F., Lupashin, S., Hamer, M., Male, C., Hehn, M., Mueller, M.~W.,
  Willmann, J., Gramazio, F., Kohler, M., and D'Andrea, R. (2014).
\newblock {T}he {F}light {A}ssembled {A}rchitecture installation: {C}ooperative
  construction with flying machines.
\newblock {\em IEEE Control Systems Magazine}, 34(4):46--64.

\bibitem[Bachrach et~al., 2009]{Bachrach2009}
Bachrach, A., He, R., and Roy, N. (2009).
\newblock Autonomous flight in unknown indoor environments.
\newblock {\em International Journal of Micro Air Vehicles}, 1(4):217--228.

\bibitem[Bachrach et~al., 2011]{Bachrach2011}
Bachrach, A., Prentice, S., He, R., and Roy, N. (2011).
\newblock {RANGE -- Robust Autonomous Navigation in GPS-Denied Environments}.
\newblock {\em Journal of Field Robotics}, 28(5):644--666.

\bibitem[Bellingham et~al., 2002]{Bellingham2002}
Bellingham, J., Richards, A., and How, J.~P. (2002).
\newblock {Receding horizon control of autonomous aerial vehicles}.
\newblock In {\em Proceedings of the 2002 American Control Conference (IEEE
  Cat. No.CH37301)}, volume~5, pages 3741--3746 vol.5.

\bibitem[Bloesch et~al., 2015]{Bloesch2015}
Bloesch, M., Omari, S., Hutter, M., and Siegwart, R. (2015).
\newblock Robust visual inertial odometry using a direct ekf-based approach.
\newblock In {\em 2015 IEEE/RSJ International Conference on Intelligent Robots
  and Systems (IROS)}, pages 298--304.

\bibitem[Blösch et~al., 2010]{Blosch2010}
Blösch, M., Weiss, S., Scaramuzza, D., and Siegwart, R. (2010).
\newblock Vision based mav navigation in unknown and unstructured environments.
\newblock In {\em 2010 IEEE International Conference on Robotics and
  Automation}, pages 21--28.

\bibitem[Bouabdallah et~al., 2004]{Bouabdallah2004}
Bouabdallah, S., Murrieri, P., and Siegwart, R. (2004).
\newblock Design and control of an indoor micro quadrotor.
\newblock In {\em Robotics and Automation, 2004. Proceedings. ICRA '04. 2004
  IEEE International Conference on}, volume~5, pages 4393--4398 Vol.5.

\bibitem[Bouabdallah and Siegwart, 2005]{Bouabdallah2005}
Bouabdallah, S. and Siegwart, R. (2005).
\newblock Backstepping and sliding-mode techniques applied to an indoor micro
  quadrotor.
\newblock In {\em Proceedings of the 2005 IEEE International Conference on
  Robotics and Automation}, pages 2247--2252.

\bibitem[Bouabdallah and Siegwart, 2007]{Bouabdallah2007}
Bouabdallah, S. and Siegwart, R. (2007).
\newblock Full control of a quadrotor.
\newblock In {\em 2007 IEEE/RSJ International Conference on Intelligent Robots
  and Systems}, pages 153--158.

\bibitem[Boyd and Vandenberghe, 2004]{boyd_convex_2004}
Boyd, S. and Vandenberghe, L. (2004).
\newblock {\em Convex {Optimization}}.
\newblock Cambridge University Press, New York, NY, USA.

\bibitem[Civera et~al., 2008]{Civera08tro}
Civera, J., Davison, A., and Montiel, J. (2008).
\newblock {Inverse Depth Parametrization for Monocular SLAM}.
\newblock {\em {IEEE} Trans. Robotics}, 24(5).

\bibitem[Deits and Tedrake, 2015a]{deits2015computing}
Deits, R. and Tedrake, R. (2015a).
\newblock {Computing large convex regions of obstacle-free space through
  semidefinite programming}.
\newblock In {\em Algorithmic Foundations of Robotics XI}, pages 109--124.
  Springer.

\bibitem[Deits and Tedrake, 2015b]{Deits2015}
Deits, R. and Tedrake, R. (2015b).
\newblock {Efficient Mixed-Integer Planning for UAVs in Cluttered
  Environments}.
\newblock In {\em 2015 IEEE International Conference on Robotics and Automation
  (ICRA)}, pages 42--49.

\bibitem[Escareno et~al., 2006]{Escareno2006}
Escareno, J., Salazar-Cruz, S., and Lozano, R. (2006).
\newblock Embedded control of a four-rotor uav.
\newblock In {\em 2006 American Control Conference}, pages 6 pp.--.

\bibitem[Forster et~al., 2014]{Forster2014}
Forster, C., Pizzoli, M., and Scaramuzza, D. (2014).
\newblock {SVO}: Fast semi-direct monocular visual odometry.
\newblock In {\em IEEE International Conference on Robotics and Automation
  (ICRA)}.

\bibitem[Forster et~al., 2017]{Forster2016svo}
Forster, C., Zhang, Z., Gassner, M., Werlberger, M., and Scaramuzza, D. (To
  appear, 2017).
\newblock {SVO: Semi-Direct Visual Odometry for Monocular and Multi-Camera
  Systems}.
\newblock {\em IEEE Transactions on Robotics}.

\bibitem[Grzonka et~al., 2009]{Grzonka2009}
Grzonka, S., Grisetti, G., and Burgard, W. (2009).
\newblock Towards a navigation system for autonomous indoor flying.
\newblock In {\em 2009 IEEE International Conference on Robotics and
  Automation}, pages 2878--2883.

\bibitem[Guenard et~al., 2005]{Guenard2005}
Guenard, N., Hamel, T., and Moreau, V. (2005).
\newblock {Dynamic modeling and intuitive control strategy for an
  ``X4-flyer''}.
\newblock In {\em 2005 International Conference on Control and Automation},
  volume~1, pages 141--146.

\bibitem[Harabor and Grastien, 2011]{Harabor2011}
Harabor, D. and Grastien, A. (2011).
\newblock Online graph pruning for pathfinding on grid maps.
\newblock In {\em Proceedings of the Twenty-Fifth AAAI Conference on Artificial
  Intelligence}, AAAI'11, pages 1114--1119. AAAI Press.

\bibitem[He et~al., 2008]{He2008}
He, R., Prentice, S., and Roy, N. (2008).
\newblock Planning in information space for a quadrotor helicopter in a
  gps-denied environment.
\newblock In {\em 2008 IEEE International Conference on Robotics and
  Automation}, pages 1814--1820.

\bibitem[Hoffmann et~al., 2007]{Hoffmann2007}
Hoffmann, G., Huang, H., Waslander, S., and Tomlin, C. (2007).
\newblock Quadrotor helicopter flight dynamics and control: Theory and
  experiment.
\newblock In {\em AIAA Guidance, Navigation and Control Conference and
  Exhibit}, page 6461.

\bibitem[Jones and Soatto, 2011]{Jones2011}
Jones, E.~S. and Soatto, S. (2011).
\newblock Visual-inertial navigation, mapping and localization: A scalable
  real-time causal approach.
\newblock {\em The International Journal of Robotics Research}, 30(4):407--430.

\bibitem[Julier et~al., 1995]{Julier1995}
Julier, S.~J., Uhlmann, J.~K., and Durrant-Whyte, H.~F. (1995).
\newblock {A new approach for filtering nonlinear systems}.
\newblock In {\em American Control Conference, Proceedings of the 1995},
  volume~3, pages 1628--1632 vol.3.

\bibitem[Karydis and Kumar, 2016]{Karydis_RSIF_17}
Karydis, K. and Kumar, V. (2016).
\newblock Energetics in robotic flight at small scales.
\newblock {\em Interface Focus}, 7(1).

\bibitem[Klein and Murray, 2007]{Klein2007}
Klein, G. and Murray, D. (2007).
\newblock Parallel tracking and mapping for small {AR} workspaces.
\newblock In {\em Proc. Sixth {IEEE} and {ACM} International Symposium on Mixed
  and Augmented Reality {(ISMAR'07)}}, Nara, Japan.

\bibitem[Lee et~al., 2010]{Lee2010}
Lee, T., Leok, M., and McClamroch, N.~H. (2010).
\newblock {Geometric Tracking Control of a Quadrotor UAV on SE(3)}.
\newblock In {\em 2010 49th IEEE Conference on Decision and Control (CDC)},
  pages 5420--5425.

\bibitem[Liu et~al., 2016]{Sikang2016}
Liu, S., Watterson, M., Tang, S., and Kumar, V. (2016).
\newblock High speed navigation for quadrotors with limited onboard sensing.
\newblock In {\em 2016 IEEE International Conference on Robotics and Automation
  (ICRA)}, pages 1484--1491.

\bibitem[Meier et~al., 2015]{pixhawk}
Meier, L., Honegger, D., and Pollefeys, M. (2015).
\newblock {PX4: A Node-Based Multithreaded Open Source Robotics Framework for
  Deeply Embedded Platforms}.
\newblock In {\em 2015 IEEE International Conference on Robotics and Automation
  (ICRA)}, pages 6235--6240.

\bibitem[Mellinger and Kumar, 2011]{Mellinger2011}
Mellinger, D. and Kumar, V. (2011).
\newblock {Minimum Snap Trajectory Generation and Control for Quadrotors}.
\newblock In {\em 2011 IEEE International Conference on Robotics and
  Automation}, pages 2520--2525.

\bibitem[Mellinger et~al., 2012]{Mellinger2012}
Mellinger, D., Kushleyev, A., and Kumar, V. (2012).
\newblock {Mixed-Integer Quadratic Program Trajectory Generation for
  Heterogeneous Quadrotor Teams}.
\newblock In {\em 2012 IEEE International Conference on Robotics and
  Automation}, pages 477--483.

\bibitem[Michael et~al., 2010]{Michael2010}
Michael, N., Mellinger, D., Lindsey, Q., and Kumar, V. (2010).
\newblock {The GRASP Multiple Micro-UAV Testbed}.
\newblock {\em IEEE Robotics \& Automation Magazine}, 17(3):56--65.

\bibitem[Mohta et~al., 2016]{Mohta2016}
Mohta, K., Turpin, M., Kushleyev, A., Mellinger, D., Michael, N., and Kumar, V.
  (2016).
\newblock {\em QuadCloud: A Rapid Response Force with Quadrotor Teams}, pages
  577--590.
\newblock Springer International Publishing.

\bibitem[Mourikis and Roumeliotis, 2007]{Mourikis2007}
Mourikis, A.~I. and Roumeliotis, S.~I. (2007).
\newblock {A Multi-State Constraint Kalman Filter for Vision-aided Inertial
  Navigation}.
\newblock In {\em Proceedings 2007 IEEE International Conference on Robotics
  and Automation}, pages 3565--3572.

\bibitem[Mulgaonkar et~al., 2014]{Mulgaonkar2014SPIEDSS}
Mulgaonkar, Y., Whitzer, M., Morgan, B., Kroninger, C.~M., Harrington, A.~M.,
  and Kumar, V. (2014).
\newblock Power and weight considerations in small, agile quadrotors.
\newblock In {\em Proc. SPIE 9083, Micro- and Nanotechnology Sensors, Systems,
  and Applications VI}, volume 9083, pages 90831Q--90831Q--16.

\bibitem[Mur-Artal et~al., 2015]{Mur-Artal2015}
Mur-Artal, R., Montiel, J. M.~M., and Tardós, J.~D. (2015).
\newblock {ORB-SLAM: A Versatile and Accurate Monocular SLAM System}.
\newblock {\em IEEE Transactions on Robotics}, 31(5):1147--1163.

\bibitem[Pizzoli et~al., 2014]{Pizzoli14icra}
Pizzoli, M., Forster, C., and Scaramuzza, D. (2014).
\newblock {REMODE: Probabilistic, Monocular Dense Reconstruction in Real Time}.
\newblock In {\em {IEEE} Int. Conf. on Robotics and Automation (ICRA)}, pages
  2609--2616.

\bibitem[Press, 1992]{press_numerical_1992}
Press, W.~H., editor (1992).
\newblock {\em Numerical recipes in {C}: the art of scientific computing}.
\newblock Cambridge University Press, Cambridge ; New York, 2nd ed edition.

\bibitem[Richter et~al., 2016]{richter2016polynomial}
Richter, C., Bry, A., and Roy, N. (2016).
\newblock Polynomial trajectory planning for aggressive quadrotor flight in
  dense indoor environments.
\newblock In {\em Robotics Research}, pages 649--666. Springer.

\bibitem[Rosten et~al., 2010]{Rosten10pami}
Rosten, E., Porter, R., and Drummond, T. (2010).
\newblock {Faster and Better: A Machine Learning Approach to Corner Detection}.
\newblock {\em {IEEE} Trans. Pattern Anal. Machine Intell.}, 32(1):105--119.

\bibitem[Schmid et~al., 2014]{Schmid2014}
Schmid, K., Lutz, P., Tomić, T., Mair, E., and Hirschmüller, H. (2014).
\newblock Autonomous vision-based micro air vehicle for indoor and outdoor
  navigation.
\newblock {\em Journal of Field Robotics}, 31(4):537--570.

\bibitem[Shen et~al., 2011]{Shen2011}
Shen, S., Michael, N., and Kumar, V. (2011).
\newblock {Autonomous Multi-Floor Indoor Navigation with a Computationally
  Constrained MAV}.
\newblock In {\em 2011 IEEE International Conference on Robotics and
  Automation}, pages 20--25.

\bibitem[Shen et~al., 2013]{Shen2013}
Shen, S., Mulgaonkar, Y., Michael, N., and Kumar, V. (2013).
\newblock Vision-based state estimation and trajectory control towards
  high-speed flight with a quadrotor.
\newblock In {\em Proceedings of Robotics: Science and Systems}, Berlin,
  Germany.

\bibitem[Theys et~al., 2016]{Theys2016}
Theys, B., Dimitriadis, G., Hendrick, P., and Schutter, J.~D. (2016).
\newblock Influence of propeller configuration on propulsion system efficiency
  of multi-rotor unmanned aerial vehicles.
\newblock In {\em 2016 International Conference on Unmanned Aircraft Systems
  (ICUAS)}, pages 195--201.

\bibitem[Thomas et~al., 2014]{Thomas_BB_2014}
Thomas, J., Loianno, G., Polin, J., Sreenath, K., and Kumar, V. (2014).
\newblock Toward autonomous avian-inspired grasping for micro aerial vehicles.
\newblock {\em Bioinspiration \& Biomimetics}, 9(2):025010.

\bibitem[Valenti et~al., 2014]{Valenti2014}
Valenti, R.~G., Dryanovski, I., Jaramillo, C., Ström, D.~P., and Xiao, J.
  (2014).
\newblock Autonomous quadrotor flight using onboard rgb-d visual odometry.
\newblock In {\em 2014 IEEE International Conference on Robotics and Automation
  (ICRA)}, pages 5233--5238.

\bibitem[Vogiatzis and Hern\'{a}ndez, 2011]{Vogiatzis11jivc}
Vogiatzis, G. and Hern\'{a}ndez, C. (2011).
\newblock {Video-based, Real-Time Multi View Stereo}.
\newblock {\em Image Vision Comput.}, 29(7):434--441.

\bibitem[Wan and Merwe, 2000]{Wan2000}
Wan, E.~A. and Merwe, R. V.~D. (2000).
\newblock {The Unscented Kalman Filter for Nonlinear Estimation}.
\newblock In {\em Proceedings of the IEEE 2000 Adaptive Systems for Signal
  Processing, Communications, and Control Symposium (Cat. No.00EX373)}, pages
  153--158.

\bibitem[Watterson and Kumar, 2015]{Watterson2015}
Watterson, M. and Kumar, V. (2015).
\newblock {Safe receding horizon control for aggressive MAV flight with limited
  range sensing}.
\newblock In {\em 2015 IEEE/RSJ International Conference on Intelligent Robots
  and Systems (IROS)}, pages 3235--3240.

\bibitem[Wolff et~al., 2014]{Wolff2014}
Wolff, E.~M., Topcu, U., and Murray, R.~M. (2014).
\newblock {Optimization-based Trajectory Generation with Linear Temporal Logic
  Specifications}.
\newblock In {\em 2014 IEEE International Conference on Robotics and Automation
  (ICRA)}, pages 5319--5325.

\bibitem[Özaslan et~al., 2016]{Ozaslan2016}
Özaslan, T., Mohta, K., Keller, J., Mulgaonkar, Y., Taylor, C.~J., Kumar, V.,
  Wozencraft, J.~M., and Hood, T. (2016).
\newblock {Towards fully autonomous visual inspection of dark featureless dam
  penstocks using MAVs}.
\newblock In {\em 2016 IEEE/RSJ International Conference on Intelligent Robots
  and Systems (IROS)}, pages 4998--5005.

\end{thebibliography}

\end{document}